%% file: arxiv.tex
\pdfoutput = 1
\documentclass[twocolumn]{article}
\input{macros}
\usepackage{style}

\makeatletter
\@namedef{ver@everyshi.sty}{}
\newcommand{\removelatexerror}{\let\@latex@error\@gobble}

\newcommand\red[1]{{\color{red}#1}}

\title{F-CAM: Full Resolution Class Activation Maps  via Guided Parametric Upscaling}

\renewcommand\footnotemark{}

\author{Soufiane~Belharbi$^{1}$,
  ~Aydin~Sarraf$^{3}$,
  ~Marco~Pedersoli$^{1}$,
  ~Ismail~Ben~Ayed$^{1}$,
  ~Luke~McCaffrey$^{2}$, and
  ~Eric~Granger$^{1}$\\
 	$^1$ LIVIA, Dept. of Systems Engineering, École de technologie supérieure, Montreal, Canada \\
	$^2$ Goodman Cancer Research Centre, Dept. of Oncology, McGill University, Montreal, Canada\\
  $^3$ Ericsson,  Global AI Accelerator, Montreal, Canada\\
{\tt\footnotesize \textcolor{black}{soufiane.belharbi.1@ens.etsmtl.ca, luke.mccaffrey@mcgill.ca}, }\\
{\tt\footnotesize \textcolor{black}{\{ismail.benayed, eric.granger, marco.pedersoli\}@etsmtl.ca}, {\tt\footnotesize \textcolor{black}{aydin.sarraf@ericsson.com}}}
}

\newcommand{\ignore}[1]{}

\begin{document}
\maketitle\thispagestyle{fancy}

\maketitle

\begin{abstract}
  Class Activation Mapping (CAM) methods have recently gained much attention for weakly-supervised object localization (WSOL) tasks. They allow for CNN visualization and interpretation without training on fully annotated image datasets. CAM methods are typically integrated within off-the-shelf CNN backbones, such as ResNet50. Due to convolution and pooling operations, these backbones yield low resolution CAMs with a down-scaling factor of up to 32, contributing to inaccurate localizations. Interpolation is required to restore full size CAMs, yet it does not consider the statistical properties of objects, such as color and texture, leading to activations with inconsistent boundaries, and inaccurate localizations.
  As an alternative, we introduce a generic method for parametric upscaling of CAMs that allows constructing accurate full resolution CAMs (F-CAMs). In particular, we propose a trainable decoding architecture that can be connected to any CNN classifier to produce highly accurate CAM localizations. Given an original low resolution CAM, foreground and background pixels are randomly sampled to fine-tune the decoder. Additional priors such as image statistics and size constraints are also considered to expand and refine object boundaries.
  Extensive experiments\footnote{{\small Code: \textcolor{red}{ \href{https://github.com/sbelharbi/fcam-wsol}{https://github.com/sbelharbi/fcam-wsol}}}}, over three CNN backbones and six WSOL baselines on the CUB-200-2011 and OpenImages datasets, indicate that our F-CAM method yields a significant improvement in CAM localization accuracy. F-CAM performance is competitive with state-of-art WSOL methods, yet it requires fewer computations during inference.
\end{abstract}


\begin{figure}[ht!]
\centering
  \centering
  \includegraphics[width=.95\linewidth]{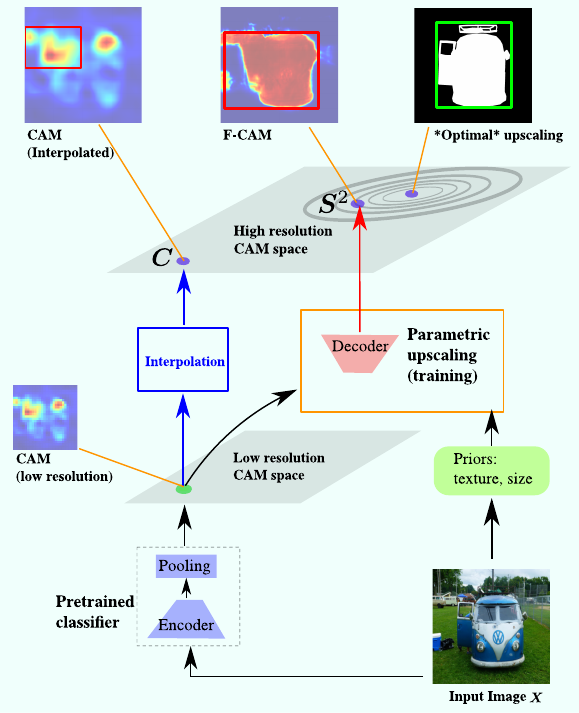}
  \caption{An illustration of the differences between interpolation and our trainable parametric upscaling with priors. ${\bm{C}}$ is the interpolated CAM, and ${\bm{S}^2}$ is the F-CAM produced using our proposed trainable decoder architecture. The figure also shows the elements required to train our method. More details are presented in Fig. \ref{fig:proposal}.}
  \label{fig:intuition}
\end{figure}

\section{Introduction}
\label{sec:intro}

Deep learning (DL) models, and in particular CNNs, provide state-of-the-art performance in many visual recognition applications, such as image classification and object detection. However, they remain complex  models with millions of parameters that typically require supervised end-to-end training on large annotated datasets. Weakly supervised learning (WSL) has recently emerged as an appealing approach to mitigate the cost and burden of annotating large datasets, by exploiting data with limited or coarse labels \cite{zhou2017brief}. In particular, WSL is largely beneficial in object localization to avoid costly annotations, such as bounding boxes. Weakly-supervised object localization (WSOL) methods have drawn much attention because they rely on image-class labels, that is less costly to acquire for training.

An important family of WSOL methods includes class-activation maps (CAMs), which visualize CNN decisions by resorting to the feature-map activations of the deep layers \cite{zhou2016learning}. Such feature maps yield spatial information allowing access to a coarse localization of the object. Despite the growing interest and success of CAM methods, they tend to cover only small discriminative parts of an object.
Several approaches have been proposed to improve CAMs, such as data-enhancement methods ~\cite{ChoeS19,SinghL17, WeiFLCZY17,YunHCOYC19, ZhangWF0H18} or methods that seek to improve the feature maps~\cite{LeeKLLY19,RahimiSAHB20,wei2021shallowspol,WeiXSJFH18,XueLWJJY19iccvdanet,YangKKK20,ZhangCW20,ZhangWKYH18}. Gradients are also used to interrogate a CNN such that specific target labels are localized~\cite{fu2020axiom, lin2013network, PinheiroC15cvpr}.

All these methods are typically applied with an off-the-shelf CNN backbone for feature extraction, such as Inception, VGG, or ResNet families. Given the multiple strided convolutions and pooling operations, these  backbones yield low resolution CAMs with a downscale factor up to 32\footnote{In practice, it is common to modify the convolution stride and max-pooling layers to reduce the downsampling factor.}.
Therefore, each pixel in a CAM covers a patch of ${32\times32}$ pixels in the input image, making the CAM vulnerable to inaccuracies in object localization. Interpolation is often used to generate full size CAMs, but it does not take into consideration statistical properties of an object such as color and texture or its shape. This results in a well-known issue with CAMs, where they cover only small discriminative image regions, leading to bloby localization with inaccurate boundaries (see Figs. \ref{fig:mainvisucub} and \ref{fig:mainvisuopenimages}).  The downscale factor of CAMs can therefore be a bottleneck in localization tasks\footnote{In the supplementary materials, we provide a simulation to show that the downscale factor of a CAM does indeed impose an upper bound performance in localization tasks. We measure the pixel-wise localization with respect to the downscale factor using a simulated CAM.}.
Moreover, CAMs do not explicitly model the background, which plays a central role in increasing false positives/negatives because, e.g., parts of the background could be considered as part of the object~\cite{rony2019weak-loc-histo-survey}.
Finally, using only global labels for supervision (without any pixel-level information), is considered to be an ill-posed problem~\cite{choe2020evaluating,WanWHJY19} that may lead to sub-optimal solutions.

\begin{figure}[ht!]
\centering
  \centering
  \includegraphics[width=\linewidth]{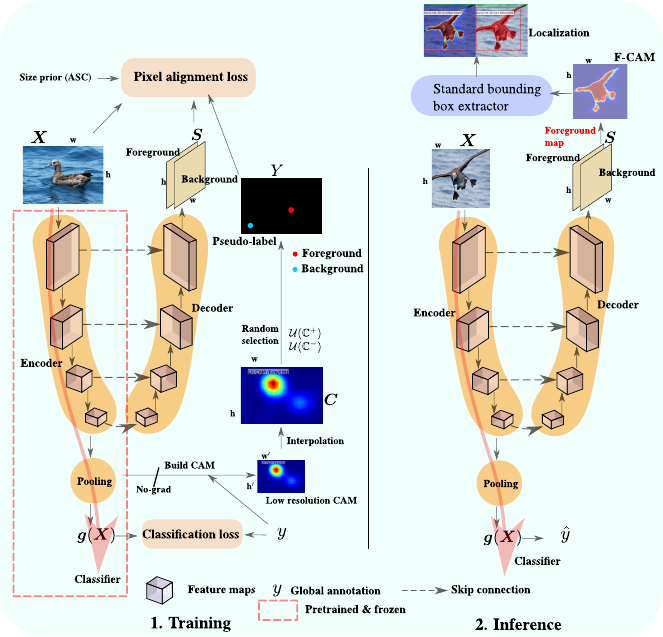}
  \caption{Our proposal: training (left), inference (right).}
  \label{fig:proposal}
\end{figure}

The issue of low resolution CAMs is not sufficiently addressed in the literature~\cite{TagarisSS19,YuKF17,ZhangXWSH20RELI}. For instance, \cite{YuKF17} proposes dilated residual networks that yield a downscale factor of $8$. \cite{ZhangXWSH20RELI} considers upscaling the feature maps then performing classification and segmentation using two branches. \cite{TagarisSS19} uses a U-Net architecture to reconstruct the image. To obtain full resolution CAMs during inference, the reconstruction image is used in combination with the upscaled CAM, in addition to other post-processing methods such as Sobel filtering and region growing methods. Such methods either yield small CAMs, are difficult to scale to large number of classes, or require post-processing steps.
In this paper, we explicitly investigate this issue and propose a method to improve the resolution and localization accuracy of CAMs. As an alternative to interpolation, we propose to equip a classifier employed for WSOL with a \emph{parametric} decoder architecture. The decoder is trained to gradually upscale the resolution of feature maps, yielding a full resolution CAM (Fig.\ref{fig:proposal}). We explicitly model the foreground and background using the decoder, allowing for robust localization. The decoder outputs two activation maps\footnote{This could be easily extended to multi-label case.}, one for the foreground and the other for the background, with the same size as the input image.

Using a decoder such as in Fig. \ref{fig:proposal}, which has a U-Net form~\cite{Ronneberger-unet-2015}, is mainly motivated by deep image prior~\cite{Ulyanov2020deepip}. It has been successfully applied to super-resolution task and other tasks, including denoising and inpainting~\cite{Ulyanov2020deepip}.
It was shown in \cite{Ulyanov2020deepip} that such U-Net architectures with skip connections capture a large part of low-level image statistics. As shown in the recent research in \cite{wei2021shallowspol}, such low-level details play a critical role in improving localization accuracy. The authors of \cite{wei2021shallowspol} aggregate low-level features to yield detailed CAMs that are also used to collect pseudo-labels for pixel-alignment. This suggests that exploiting fine-grained details in deep networks is a promising direction to improve localization accuracy. Fig.\ref{fig:intuition} shows the intuition of our method and its connection to deep image prior~\cite{Ulyanov2020deepip}.
In contrast the transductive learning setup used in \cite{Ulyanov2020deepip} for unsupervised super-resolution, we consider an inductive learning allowing for fast inference. In particular, we propose using local and global constraints to train the decoder \emph{without additional supervision}. The local constraints entail pixel alignment with pseudo-annotation that are collected from CAMs produced by a WSOL classifier. In addition, conditional random field (CRF)~\cite{tang2018regularized} is used as a consistency loss to align the activations with  object boundaries, by exploiting statistical properties of the image such as color and local proximity between pixels. An additional global constraint allows recovering the complete part of the object using a size prior.

\textbf{Our main contributions are summarized below.}

\noindent \textbf{(1)} We propose a simple, yet efficient alternative to interpolation to improve the quality of CAMs for WSOL task. Given a trained CNN classifier for WSOL, we propose integrating a decoder architecture to perform a parametric upscaling of CAMs. It leverages low/top-level features from the classifier, and original low resolution CAMs to produce full-resolution CAMs (F-CAMs) that yield accurate localizations. In addition, the decoder explicitly models the foreground and background. The proposed guidance loss is named \textit{pixel alignment loss}.

\noindent \textbf{(2)} Training of the decoder is performed using loss terms that enhance localization accuracy, consistency, and completeness of F-CAMs, using only global image-class annotations. We exploit low resolution CAMs in addition to image statistics and object size priors to  train the decoder. Our method is generic, and can be combined with any WSOL method with a CNN classifier, such as ResNet, VGG, and Inception families. It aims at improving CAM quality and, therefore, CAM localization accuracy without altering the classification performance.

\noindent \textbf{(3)} Following the experimental WSOL protocol in \cite{choe2020evaluating}, we combine F-CAM with several baseline methods, including CAM~\cite{zhou2016learning}, GradCAM~\cite{SelvarajuCDVPB17iccvgradcam}, GradCam++~\cite{ChattopadhyaySH18wacvgradcampp}, Smooth-GradCAM++~\cite{omeiza2019corr}, XGradCAM~\cite{fu2020axiom}, LayerCAM~\cite{JiangZHCW21layercam}, on two challenging benchmark datasets for WSOL, CUB-200-2011~\cite{WahCUB2002011} and OpenImages~\cite{BenensonPF19,choe2020evaluating}.  The results indicate that F-CAM can provide significant improvements in CAM localization for these baselines. In addition, F-CAMs produced with our method have better properties than standard CAM methods for WSOL, e.g., better robustness to the threshold values. The enhanced results made these simple WSOL baselines competitive with recent state-of-the-art WSOL methods. Finally, we provide ablation studies and a time complexity analysis. The inference time obtained with F-CAM is lower than the average time of fine-tuned baseline methods, and is competitive with other CAM methods.

\section{Related work}
\label{sec:proposal}

Authors in \cite{zhou2016learning} introduce Class Activation Maps (CAMs), showing that spatial feature maps of standard DL models, trained using only image-class labels already, can rich spatial information which can be used for object localization without additional supervision. CAMs allow highlighting important regions of an input image associated with a CNN’s class predictions. Since the CNN is trained for a  classification task using global image-class labels, CAMs tend to activate only on small discriminative regions while missing coverage for large parts of the body.

Several extensions have been proposed to alleviate this issue. In particular, WSOL methods based on data enhancement~\cite{ChoeS19,SinghL17, WeiFLCZY17,YunHCOYC19, ZhangWF0H18} aim to encourage the model to be less dependent on most discriminative regions and seek additional regions. For instance, \cite{SinghL17} divides the input image into patches, and only a few of them are randomly selected during training. This forces the model to look for diverse discriminative regions. However, given this random information suppression in the input image, the CNN can easily confuse objects from the background because most discriminative regions were deleted. This leads to high false positives.

Other methods consider improving the feature maps~\cite{LeeKLLY19,RahimiSAHB20,wei2021shallowspol,WeiXSJFH18,XueLWJJY19iccvdanet,YangKKK20,ZhangCW20,ZhangWKYH18}. For instance, \cite{WeiXSJFH18} considers using dilated convolutions to adapt to objects with different sizes. \cite{ZhangCW20} argues that a WSOL task must be divided into object classification and class-agnostic localization tasks. The latter generates noisy pseudo-annotations, then performs bounding box regression with them for an accurate localization. This is achieved separately from the classification task to avoid undesired interaction between both tasks. In \cite{wei2021shallowspol}, authors enhance the features by considering shallow features of DL models, yielding state-of-the-art localization accuracy, and demonstrating the benefit of shallow features for object localization.

The aforementioned methods are model-dependent, \ie they require training a specific model architecture. Other families of WSOL methods are model-independent, and allow to interrogate the localization of target label over a pretrained classifier, e.g., Gradient-weighted Class Activation Mapping (Grad-CAM), Grad-CAM++, Ablation-CAM, and Axiom-based Grad-CAM \cite{fu2020axiom, lin2013network, PinheiroC15cvpr}. Dilated residual networks (DRN)~\cite{YuKF17} explicitly addressed the issue of low resolution of CAMs. Given an input image of ${224\times224}$, the DRN can produce a map of size ${28\times28}$ which is an improvement compared to ResNet family~\cite{heZRS16} that can produce a map of ${7\times7}$. Despite this improvement, the CAMs are still low resolution with a downscale factor of 8. \cite{TagarisSS19} Uses U-Net model to reconstruct the image. During inference, a series of post-processing steps are used including: applying Sobel filter, and region growing algorithm guided with the upscaled low resolution CAM to yield full resolution CAM. While this provides high resolution CAMs, they still required different post-processing steps.

Finally, authors in \cite{ZhangXWSH20RELI} use a shared backbone, which is followed by two branches for classification over high resolution feature maps and segmentation. CAMs are refined using CRFs, with the collected labels used for segmentation. Our work differs from \cite{ZhangXWSH20RELI} in terms of the architecture and loss, in addition to the main application. We consider U-Net architecture to exploit low and high-level features of the classifier. The classifier's CAMs are low-resolution, enabling scaling to large numbers of classes. We also exploit an unsupervised size constraint, and other methods to collect reliable regions.

\section{Proposed approach}
\label{sec:proposal}

\noindent \textbf{Notation}. Consider a training set ${\mathbb{D} = \{(\bm{X}, y)_i\}_{i=1}^N}$, where ${\bm{X}: \Omega \subset \reals^2}$ denotes an image and ${y \in \{1, \cdots, K\}}$ its global label, with $K$ the number possible classes. Our model (Fig. \ref{fig:proposal}) is composed of: (a) module ${g}$ for the classification task, and (b) decoder ${f}$ to output two activation maps, one for the foreground and the other for background, used for the object localization task. The classifier is composed of: (1) feature encoder backbone for constructing features, and (2) pooling head to compute classification scores. For simplicity, ${\bm{\theta}}$ refers to the parameters of the entire model (Fig. \ref{fig:proposal}).  Furthermore, ${g(\bm{X}) \in [0, 1]^K}$ denotes the per-class classification probabilities where ${g(\bm{X})_k = \mbox{Pr}(k | \bm{X})}$.
The softmax activation maps generated by the decoder are denoted ${\bm{S} = f(\bm{X}) \in [0, 1]^{\abs{\Omega} \times 2}}$ where ${\bm{S}^1, \bm{S}^2}$ are the background and foreground maps, respectively. Map ${\bm{S}^2}$ is class-agnostic, meaning it can hold the activation of any class $y$.
The classifier yields a low-resolution CAM of the target ${y}$, which is then interpolated to have the same resolution as the image. The high-resolution CAM is referred to as ${\bm{C}}$. We denote ${\bm{S}_p \in [0, 1]^2}$ as a row of matrix ${\bm{S}}$, with index ${p \in \Omega}$ denoting a given point.

\noindent \textbf{Generation of sampling regions (SRs)}. In order to guide the fine-tuning of the decoder, we employ local information at pixel level as a supervisory signal. A WSOL task aims at producing a bounding box, with the foreground inside and the remaining region considered as background. Similarly, we rely on the activation magnitude in ${\bm{C}}$ to decide a pixel state (\ie foreground or background). Since such activations hint the presence or absence of an object, we can  assume that pixels with high activations are more likely to be foreground, while lower activations are background~\cite{durand2017wildcat}. We denote ${\mathbb{C}^+}$ and ${\mathbb{C}^-}$ as foreground and background regions, respectively, estimated as follows,
\begin{equation}
    \label{eq:sets}
    \mathbb{C}^+ = \psi^+(\bm{C}), \quad \mathbb{C}^- = \psi^-(\bm{C}, n^-) \;,
\end{equation}
where ${\psi^+(\bm{C})}$ is the set of top\footnote{In this context, the top $n$ elements of a list ordered from value $a$ toward $b$ are the initial $n$ elements of the list.} pixels in ${\bm{C}}$, which is ordered from high to low activation. Without adding additional hyper-parameters, ${\psi^+(\bm{C})}$ takes all pixels in ${\bm{C}}$ with activation magnitude above Otsu~\cite{otsuthresh} threshold obtained over ${\bm{C}}$. ${\psi^-(\bm{C}, n^-)}$ is the set of top ${n^-\%}$ pixels in ${\bm{C}}$, which is ordered from low to high activation.  ${\psi^-(\bm{C}, n^-)}$ sorts activation magnitudes in ${\bm{C}}$ from low to high and takes the ${n^-\%}$ of top pixels. We denote by ${n^-\%}$ the portion of pixels that we are allowed to consider as background. Pixels in ${\mathbb{C}^+}$ are assigned pseudo-label ${1}$ for foreground and pixels in ${\mathbb{C}^-}$ are assigned $0$ for background.

The estimated sampling regions ${\mathbb{C}^+}$ and ${\mathbb{C}^-}$ are uncertain and can contain incorrect labels. Since standard CAMs are often bloby, the foreground ${\mathbb{C}^+}$ could contain background regions. Similarly, ${\mathbb{C}^-}$  is expected to hold background pixels but also parts of the object since CAMs are typically incomplete. Due to this uncertainty and noise in labels, we avoid fitting the model directly on ${\mathbb{C}^+/ \mathbb{C}^-}$ all at once. Instead, we randomly select a few pixels at each training iteration, while dropping the remaining pixels~\cite{SinghL17,srivastava2014dropout}. This prevents overfitting over ${\mathbb{C}^+}$ and ${\mathbb{C}^-}$. To this end, we define a stochastic set of pixels randomly selected from foreground and background for an image at a training iteration,
\begin{equation}
    \label{eq:sset}
    \Omega^{\prime} = \mathcal{U}(\mathbb{C}^+)\; \cup \; \mathcal{U}(\mathbb{C}^-) \;,
\end{equation}
where ${\mathcal{U}(\mathbb{C})}$ consists of a sampled set of one pixel\footnote{We can however sample more pixels at once.} uniformly sampled from the set ${\mathbb{C}}$.  We denote by ${Y}$ the \emph{partially} pseudo-labeled mask for the sample ${\bm{X}}$, where ${Y_p \in \{0, 1\}^2}$ with labels ${0}$ for background, and ${1}$ for foreground.

\noindent \textbf{Overall training loss}. Our training loss combines two main terms: a standard classification loss, and our proposed loss for fine-tuning the decoder named \emph{pixel alignment loss}. This loss entails local (\ie, pixel level) and global terms. The local term aims at aligning the output activations ${\bm{S}}$ with the pseudo-labeled pixels selected in ${\Omega^{\prime}}$ using standard partial cross-entropy ${\bm{H}}$.  To promote the consistency in activations, and align them with the object boundaries, we exploit statistical properties of the image such as the color, and pixel proximity allowing nearby pixels with similar color to be assigned similar state (\ie, foreground or background). To this end, we include the CRF loss~\cite{tang2018regularized} denoted by ${\mathcal{R}}$. (See supplementary materials for more details.)

CAMs are known to highlight minimal discriminative regions. This could easily lead to unbalanced partitioning, where the background region dominates the foreground in term of size, which gives raise to false positives. To circumvent this issue, the activations of the foreground are explicitly pushed to expand by constraining their total area to be large. In parallel, the background is pushed to be large as well, so as to avoid foreground dominance. By doing so, we push both regions to compete over pixels but without violating other competing losses, \ie CRF and portial cross entropy. In particular, we consider the absolute size constraint (ASC)~\cite{belharbi2020minmaxuncer} over both the foreground and background. We do not assume whether the background is larger than the foreground~\cite{pathak2015constrained} nor the opposite. The ASC loss encourages both regions to be large. To avoid a trivial solution, with half the image as foreground and the other half as background, control terms are necessary. In this work, partial cross entropy and CRF losses control the growth of the size, so as to ensure consistency with the object boundaries and sampling regions. The size loss is {\em unsupervised} and formulated through inequality constraints to maximize the area of activation of the map at hand. The constraints are solved via standard log-barrier method~\cite{boyd2004convex}.
We note: ${\bm{H}(Y_p, \bm{S}_p) = - \sum_{t}^2 Y_p^t\; \log(\bm{S}_p^t)}$ as the standard cross-entropy between the ${S_p}$ and the pseudo-label mask ${Y_p}$ at pixel ${p}$, and ${\alpha, \lambda}$ are balancing coefficients. Our overall pixel alignment loss is formulated as,
\begin{equation}
\label{eq:totalloss}
\begin{aligned}
\min_{\bm{\theta}} \quad & - \log(\bm{g}(\bm{X})[y]) + \alpha \sum_{p \in \Omega^{\prime}} \bm{H}(Y_p, \bm{S}_p) + \lambda\; \mathcal{R}(\bm{S}, \bm{X}) \;,\\
\textrm{s.t.} \quad & \sum \bm{S}^r \geq 0 \;, \quad r \in \{1, 2\} \;,\\
\end{aligned}
\end{equation}

It is important to note that training with our method (Eq. \ref{eq:totalloss}) does not require any additional supervision besides the already provided global image-class annotation ${y}$. This label is used to build the CAM of the target as presented in Fig. \ref{fig:proposal}. Another important aspect is the semantic meaning of the foreground in ${\bm{S}^2}$. Since ${\mathbb{C}^+}$ holds pixels that are assumed to be foreground estimated from the CAM ${\bm{C}}$ of \emph{the true label} ${y}$, the foreground predicted in ${\bm{S}^2}$ is expected to be consistent with the global annotation $y$ of the image. Once the foreground full-resolution CAM is obtained, localization is carried out using the same standard method used for any CAM (see Fig. \ref{fig:proposal}, \emph{inference}).

\section{Results and discussion}
\label{sec:experiments}

\subsection{Experimental methodology:}
\label{subsec:expsettings}

\noindent \textbf{Datasets.} To evaluate our method, two datasets from \cite{choe2020evaluating} are adopted: CUB-200-2011 (CUB) \cite{WahCUB2002011} and OpenImages \cite{BenensonPF19,choe2020evaluating}. CUB contains 200 categories of birds with 5,994 training images and 5,794 testing images. In addition, 1000 extra images annotated in \cite{choe2020evaluating} are used as a validation set for model and hyper-parameters selection. OpenImages contains 37,319 images of 100 classes. 29,819 samples are used for training, while 2,500 samples are used for validation. The 5,000 remaining images are used for test. Different from CUB, OpenImages WSOL dataset provides pixel annotation of objects instead of bounding boxes for a fine localization. We follow the protocol in \cite{choe2020evaluating} for both datasets.

\noindent \textbf{Evaluation metrics.} Following \cite{choe2020evaluating}, we report 5 localization metrics and one classification metric. For localization, we report: (1) \maxboxacc (also known as CorLoc \cite{deselaers2012weakly}, and GT-known \cite{SinghL17}): fraction of images for which the predicted bounding box has more than $\sigma = 50\%$ IoU with the ground truth,  independently from classification prediction, (2) \newmaxboxacc: the same as \maxboxacc but averaged over threes sizes ${\sigma \in \{30\%, 50\%, 70\%\}}$, (3) \topone localization accuracy: fraction of images with the correct class prediction and more than ${\sigma = 50\%}$ IoU with the ground truth box, and (4) \topfive localization accuracy: fraction of images with class label belonging to the \topfive predictions and more than ${\sigma=50\%}$ IoU.  With OpenImages, (5) we report the \pxap metric proposed in \cite{choe2020evaluating} which computes the area under the precision-recall curve. As in \cite{choe2020evaluating}, the CAM's threshold is marginalized over the interval ${\tau \in [0, 1]}$ with a step of $0.001$.

\noindent \textbf{Implementation details.} In all experiments, we follow the same protocol as  \cite{choe2020evaluating} including backbones, training epochs (50 for CUB and 10 for OpenImages), and batch size of 32. We validated our method over three backbones, VGG16 \cite{SimonyanZ14a}, InceptionV3 \cite{SzegedyVISW16}, and ResNet50 \cite{heZRS16}. In Eq.\ref{eq:totalloss}, the hyper-parameter $\lambda$ for the CRF is set to the same value as in \cite{tang2018regularized} which is ${2e^{-9}}$. For log-barrier optimization, hyper-parameter $t$ is set to the same value as in \cite{belharbi2019unimoconstraints,kervadec2019log}. It is initialized to $1$, and increased by a factor of ${1.01}$ in each epoch with a maximum value of $10$. $\alpha$ is searched in ${\{1, 0.1\}}$ through validation. We find that large values, such as 1, do not harm the performance, suggesting that sampling regions are less noisy. In all experiments with our method, we used a learning rate of $0.01$ using SGD for optimization. Similar to \cite{choe2020evaluating}, images are resized to ${256\times256}$, then randomly cropped to ${224\times224}$ for training. ${n^-}$ is chosen using the validation set from the set ${[0.1, 0.7]}$ with a step of 0.1.

\noindent \textbf{Baseline models.}
To validate our F-CAM method, we compare with recent WSOL methods, including: CAM~\cite{zhou2016learning}, HaS~\cite{SinghL17}, ACoL~\cite{ZhangWF0H18}, SPG~\cite{ZhangWKYH18}, ADL~\cite{ChoeS19}, CutMix~\cite{YunHCOYC19}, CSTN~\cite{MeethalPBG20icprcstn}, TS-CAM~\cite{gao2021tscam}, MEIL~\cite{Mai20CVPRmeil}, DANet~\cite{XueLWJJY19iccvdanet}, SPOL~\cite{wei2021shallowspol}, ICL~\cite{KiU0B20icl}, NL-CCAM~\cite{YangKKK20nlccam}, I${^2}$C~\cite{ZhangW020i2c}, GradCAM~\cite{SelvarajuCDVPB17iccvgradcam}, GradCam++~\cite{ChattopadhyaySH18wacvgradcampp}, Smooth-GradCAM++~\cite{omeiza2019corr}, XGradCAM~\cite{fu2020axiom}, LayerCAM~\cite{JiangZHCW21layercam}. We present the results reported in \cite{choe2020evaluating} for the methods: CAM, HaS, ACoL, SPG, ADL, and CutMix. For GradCAM, GradCam++, Smooth-GradCAM, XGradcam, and LayerCAM, we have reproduced their results. We also reproduced results for CAM~\cite{zhou2016learning}, and name its results as CAM* to distinguish it from CAM's results in \cite{choe2020evaluating}. For the rest of the methods, we present what was reported in the original papers. Missing values are shown here by ${--}$. \cite{choe2020evaluating} provides results using few-shot learning (FSL) where a few fully supervised samples are used to train the model. A simple baseline is also provided in \cite{choe2020evaluating} which is a center-Gaussian baseline. It generates isotropic Gaussian score maps centered at the image. This represents a lower bound performance obtained without any training.
We study the impact of combining our method with 6 baseline WSOL methods over localization performance. Our choice is based on low complexity -- we chose methods that yield CAMs by a simple forward pass, such as CAM~\cite{zhou2016learning}, or a forward and a backward pass, such as GradCAM family while using standard pretrained classifiers. In addition, these methods simply interrogate a classifier without changing its architecture, making the integration of our decoder with the classifier straightforward. To this end, we select the following WSOL methods: CAM, GradCAM, GradCAM++, Smoth-GradCAM, XGradCAM, and LayerCAM. All the baselines use the same pooling method which is a global average pooling~\cite{zhou2016learning}.
To integrate our F-CAM method, the WSOL baseline method is trained only using the classification term in Eq.\ref{eq:totalloss} until convergence. Then, we freeze the classifier, and continue fine-tuning our decoder using the pixel alignment loss in Eq.\ref{eq:totalloss}. Such separation in training is meant to avoid any undesirable interaction between classification and pixel-wise assignment tasks \cite{belharbi2020DeepAlJoinClSegWeakAnn}, and to provide clear conclusions about the results. Moreover, it allows the baseline method to converge and yield accurate localization which will be used to guide the decoder's fine-tuning.

\subsection{Comparison with state-of-the-art:}
\label{subsec:sota}

{
\setlength{\tabcolsep}{3pt}
\renewcommand{\arraystretch}{1.1}
\begin{table}[ht!]
\centering
\resizebox{.47\textwidth}{!}{%
\centering
\small
\begin{tabular}{lc*{3}{c}gc*{3}{c}g}
& &  \multicolumn{4}{c}{CUB (\maxboxacc)}  & & \multicolumn{4}{c}{OpenImages (\pxap)} \\
Methods   &  & VGG & Inception & ResNet & Mean &  &  VGG &  Inception & ResNet & Mean  \\
\cline{1-1}\cline{3-6}\cline{8-11} \\
CAM~\cite{zhou2016learning} {\small \emph{(cvpr,2016)}} &  & 71.1 & 62.1 & 73.2 & 68.8 &  & 58.1 & 61.4 & 58.0 & 59.1 \\
HaS~\cite{SinghL17} {\small \emph{(iccv,2017)}} &  &  76.3 & 57.7 & 78.1 & 70.7 &  & 56.9 & 59.5 & 58.2 & 57.8 \\
ACoL~\cite{ZhangWF0H18} {\small \emph{(cvpr,2018)}} &  &  72.3 & 59.6 & 72.7 & 68.2 &  & 54.7 & 63.0 & 57.8 & 58.4 \\
SPG~\cite{ZhangWKYH18} {\small \emph{(eccv,2018)}} &  &  63.7 & 62.8 & 71.4 & 66.0 &  & 55.9 & 62.4 & 57.7 & 58.6 \\
ADL~\cite{ChoeS19} {\small \emph{(cvpr,2019)}} &  &  75.7 & 63.4 & 73.5 & 70.8 &  & 58.3 & 62.1 & 54.3 & 58.2 \\
CutMix~\cite{YunHCOYC19} {\small \emph{(eccv,2019)}} &  &  71.9 & 65.5 & 67.8 & 68.4 &  & 58.2 & 61.7 & 58.7 & 59.5 \\
\cline{1-1}\cline{3-6}\cline{8-11} \\
Best WSOL &  & 76.3 & 65.5 & 78.1 & 70.8 &  & 58.3 & 63.0 & 58.7 & 59.5 \\
FSL baseline &  & 86.3 & 94.0 & 95.8 & 92.0 &  & 61.5 & 70.3 & 74.4 & 68.7 \\
Center baseline &  &  59.7 & 59.7 & 59.7 & 59.7 &  & 45.8 & 45.8 & 45.8 & 45.8 \\
\cline{1-1}\cline{3-6}\cline{8-11} \\
CSTN~\cite{MeethalPBG20icprcstn} {\small \emph{(icpr,2020)}} &  &  \multicolumn{4}{c}{Resnet101~\cite{heZRS16}: 76.0} &  & -- & -- & -- & -- \\
TS-CAM~\cite{gao2021tscam} {\small \emph{(corr,2021)}} &  &  \multicolumn{4}{c}{Deit-S~\cite{TouvronCDMSJ21}: 83.8} &  & -- & -- & -- & -- \\
MEIL~\cite{Mai20CVPRmeil} {\small \emph{(cvpr,2020)}} &  &  73.8 & -- & -- & -- &  & -- & -- & -- & -- \\
DANet~\cite{XueLWJJY19iccvdanet} {\small \emph{(iccv,2019)}} &  &  67.7 & 67.03 & -- & -- &  & -- & -- & -- & -- \\
SPOL~\cite{wei2021shallowspol} {\small \emph{(cvpr,2021)}} &  &  -- & -- & 96.4 & -- &  & -- & -- & -- & -- \\
\cline{1-1}\cline{3-6}\cline{8-11} \\
CAM*~\cite{zhou2016learning} {\small \emph{(cvpr,2016)}} &  &    61.6 & 58.8 & 71.5 & 63.9 &  & 53.0 & 62.7 & 56.8 & 57.5 \\
GradCAM~\cite{SelvarajuCDVPB17iccvgradcam} {\small \emph{(iccv,2017)}} &  &                        69.3 & 62.3 & 73.1 & 68.2 &  & 59.6 & 63.9 & 60.1 & 61.2 \\
GradCAM++~\cite{ChattopadhyaySH18wacvgradcampp} {\small \emph{(wacv,2018)}}&  &                       84.1 & 63.3 & 81.9 & 76.4 &  & 60.5 & 64.0 & 60.2 & 61.5 \\
Smooth-GradCAM++~\cite{omeiza2019corr} {\small \emph{(corr,2019)}} &  &               69.7 & 66.9 & 76.3 & 70.9 &  & 52.2 & 61.7 & 54.3 & 56.0 \\
XGradCAM~\cite{fu2020axiom} {\small \emph{(bmvc,2020)}} &  &                        69.3 & 60.9 & 72.7 & 67.6 &  & 59.0 & 63.9 & 60.2 & 61.0 \\
LayerCAM~\cite{JiangZHCW21layercam} {\small \emph{(ieee,2021)}} &  &                        84.3 & 66.5 & 85.2 & 78.6 &  & 59.5 & 63.5 & 61.1 & 61.3 \\
\cline{1-1}\cline{3-6}\cline{8-11}\\
\cline{1-1}\cline{3-6}\cline{8-11}\\
CAM*~\cite{zhou2016learning} + ours &  &                   87.3 & 82.0 & 90.3 & 86.5 &  & 67.8 & 71.9 & 72.1 & 70.6 \\
GradCAM~\cite{SelvarajuCDVPB17iccvgradcam} + ours &  &                87.5 & 84.4 & 90.5 & 87.4 &  & 68.6 & 70.0 & 70.9 & 69.8 \\
GradCAM++~\cite{zhou2016learning} + ours       &  &        91.5 & 84.6 & 91.0 & 89.0 &  & 64.8 & 67.1 & 66.3 & 66.0 \\
Smooth-GradCAM++~\cite{zhou2016learning} + ours       &  & 89.1 & 86.8 & 90.7 & 88.8 &  & 60.3 & 65.4 & 64.4 & 63.3 \\
XGradCAM~\cite{zhou2016learning} + ours       &  &         86.8 & 84.4 & 90.4 & 88.8 &  & 68.7 & 71.3 & 70.4 & 70.1 \\
LayerCAM~\cite{zhou2016learning}    + ours   &  &         91.0 & 85.3 & 92.4 & 89.7 &  & 64.3 & 64.9 & 65.3 & 64.8 \\
\cline{1-1}\cline{3-6}\cline{8-11} \\
Best WSOL + ours &  &                                      91.5 & 86.8 & 92.4 & 89.7 &  & 68.7 & 71.9 & 72.1 & 70.6 \\
\cline{1-1}\cline{3-6}\cline{8-11}\\
\end{tabular}
}
\caption{Performance on \maxboxacc and \pxap metrics.}
\label{tab:maxbox-pxap}
\vspace{-1em}
\end{table}
}

{
\setlength{\tabcolsep}{3pt}
\renewcommand{\arraystretch}{1.1}
\begin{table}[ht!]
\centering
\resizebox{.47\textwidth}{!}{
\centering
\small
\begin{tabular}{lc*{3}{c}c*{3}{c}}
& &  \multicolumn{3}{c}{\topone localization}  & & \multicolumn{3}{c}{\topfive localization}\\
Methods  &  &  VGG & Inception & ResNet  &  & VGG & Inception & ResNet \\
\cline{1-1}\cline{3-5}\cline{7-9} \\
CAM~\cite{zhou2016learning} {\small \emph{(cvpr,2016)}}&  &     45.8 & 40.4 & 56.1  &  & -- & -- & --  \\
HaS~\cite{SinghL17} {\small \emph{(iccv,2017)}}&  &             55.6 & 41.1 & 60.7  &  & -- & -- & --  \\
ACoL~\cite{ZhangWF0H18} {\small \emph{(cvpr,2018)}}&  &         44.8 & 46.8 & 57.8  &  & -- & -- & --  \\
SPG~\cite{ZhangWKYH18} {\small \emph{(eccv,2018)}}&  &          42.9 & 44.9 & 51.5  &  & -- & -- & --  \\
ADL~\cite{ChoeS19} {\small \emph{(cvpr,2019)}}&  &              39.2 & 35.2 & 41.1  &  & -- & -- & --  \\
CutMix~\cite{YunHCOYC19} {\small \emph{(eccv,2019)}}&  &        47.0 & 48.3 & 54.5  &  & -- & -- & --  \\
\cline{1-1}\cline{3-5}\cline{7-9}\\
ICL~\cite{KiU0B20icl} {\small \emph{(accv,2020)}}&  &           57.5 & 56.1 & 56.1  &  & -- & -- & --  \\
CSTN~\cite{MeethalPBG20icprcstn} {\small \emph{(icpr,2020)}}&  &           \multicolumn{3}{c}{Resnet101~\cite{heZRS16}: 49.0} &  & -- & -- & --  \\
TS-CAM~\cite{gao2021tscam} {\small \emph{(corr,2021)}}&  &           \multicolumn{3}{c}{Deit-S~\cite{TouvronCDMSJ21}: 71.3} &  & \multicolumn{3}{c}{Deit-S~\cite{TouvronCDMSJ21}: 83.8}  \\
I$^2$C~\cite{ZhangW020i2c} {\small \emph{(eccv,2020)}}&  &           -- & 56.0 & --  &  & -- & 68.3 & --  \\
MEIL~\cite{Mai20CVPRmeil} {\small \emph{(cvpr,2020)}}&  &           57.4 & -- & --  &  & -- & -- & --  \\
DANet~\cite{XueLWJJY19iccvdanet} {\small \emph{(iccv,2019)}}&  &         52.5   & 49.4 & --  &  & 61.9 & 60.4 & --  \\
NL-CCAM~\cite{YangKKK20nlccam} {\small \emph{(wacv,2020)}}&  &         52.4   & -- & --  &  & 65.0 & -- & --  \\
SPOL~\cite{wei2021shallowspol} {\small \emph{(cvpr,2021)}}&  &         --   & -- & 80.1  &  & -- & -- & 93.4  \\
\cline{1-1}\cline{3-5}\cline{7-9}\\
\cline{1-1}\cline{3-5}\cline{7-9}\\
CAM*~\cite{zhou2016learning} {\small \emph{(cvpr,2016)}} &  &    33.5 & 40.9 & 47.8  &  & 52.7 & 54.8 & 66.5  \\
GradCAM~\cite{SelvarajuCDVPB17iccvgradcam} {\small \emph{(iccv,2017)}}  &  &                        18.5 & 41.6 & 32.3  &  & 41.7 & 56.2 & 56.4  \\
GradCAM++~\cite{ChattopadhyaySH18wacvgradcampp} {\small \emph{(wacv,2018)}} &  &                       20.8 & 41.4 & 34.7  &  & 47.5 & 56.8 & 61.4  \\
Smooth-GradCAM~\cite{omeiza2019corr} {\small \emph{(corr,2019)}}  &  &                 17.7 & 44.4 & 33.1  &  & 40.2 & 60.2 & 57.7 \\
XGradCAM~\cite{fu2020axiom} {\small \emph{(bmvc,2020)}} &  &                        18.5 & 40.9 & 48.3  &  & 41.7 & 55.2 & 67.3  \\
LayerCAM~\cite{JiangZHCW21layercam} {\small \emph{(ieee,2021)}}  &  &                       \red{24.8} & 43.9 & 44.2  &  & 47.8 & 59.7 & 70.3 \\
\cline{1-1}\cline{3-5}\cline{7-9}\\
\cline{1-1}\cline{3-5}\cline{7-9}\\
CAM*~\cite{zhou2016learning} + ours &  &    44.0 & 54.6 & 59.1  &  & 70.1 & 73.8 & 82.6  \\
GradCAM~\cite{SelvarajuCDVPB17iccvgradcam} + ours &  &                         22.1 & 56.1 & 38.8  &  & 50.0 & 75.4 & 68.5  \\
GradCAM++~\cite{ChattopadhyaySH18wacvgradcampp} + ours &  &                       23.2 & 55.9 & 39.0  &  & 52.0 & 76.0 & 68.9  \\
Smooth-GradCAM~\cite{omeiza2019corr}  + ours &  &                  22.9 & 57.3 & 38.9  &  & 51.2 & 77.6 & 68.6 \\
XGradCAM~\cite{fu2020axiom} + ours &  &                        22.0 & 55.8 & 59.3  &  & 49.6 & 75.3 & 82.7  \\
LayerCAM~\cite{JiangZHCW21layercam} + ours &  &                        \red{23.1} & 56.4 & 47.7  &  & 51.9 & 76.6 & 76.1 \\
\cline{1-1}\cline{3-5}\cline{7-9}\\
\end{tabular}
}
\caption{The \topone and \topfive localization accuracy of WSOL methods on CUB according to the \maxboxacc metric. In \red{red} are the cases where our method decreases performance over the corresponding baseline.}
\label{tab:top15}
\end{table}
}

\noindent \textbf{Quantitative comparison}. Tab.\ref{tab:maxbox-pxap} shows the performance obtained with the proposed and baseline methods according to the \maxboxacc and \pxap metrics. We observe that results with the 6 selected baselines range from the lower performance using CAM* to the higher performance using LayerCAM. This provides a good scenario to evaluate our method when combined with weak and strong baselines. Note that compared to the methods reported in \cite{choe2020evaluating}, GradCAM family reported much higher performance. Combing our method with each of these baselines yields a considerable improvement for all CAM methods, CNN backbones, and over both datasets. For instance, CAM* alone yields a \maxboxacc of 71.5\% over CUB using ResNet50. When combined with our method, its performance climbs to 90.3\%. The same improvement is observed over OpenImages but with a smaller margin. Note that OpenImages is more challenging than CUB as it has a high variation among classes of different objects (CUB contains only birds' species). Results indicate that by simply combing a decoder with these baselines, we can increase their performance to a level that is competitive with recent state-of-the-art methods, such as TS-CAM and CSTN and even approaching the performance of SPOL.  In addition, these results are competitive with FSL, and in some cases, surpassing FSL performance. Similar behavior is observed over \newmaxboxacc performance\footnote{\newmaxboxacc performance on CUB is reported in supp. material.}.

\begin{figure}
     \centering
     \begin{subfigure}[b]{0.45\textwidth}
         \centering
         \includegraphics[width=\textwidth]{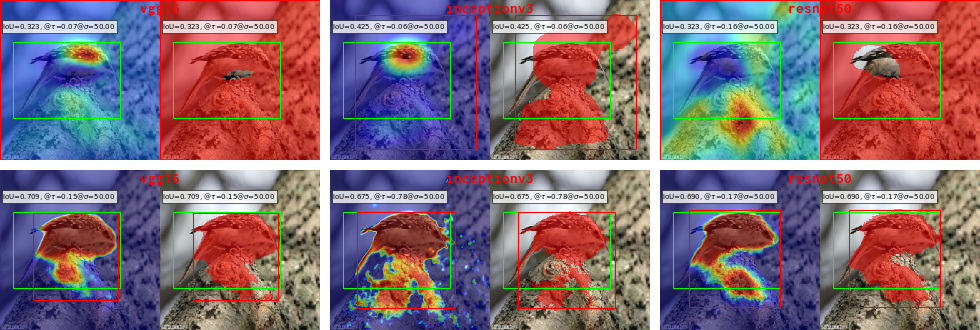}
     \end{subfigure}
     \\
     \vspace{.2cm}
     \begin{subfigure}[b]{0.45\textwidth}
         \centering
         \includegraphics[width=\textwidth]{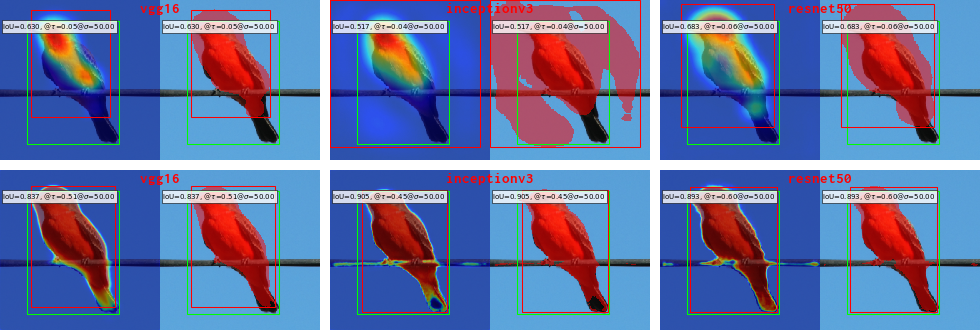}
     \end{subfigure}
     \\
     \vspace{.2cm}
     \begin{subfigure}[b]{0.45\textwidth}
         \centering
         \includegraphics[width=\textwidth]{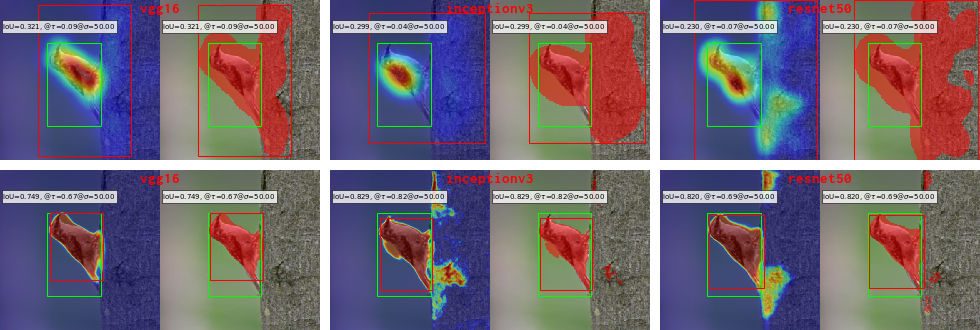}
     \end{subfigure}
        \caption{Test samples from CUB. Top: CAM*. Middle: GradCAM. Bottom: GradCAM++. First row: WSOL baseline. Next row: WSOL baseline + ours. First column: CAM. Next column: localization.
        Colors: predicted boxes in red, and true box in green. Thresholded mask is in red. ${\sigma= 50\%}$.}
        \label{fig:mainvisucub}
\end{figure}

\begin{figure}
     \centering
     \begin{subfigure}[b]{0.45\textwidth}
         \centering
         \includegraphics[width=\textwidth]{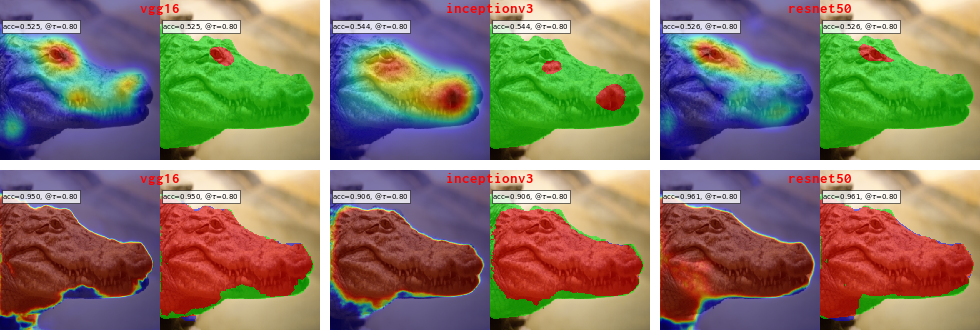}
     \end{subfigure}
     \\
     \vspace{.2cm}
     \begin{subfigure}[b]{0.45\textwidth}
         \centering
         \includegraphics[width=\textwidth]{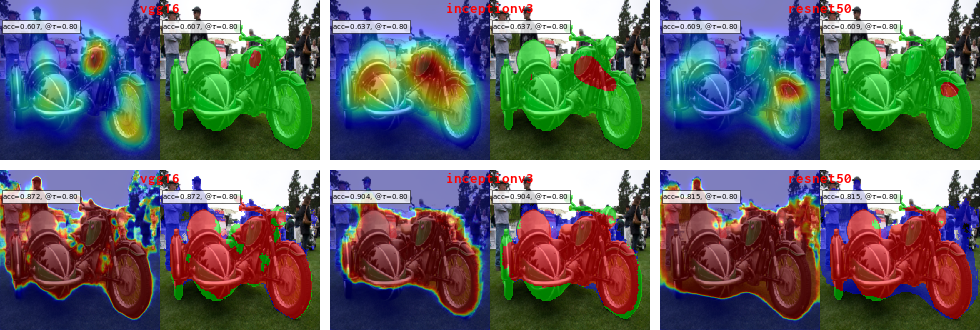}
     \end{subfigure}
     \\
     \vspace{.2cm}
     \begin{subfigure}[b]{0.45\textwidth}
         \centering
         \includegraphics[width=\textwidth]{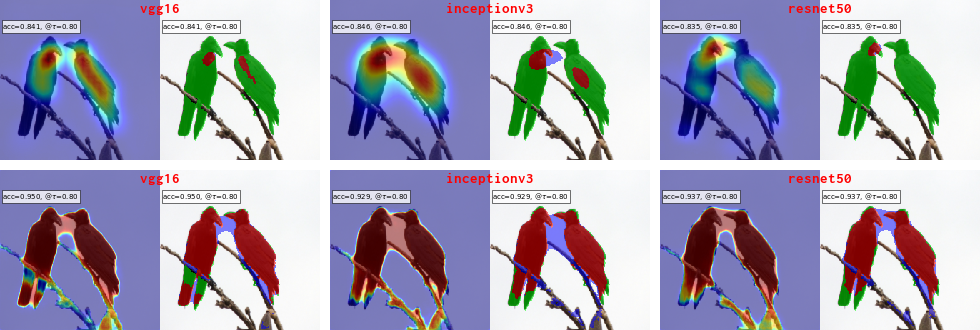}
     \end{subfigure}
        \caption{Test samples from OpenImages. Top: LayerCAM. Middle: Smooth-GradCAM. Bottom: XGradCAM. First row: WSOL baseline. Next row: WSOL baseline + ours. First column: CAM. Next column: localization. Colors: green false negative, red : true positive, blue: false positive. ${\tau=0.8}$.}
        \label{fig:mainvisuopenimages}
\end{figure}

Tab.\ref{tab:top15} shows \topone and \topfive localization performance over CUB. When using our method, WSOL baselines obtained competitive \topfive localization over Inception and ResNet architectures. However, \topone localization is poor even with the assistance of our method. Since \topone localization is directly tied to classification performance, the improvement of our method is bounded by the number of correctly classified samples. By inspecting the classification performance of WSOL baselines, only CAM* yields high accuracy over the CUB dataset for the three architectures. This explains why \topone CAM* is much higher than others. The average classification accuracy ranges from 44\% to 59\%. In addition, all WSOL baselines yield high classification using VGG16 over CUB which again explains the high \topone localization. Over OpenImages, all methods yield relatively the same classification accuracy with an average that ranges from 63\% to 70\%.
Since the aim of our method is to improve localization performance without changing the classification performance, our method can only improve the localization of correctly classified samples. This gives advantage to other methods that have high classification accuracy. Note that model selection during the training of our baselines is achieved through the localization performance on the validation set as proposed in \cite{choe2020evaluating}. It was observed in \cite{choe2020evaluating} that the localization task converges in the early training epochs while the classification task takes longer to converge. We observed similar behavior when training our baselines\footnote{Training curves and classification performance for all WSOL baselines methods are reported in the supplementary materials.}.

\noindent \textbf{Visual comparison: WSOL baselines with F-CAMs}. Figs.\ref{fig:mainvisucub} and \ref{fig:mainvisuopenimages} illustrate the impact of our method on WSOL baselines in term of activations and localization. It is well known that standard CAM methods tend to activate only over minimal discriminative regions as shown in these figures. However, the optimum threshold allows other low (invisible) activations to participate in defining the final localization such as in the first row in Fig.\ref{fig:mainvisucub}.  As a result, non-discriminative regions could be easily included in the localization. Moreover, such low activations downgrade the interpretability aspect since they yield a box that covers regions without any visible activation over an object.
In comparison, our method yields sharp and complete CAMs allowing bounding boxes to be tight around the object making it robust to thresholding while improving the interpretability aspect.
Similarly, over OpenImages, WSOL baselines yield minimal activation that are sensitive to thresholding. Note that in \pxap metric, authors in \cite{choe2020evaluating} do not define an optimum threshold as in \maxboxacc and \newmaxboxacc since \pxap defines an area under the precision-recall curve. Using high threshold such as $0.8$ yields very small true positive regions with high false negative. On the other hand, our method easily covers the object with minimal false positives.
In the next section, we analyze the score of the CAMs as an attempt to understand their sensitivity to thresholding.
\begin{figure}
     \centering
     \begin{subfigure}[b]{0.45\textwidth}
         \centering
         \includegraphics[width=\textwidth]{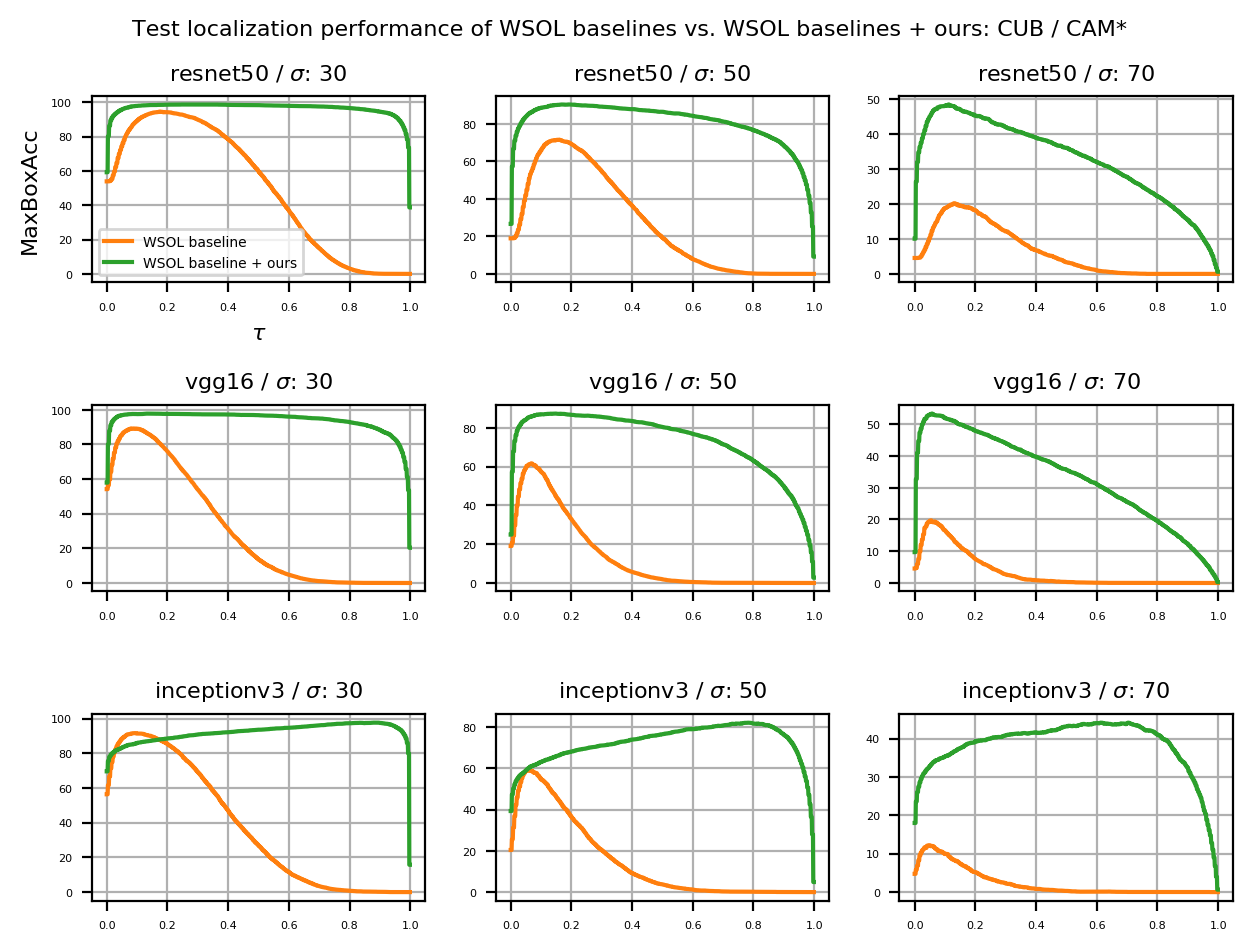}
          \caption{\maxboxacc metric.}
         \label{fig:cam-maxbo-tau}
     \end{subfigure}
     \\
     \vspace{.2cm}
     \begin{subfigure}[b]{0.45\textwidth}
         \centering
         \includegraphics[width=\textwidth]{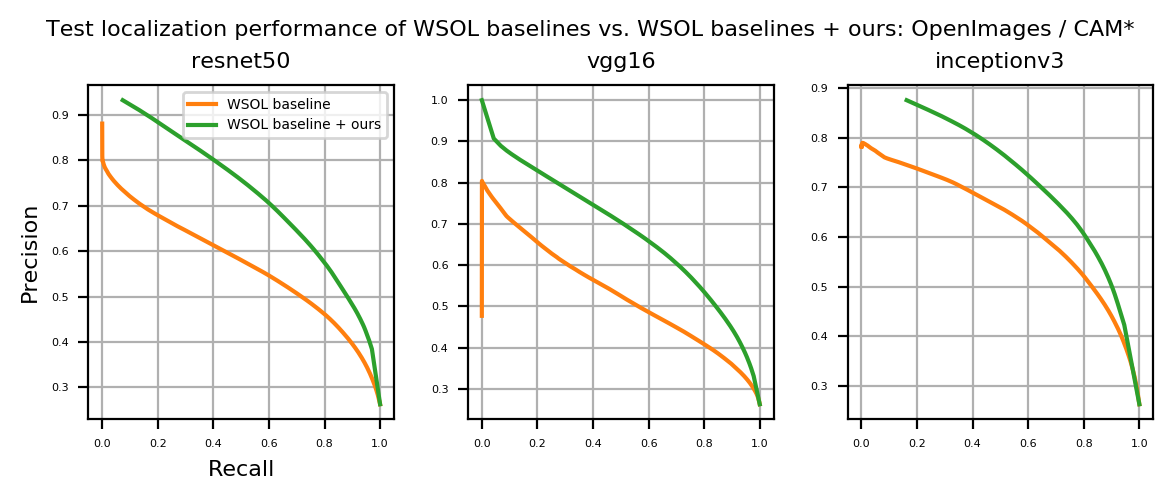}
          \caption{\pxap metric.}
         \label{fig:cam-pxap-tau}
     \end{subfigure}
        \caption{Localization accuracy as a function of threshold value ${\tau}$ for CAM* (orange) and  CAM* + ours (green). (A) CUB. (b) OpenImages.}
        \label{fig:cam-tau}
\end{figure}
\begin{figure}
     \centering
         \includegraphics[width=\linewidth]{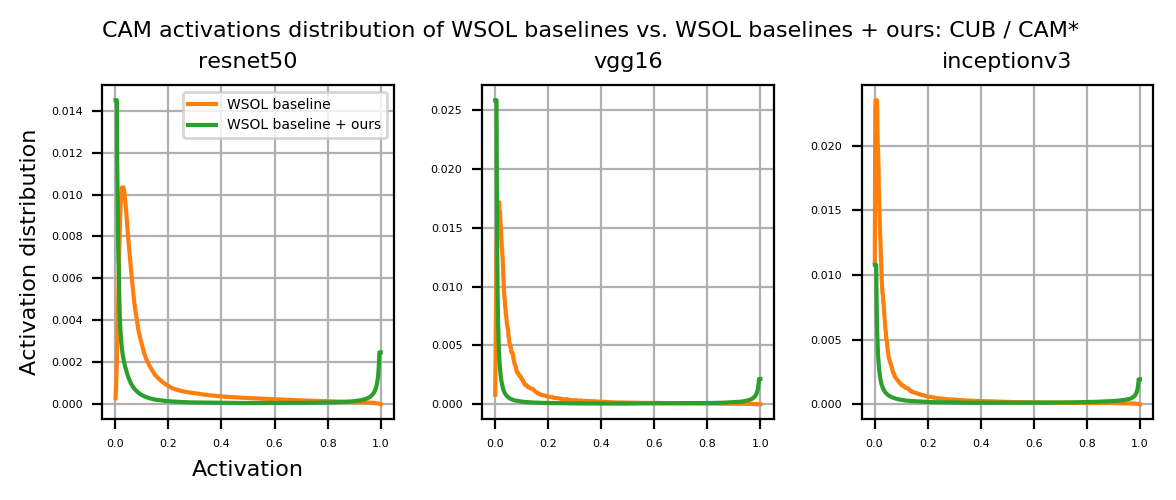}
            \\
         \includegraphics[width=\linewidth]{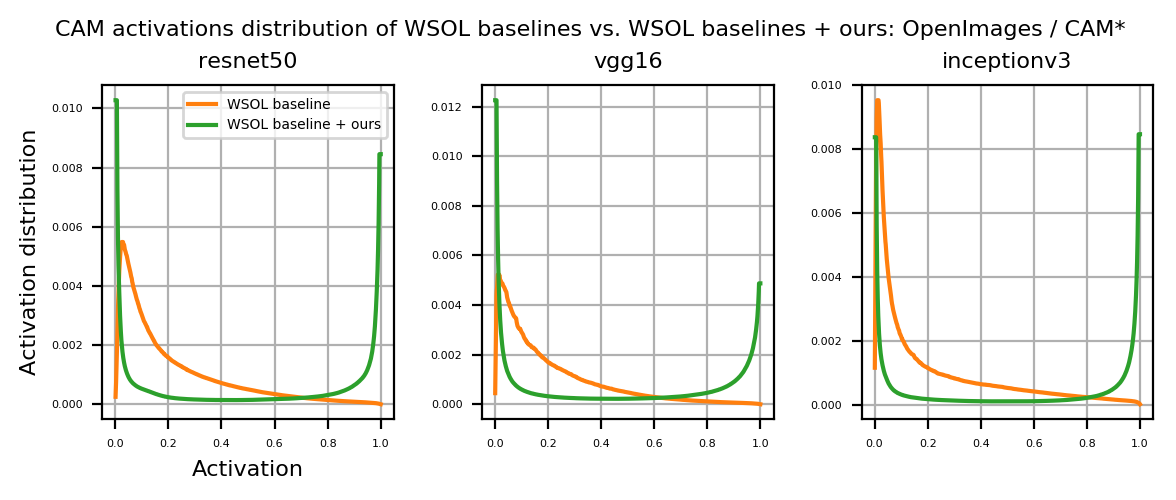}
        \caption{Distribution of activations over the test set for CAM* (orange) and CAM* + ours (green) for different backbones. Top: CUB. Bottom: OpenImages.}
        \label{fig:cam-ac-distribution}
\end{figure}

\noindent \textbf{Activation distribution shift: WSOL baselines with F-CAMs}. Fig.\ref{fig:cam-tau} shows the change in  \maxboxacc and \pxap metrics with respect to ${\tau}$ over the test set using CAM* baseline compared to CAM* + ours\footnote{The rest of the methods are presented in the supplementary material.}. For \maxboxacc metric, notice that the CAM*'s functioning region is concentrated near 0 while the localization performance completely drops to zero when ${\tau}$ starts to increase. This indicates that often the optimum threshold for these methods is very small. On the other hand, when combined with our method, \maxboxacc becomes less sensitive to ${\tau}$. Over \pxap metric (Fig.\ref{fig:cam-pxap-tau}), using our method improves the metric by pushing the curve to the top right corner.
Following \cite{choe2020evaluating}, we inspect the activation distribution of CAMs since they are related to  the shape of the metrics curves (Fig.\ref{fig:cam-ac-distribution}). CAM* shows a single mode with high concentration of activation near zero which explains the shape of the \maxboxacc and \pxap metrics (Fig.\ref{fig:cam-tau}). In addition, this makes the search of the optimum threshold difficult. Moreover, changing the threshold slightly could lead to high variation in localization performance since the threshold is expected to separate the foreground from background. The single mode in the distribution means that the method is unable to separate foreground and background regions, and reflects more uncertainty in CAM interpretability. On the other hand, using F-CAM allows the appearance of a second mode near one. (Mode near zero is for background pixels, while the mode near one is for foreground pixels.) This second mode is more clear over OpenImages dataset. Moreover, both modes are sharper than CAM* alone. This reflects assigning foreground and background to each pixel with more certainty. The reason for this behavior is that our activations are trained using  cross-entropy (Eq.\ref{eq:totalloss}) which pushes activations to be certain (0/1). These results explain why metrics over baseline CAM* perform well only near zero with large sensitivity to thresholding while our method shows less sensitivity.
It is important to note that CAM activation distributions (Fig.\ref{fig:cam-ac-distribution}) are  heavily influenced by the object size. For datasets with small objects, background pixels which are expected to have low activation magnitude will lead typically to density around zero that is higher than the density around one. This can be observed over CUB for our method. On the other hand, over datasets with large objects, density around one is typically expected to be high which is observed over OpenImages using our method.

\noindent \emph{Note:} Supplementary materials contain failure cases, CRF loss, ablation studies, run-time analysis, results on the impact of CAM size vs. localization accuracy, convergence curves of WSOL baselines, and more visual results.

\section{Conclusion}
\label{sec:conclusion}
CAM methods typically rely on interpolation in order to restore full size CAMs in WSOL tasks. To improve the localization accuracy of CAMs, this paper proposes to connect a trainable decoding architecture to a CNN classifier, allowing for parametric upscaling of CAMs to accurate full resolution CAMs (F-CAMs). Low resolution CAMs and variants priors are used to fine-tune the decoder. Evaluated in combination with six baseline WSOL methods and three CNN backbones, our F-CAM methods improves the performance of these baselines by a large margin on CUB and OpenImage datasets.  F-CAM performance is competitive with state-of-the-art WSOL methods, yet it requires less computation resources during inference.

\noindent \textbf{Acknowledgment:} This research was supported in part by the Canadian Institutes of Health Research, Natural Sciences and Engineering Research Council of Canada, Compute Canada and MITACS.

\FloatBarrier

\renewcommand{\theequation}{\thesection.\arabic{equation}}
\setcounter{equation}{0}

\renewcommand\thefigure{\thesection.\arabic{figure}}
\setcounter{figure}{0}

\renewcommand\thetable{\thesection.\arabic{table}}
\setcounter{table}{0}

\appendices

%
%
%
%

\section{Supplementary material}

We provide in this supplementary document:

\noindent \textbf{1)} General description of CRF loss ${\mathcal{R}}$ ~\cite{tang2018regularized}.

\noindent \textbf{2)} Failure cases (Sec.\ref{sec:fail}).

\noindent \textbf{3)} Ablation study (Sec.\ref{sec:ablations}).

\noindent \textbf{4)} Simulation of the impact of downscale factor of CAMs over the localization performance (Sec.\ref{sec:sim-impact-factor}).

\noindent \textbf{5)} Runtime inference for WSOL baselines and our method (Sec.\ref{sec:runtime}) and number of parameters per model.

\noindent \textbf{6)} \newmaxboxacc performance over CUB (Tab.\ref{tab:maxboxv2}).

\noindent  \textbf{7)} Classification performance of different WSOL baselines (Tab.\ref{tab:main_cls}).

\begin{figure*}
     \centering
     \begin{subfigure}[b]{0.33\textwidth}
         \centering
         \includegraphics[width=\textwidth]{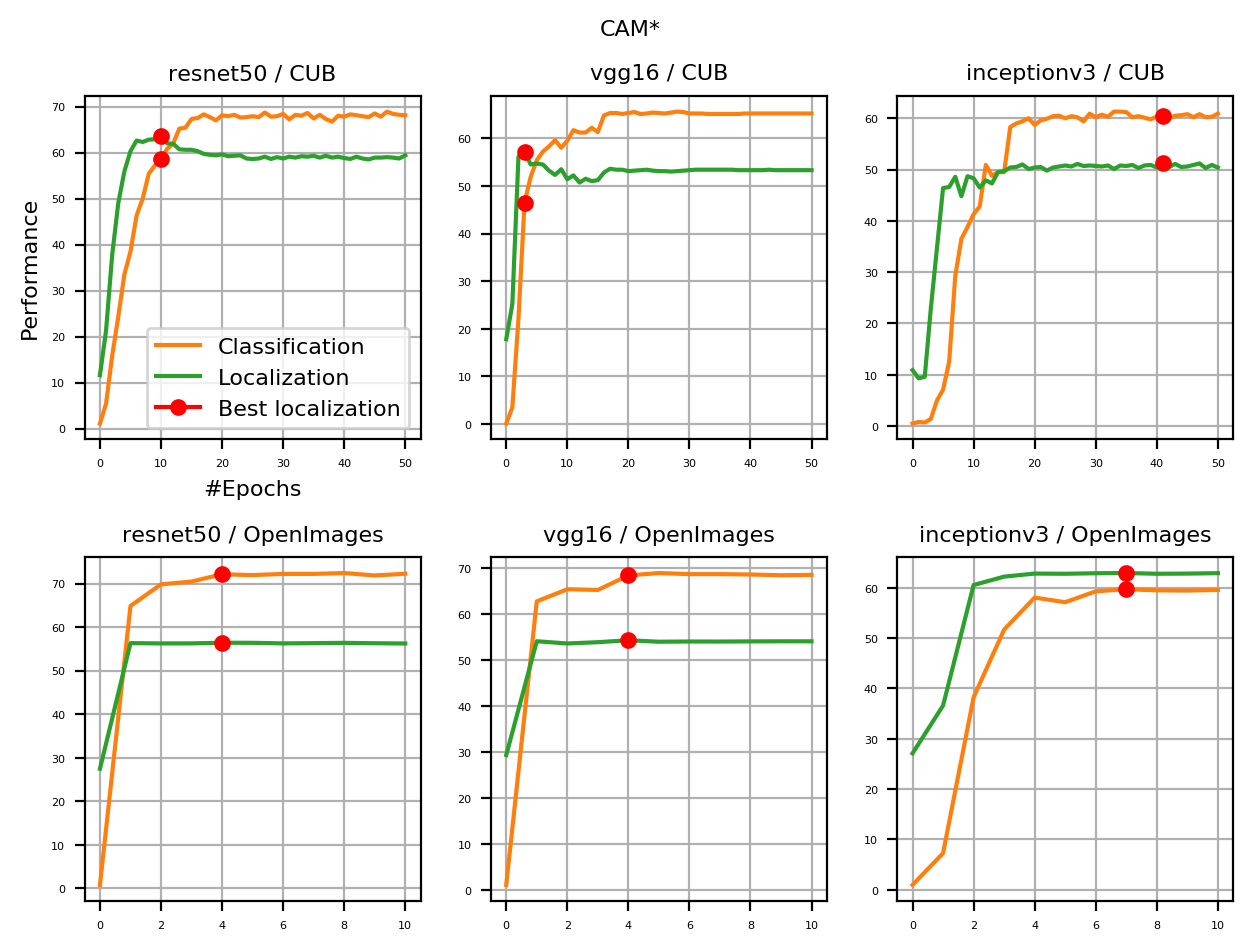}
         \caption{CAM*}
         \label{fig:y equals x}
     \end{subfigure}
     \hfill
     \begin{subfigure}[b]{0.33\textwidth}
         \centering
         \includegraphics[width=\textwidth]{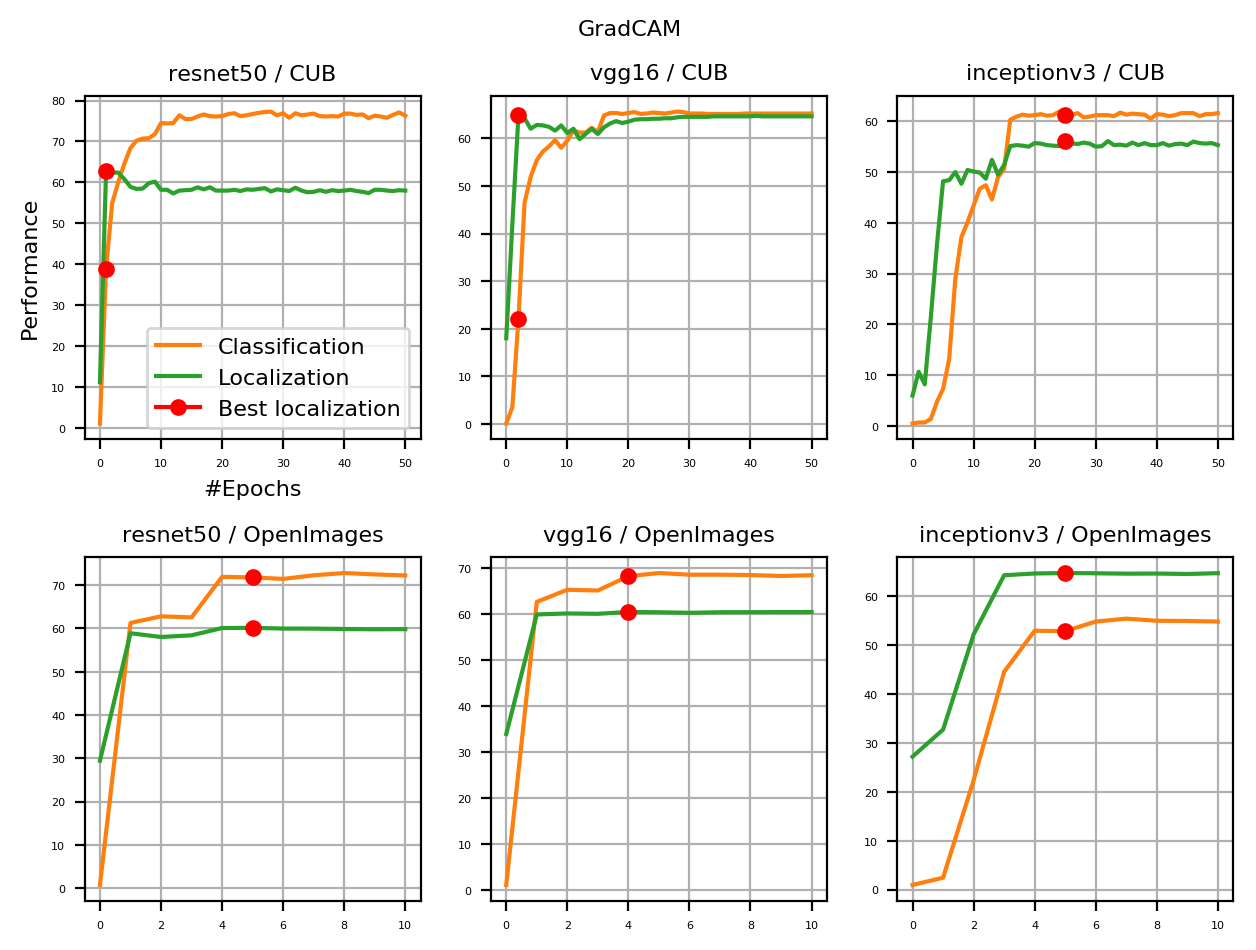}
         \caption{GradCAM}
         \label{fig:three sin x}
     \end{subfigure}
     \hfill
     \begin{subfigure}[b]{0.33\textwidth}
         \centering
         \includegraphics[width=\textwidth]{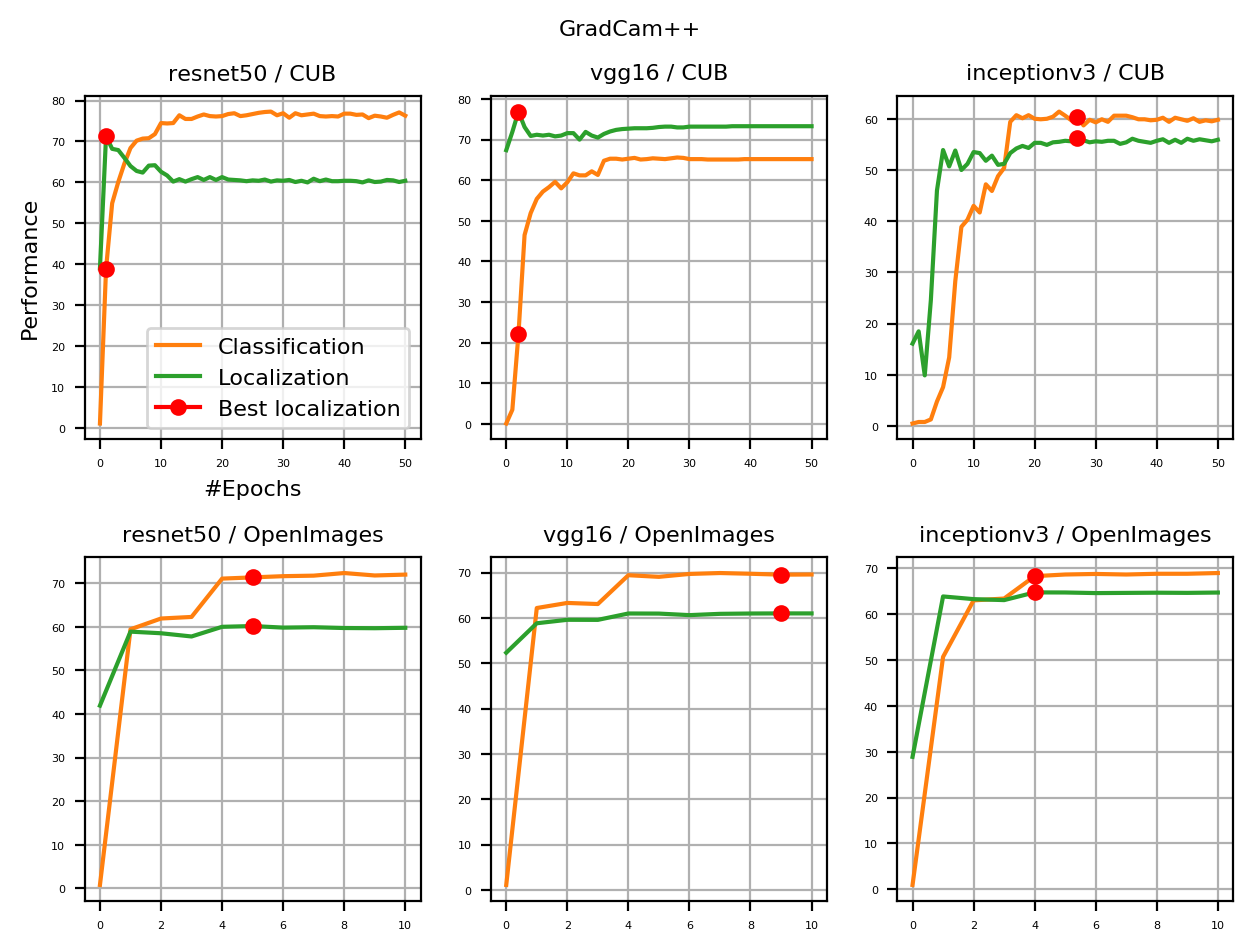}
         \caption{GradCAM++}
         \label{fig:five over x}
     \end{subfigure}
     \\
     \begin{subfigure}[b]{0.33\textwidth}
         \centering
         \includegraphics[width=\textwidth]{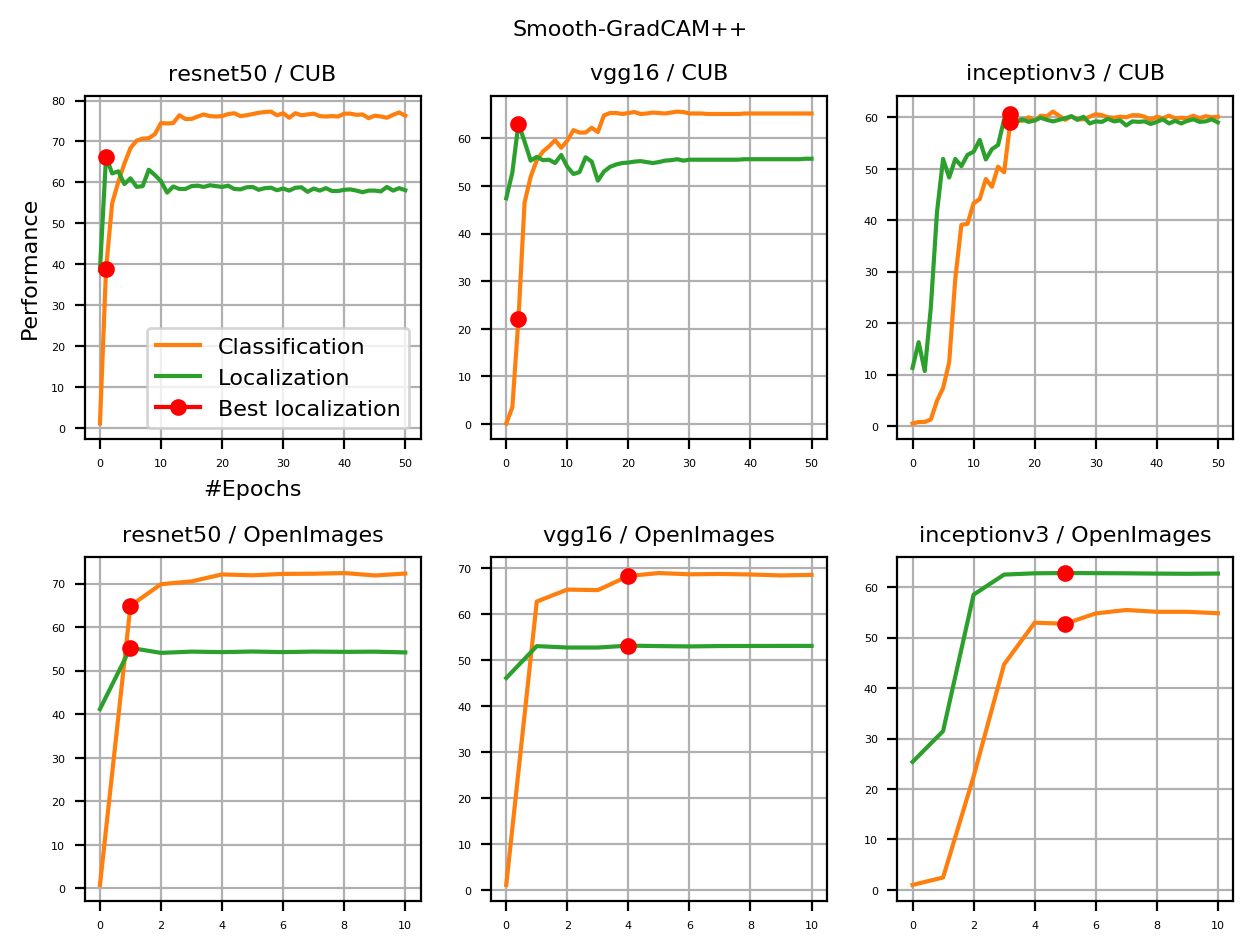}
         \caption{Smooth-GradCAM++}
         \label{fig:y equals x}
     \end{subfigure}
     \hfill
     \begin{subfigure}[b]{0.33\textwidth}
         \centering
         \includegraphics[width=\textwidth]{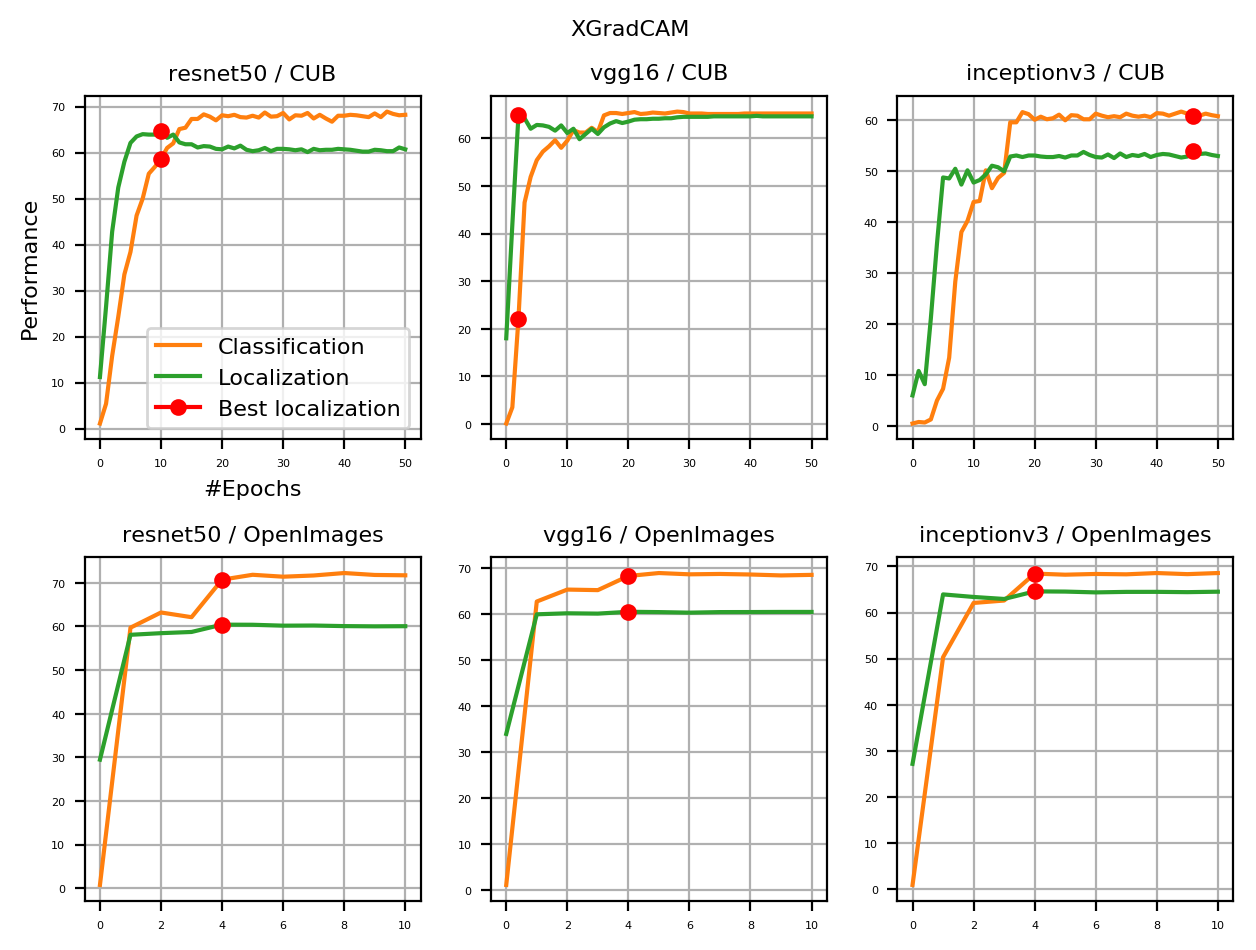}
         \caption{XGradCAM}
         \label{fig:three sin x}
     \end{subfigure}
     \hfill
     \begin{subfigure}[b]{0.33\textwidth}
         \centering
         \includegraphics[width=\textwidth]{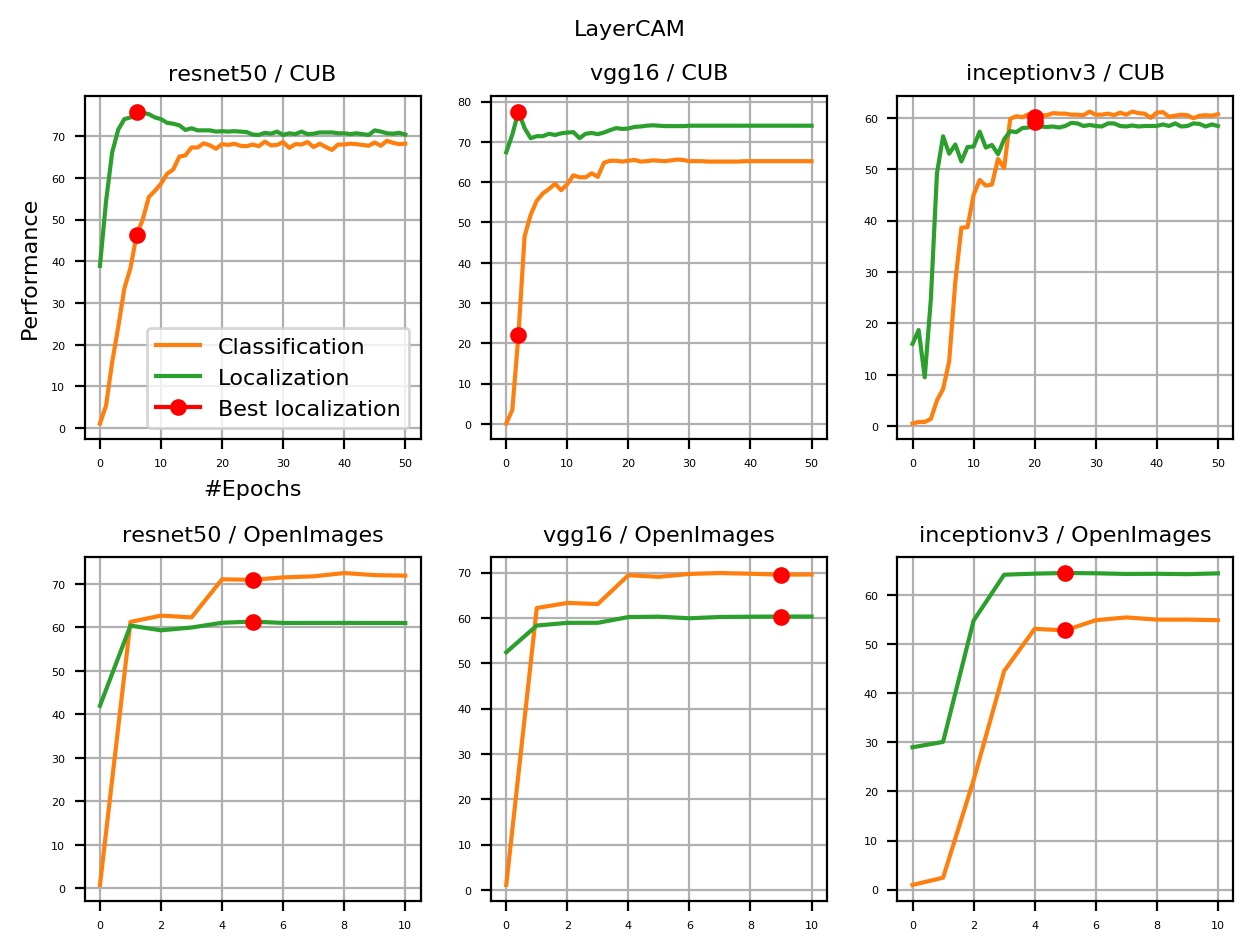}
         \caption{LayerCAM}
         \label{fig:five over x}
     \end{subfigure}
        \caption{Convergence of classification and localization tasks over validation set over baselines WSOL methods. \textcolor{red}{The red dot} is the epoch for the selected model based on the localization performance using \maxboxacc metric.}
        \label{fig:convergence-std-wsol}
\end{figure*}

\noindent  \textbf{8)} Convergence curves of WSOL baseline training. (Fig.\ref{fig:convergence-std-wsol}).

\begin{figure*}
     \centering
     \begin{subfigure}[b]{0.49\textwidth}
         \centering
         \includegraphics[width=\textwidth]{CUB-CAM-chp-best-bv2-False-subset-test}
         \caption{CAM*}
         \label{fig:cub-06}
     \end{subfigure}
     \hfill
     \begin{subfigure}[b]{0.49\textwidth}
         \centering
         \includegraphics[width=\textwidth]{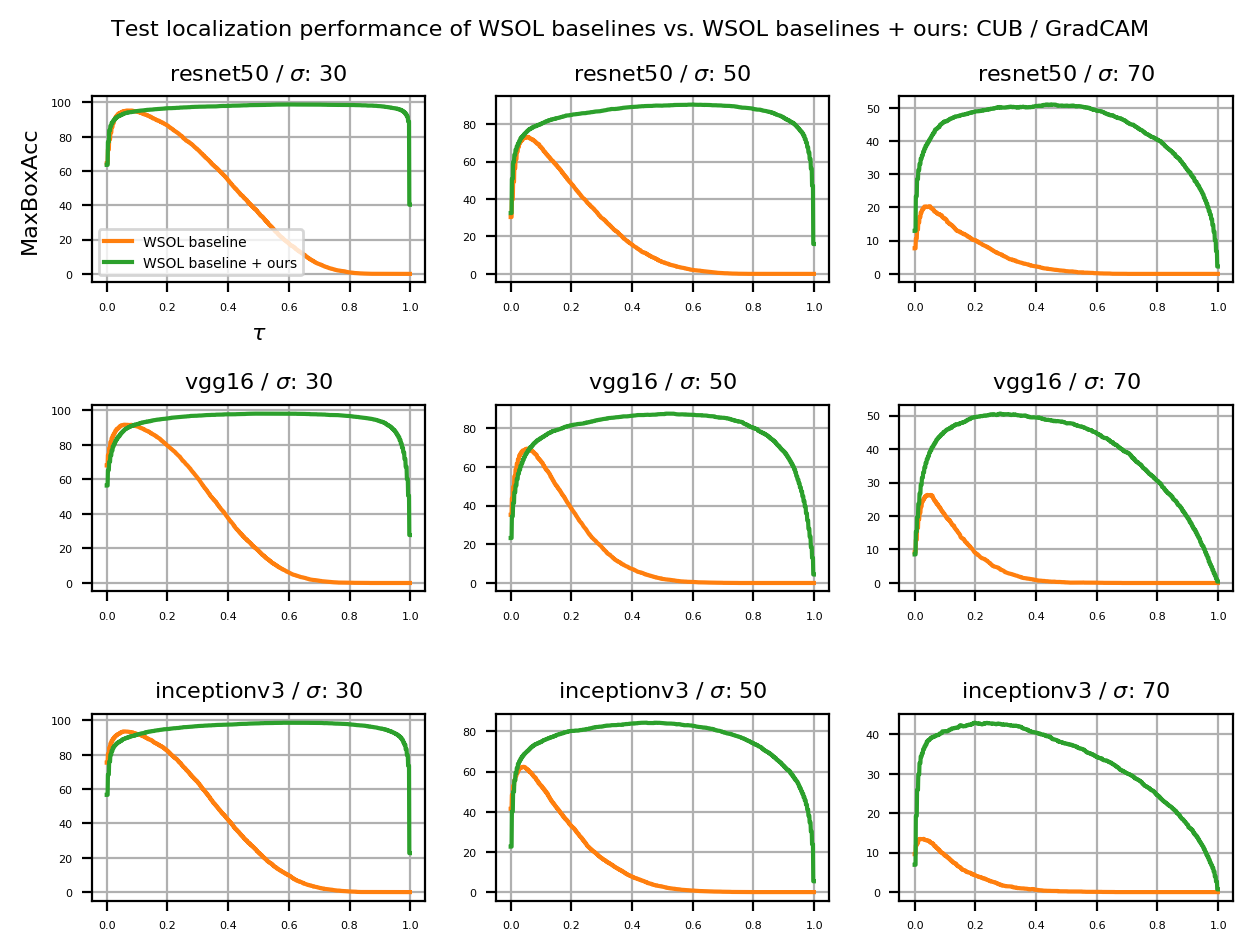}
         \caption{GradCAM}
         \label{fig:cub-05}
     \end{subfigure}
     \\
     \begin{subfigure}[b]{0.49\textwidth}
         \centering
         \includegraphics[width=\textwidth]{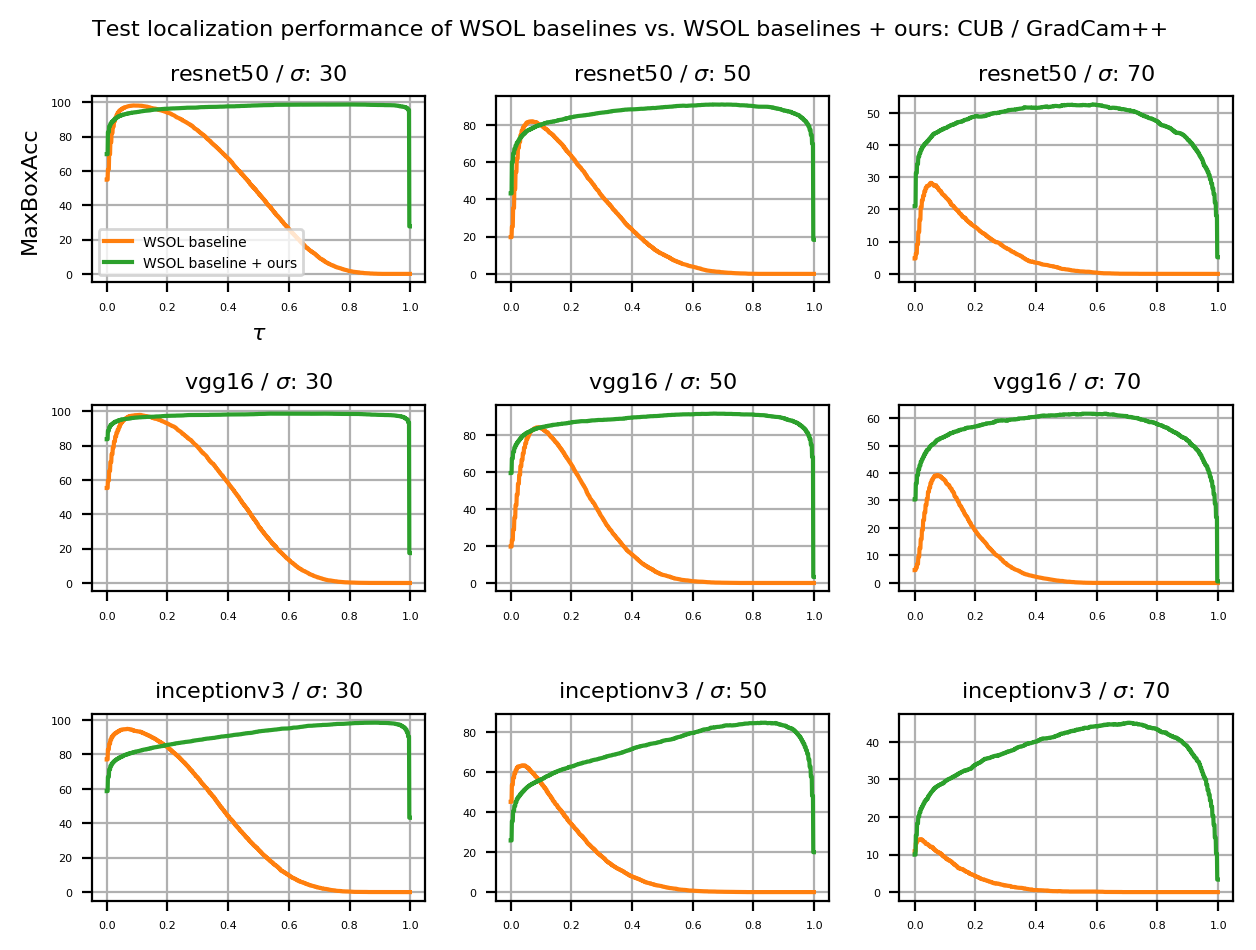}
         \caption{GradCAM++}
         \label{fig:cub-04}
     \end{subfigure}
     \begin{subfigure}[b]{0.49\textwidth}
         \centering
         \includegraphics[width=\textwidth]{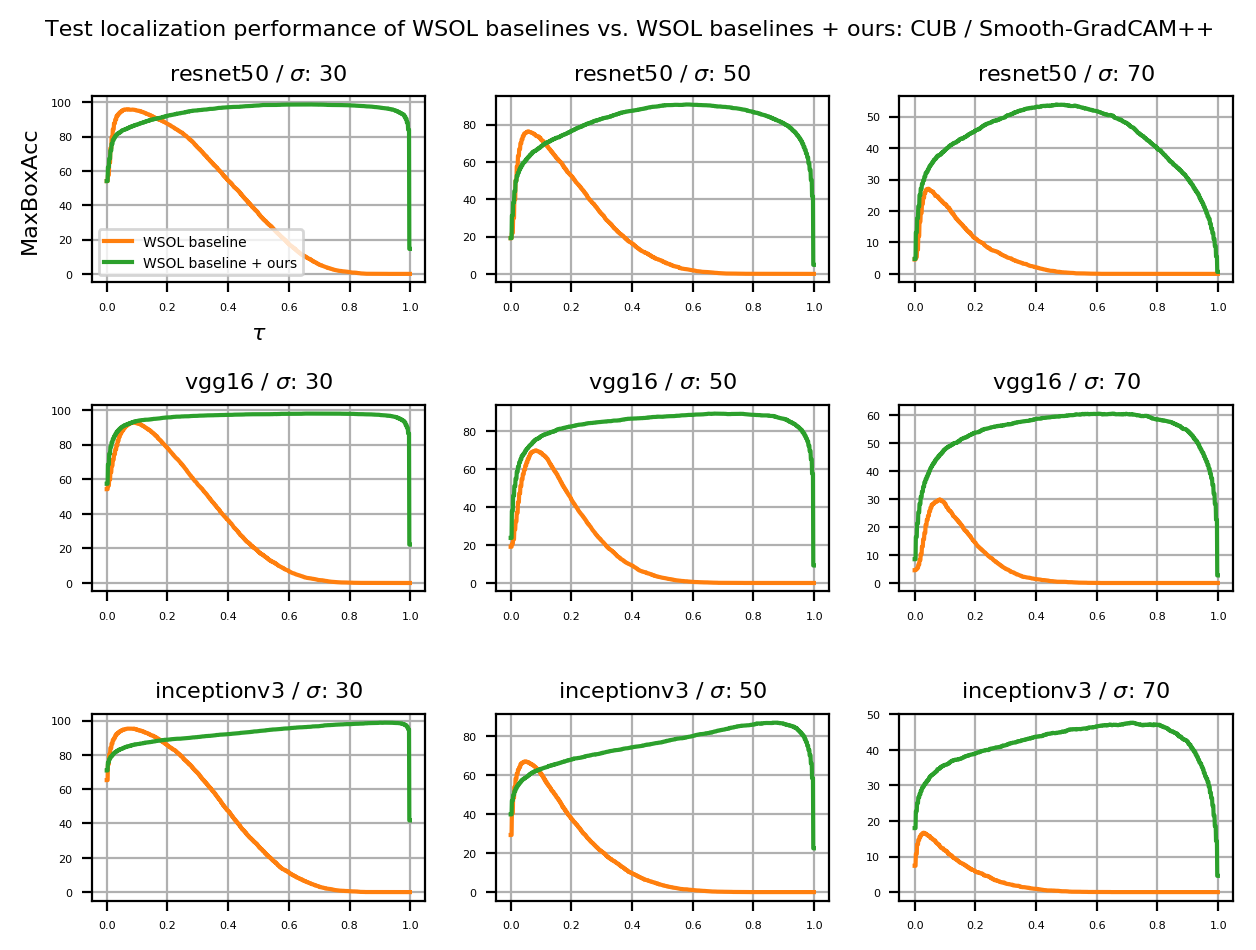}
         \caption{Smooth-GradCAM++}
         \label{fig:cub-03}
     \end{subfigure}
     \\
     \begin{subfigure}[b]{0.49\textwidth}
         \centering
         \includegraphics[width=\textwidth]{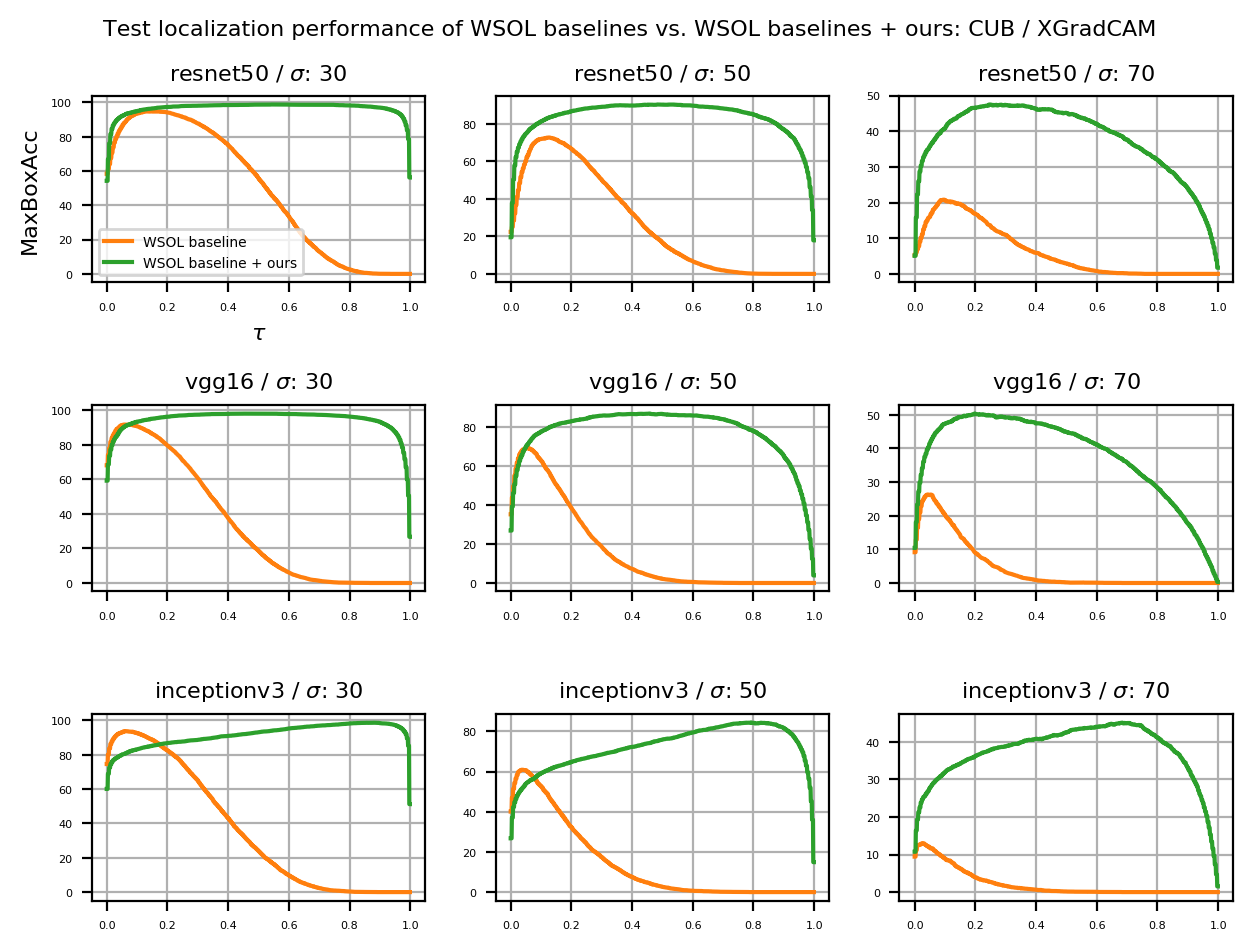}
         \caption{XGradCAM}
         \label{fig:cub-02}
     \end{subfigure}
     \hfill
     \begin{subfigure}[b]{0.49\textwidth}
         \centering
         \includegraphics[width=\textwidth]{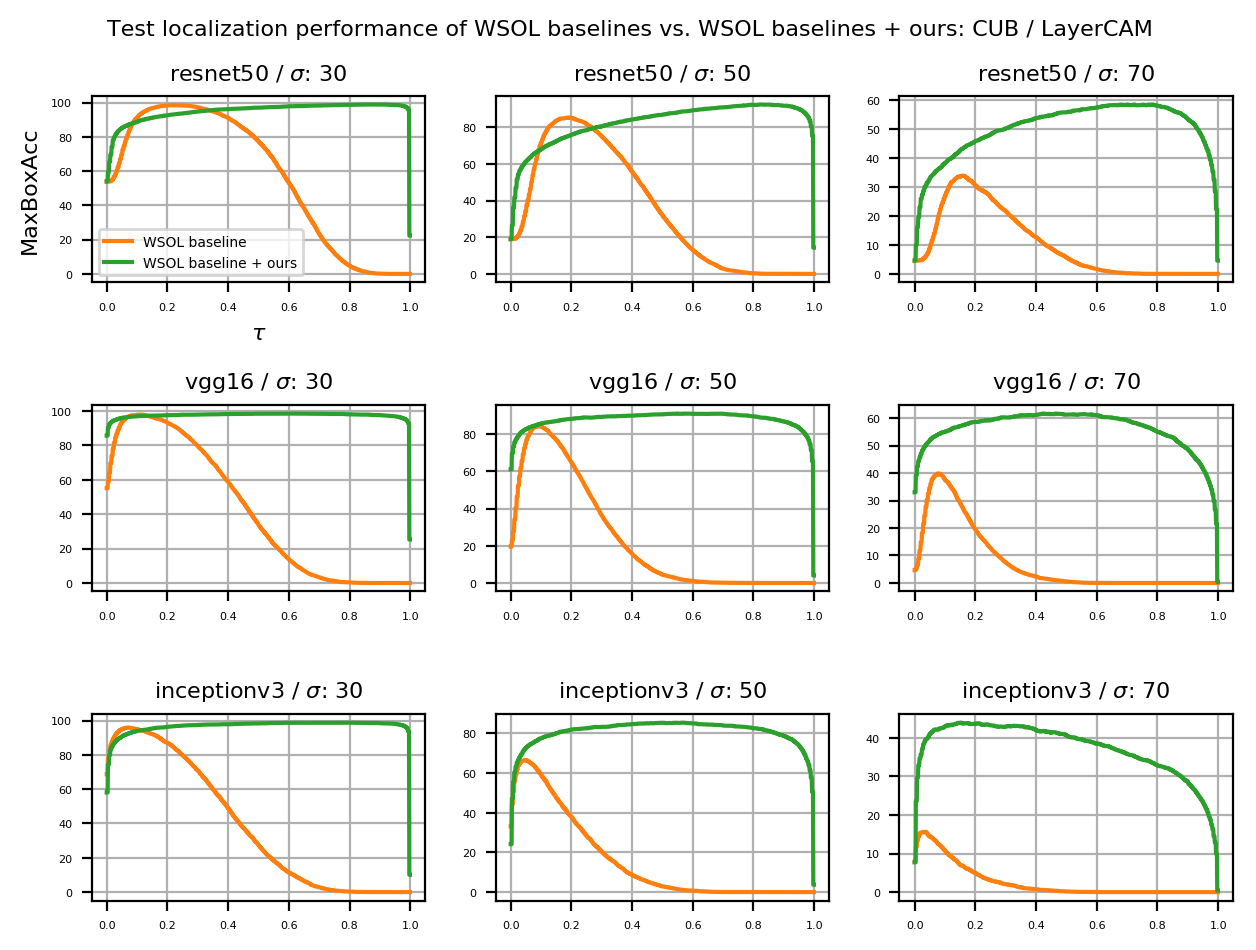}
         \caption{LayerCAM}
         \label{fig:cub-01}
     \end{subfigure}
        \caption{CUB test set: WSOL baselines vs. WSOL baselines + ours  validated with \maxboxacc.}
        \label{fig:ours-vs-baselines-perf-cub}
\end{figure*}

\begin{figure*}
     \centering
     \begin{subfigure}[b]{0.49\textwidth}
         \centering
         \includegraphics[width=\textwidth]{OpenImages-CAM-chp-best-bv2-False-subset-test}
         \caption{CAM*}
         \label{fig:opim-06}
     \end{subfigure}
     \hfill
     \begin{subfigure}[b]{0.49\textwidth}
         \centering
         \includegraphics[width=\textwidth]{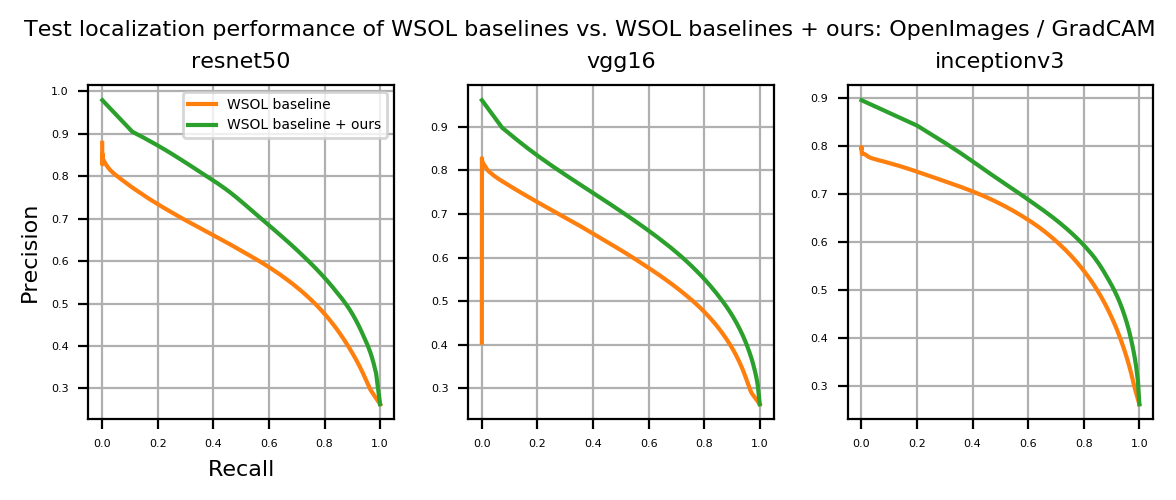}
         \caption{GradCAM}
         \label{fig:opim-05}
     \end{subfigure}
     \\
     \begin{subfigure}[b]{0.49\textwidth}
         \centering
         \includegraphics[width=\textwidth]{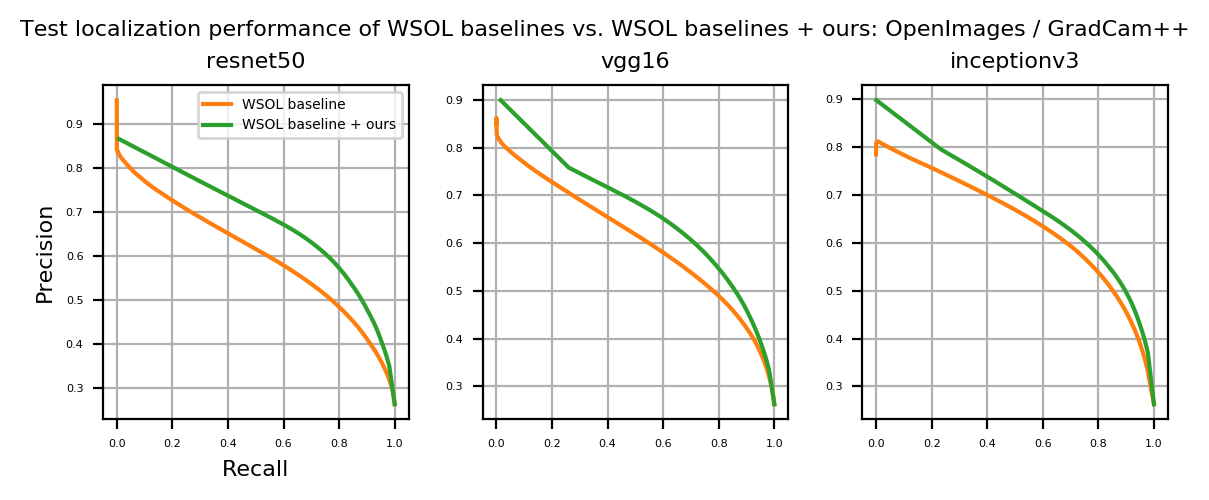}
         \caption{GradCAM++}
         \label{fig:opim-04}Aydin Sarraf
     \end{subfigure}
     \begin{subfigure}[b]{0.49\textwidth}
         \centering
         \includegraphics[width=\textwidth]{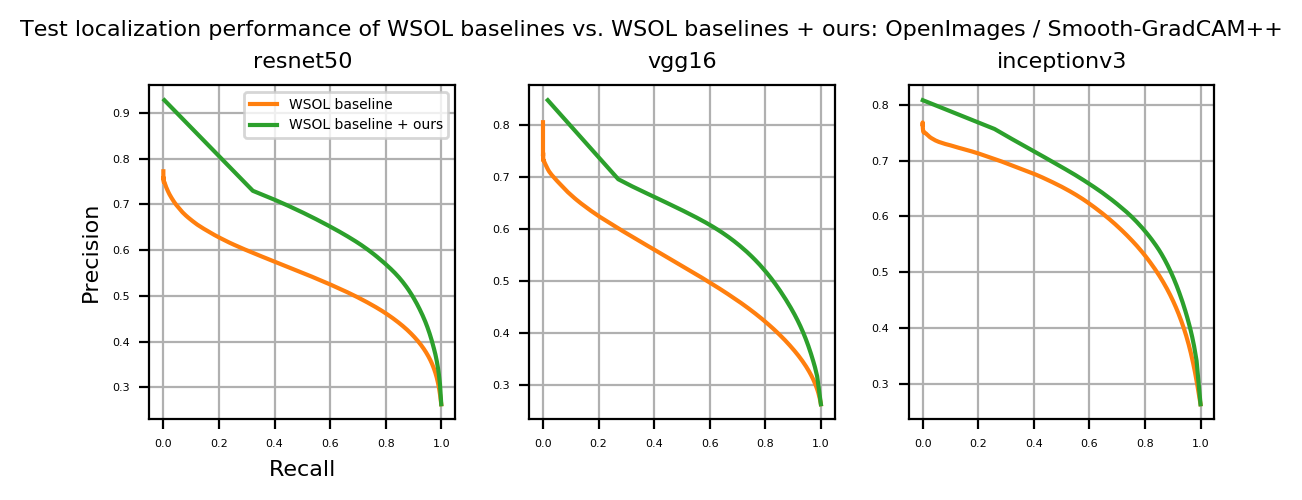}
         \caption{Smooth-GradCAM++}
         \label{fig:opim-03}
     \end{subfigure}
     \\
     \begin{subfigure}[b]{0.49\textwidth}
         \centering
         \includegraphics[width=\textwidth]{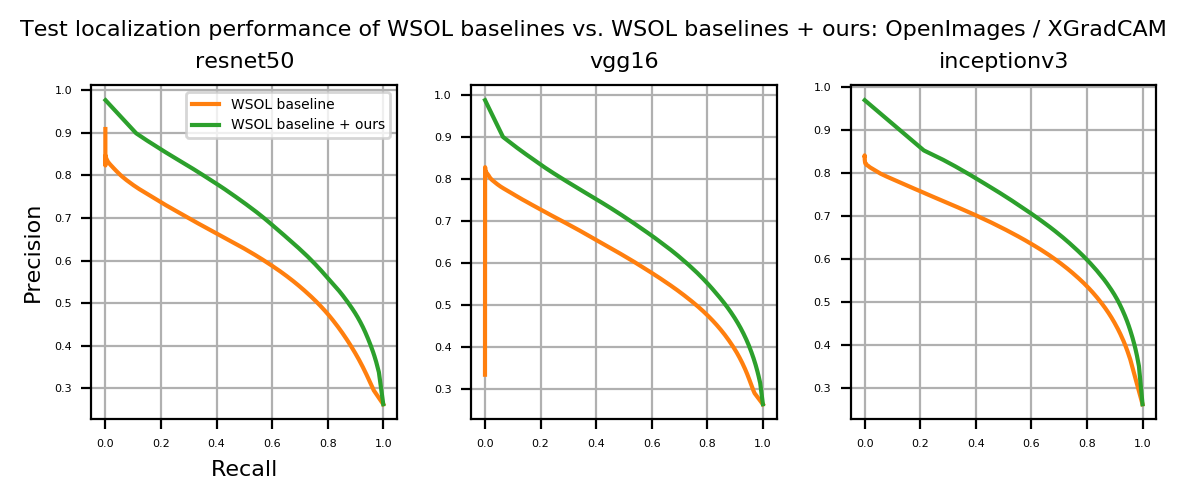}
         \caption{XGradCAM}
         \label{fig:opim-02}
     \end{subfigure}
     \hfill
     \begin{subfigure}[b]{0.49\textwidth}
         \centering
         \includegraphics[width=\textwidth]{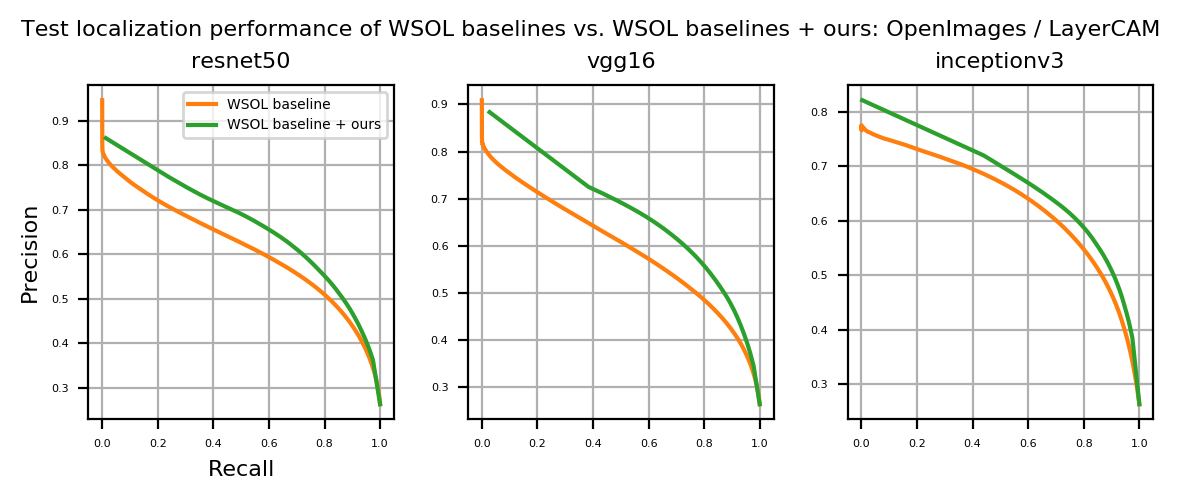}
         \caption{LayerCAM}
         \label{fig:opim-01}
     \end{subfigure}
        \caption{OpenImages test set: WSOL baselines vs. WSOL baselines + ours  validated with \maxboxacc.}
        \label{fig:ours-vs-baselines-perf-openimages}
\end{figure*}

\noindent  \textbf{9)} Comparison of localization performance curves (Fig.\ref{fig:ours-vs-baselines-perf-cub},\ref{fig:ours-vs-baselines-perf-openimages}).

\begin{figure*}
     \centering
     \begin{subfigure}[b]{0.49\textwidth}
         \centering
         \includegraphics[width=\textwidth]{cam-dist-CUB-CAM-chp-best-bv2-False-subset-test}
         \caption{CAM*}
         \label{fig:dist6cub}
     \end{subfigure}
     \hfill
     \begin{subfigure}[b]{0.49\textwidth}
         \centering
         \includegraphics[width=\textwidth]{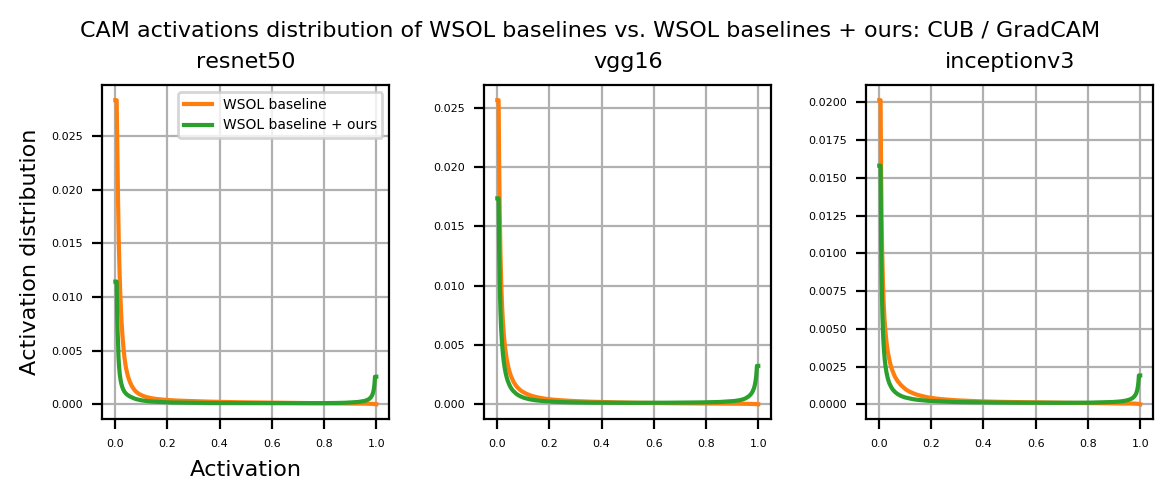}
         \caption{GradCAM}
         \label{fig:dist5cub}
     \end{subfigure}
     \\
     \begin{subfigure}[b]{0.49\textwidth}
         \centering
         \includegraphics[width=\textwidth]{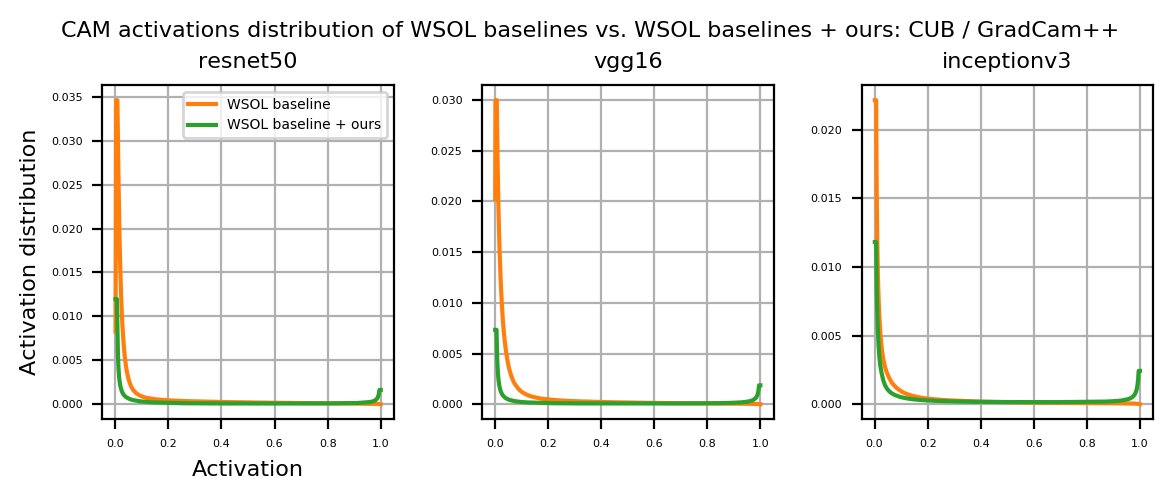}
         \caption{GradCAM++}
         \label{fig:dist4cub}
     \end{subfigure}
     \begin{subfigure}[b]{0.49\textwidth}
         \centering
         \includegraphics[width=\textwidth]{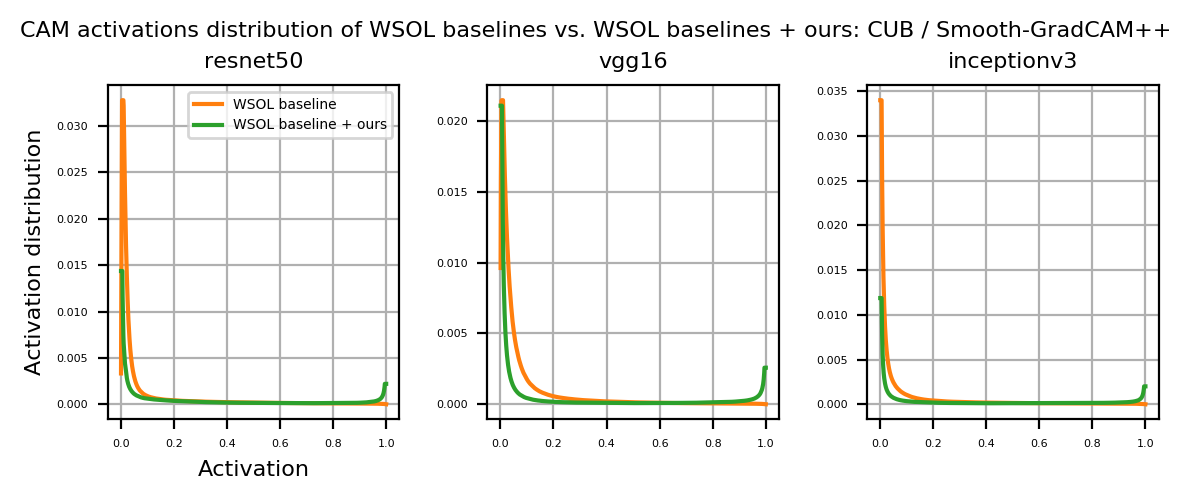}
         \caption{Smooth-GradCAM++}
         \label{fig:dist3cub}
     \end{subfigure}
     \\
     \begin{subfigure}[b]{0.49\textwidth}
         \centering
         \includegraphics[width=\textwidth]{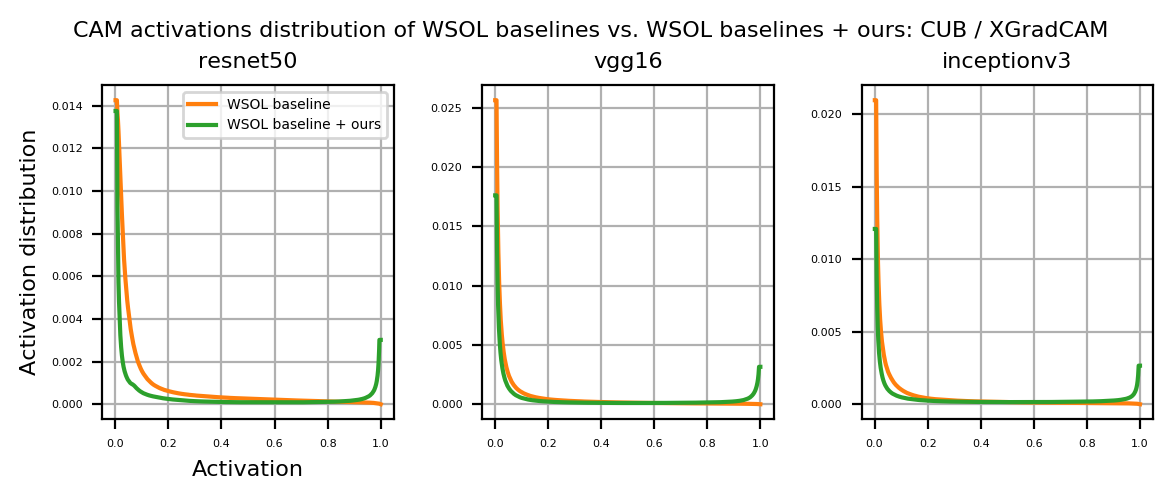}
         \caption{XGradCAM}
         \label{fig:dist2cub}
     \end{subfigure}
     \hfill
     \begin{subfigure}[b]{0.49\textwidth}
         \centering
         \includegraphics[width=\textwidth]{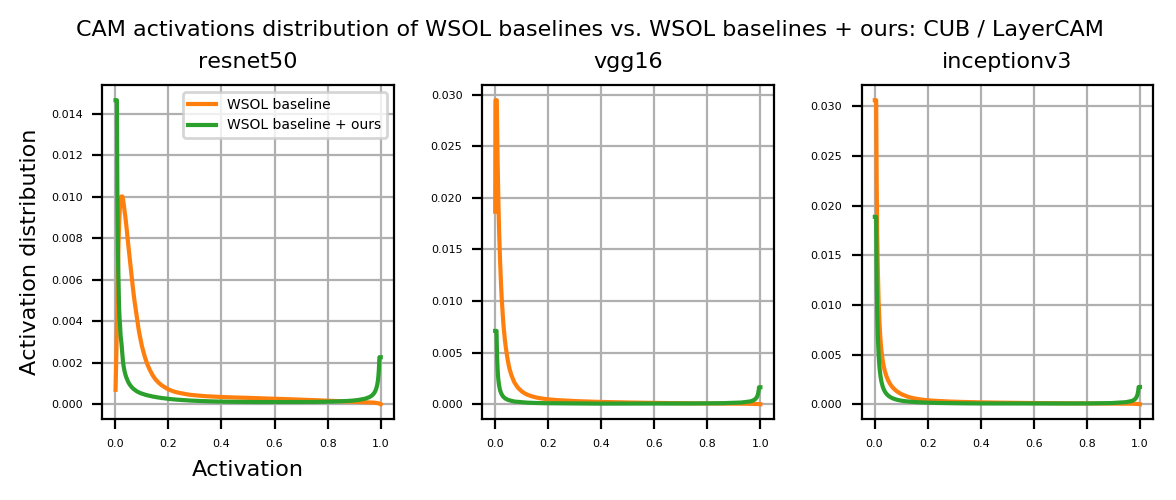}
         \caption{LayerCAM}
         \label{fig:dist1cub}
     \end{subfigure}
        \caption{CAM's activation distribution over CUB test set: WSOL baselines vs. WSOL baseline + ours  validated with \maxboxacc.}
        \label{fig:cub-cam-dist}
\end{figure*}

\begin{figure*}
     \centering
     \begin{subfigure}[b]{0.49\textwidth}
         \centering
         \includegraphics[width=\textwidth]{cam-dist-OpenImages-CAM-chp-best-bv2-False-subset-test}
         \caption{CAM*}
         \label{fig:dist6oim}
     \end{subfigure}
     \hfill
     \begin{subfigure}[b]{0.49\textwidth}
         \centering
         \includegraphics[width=\textwidth]{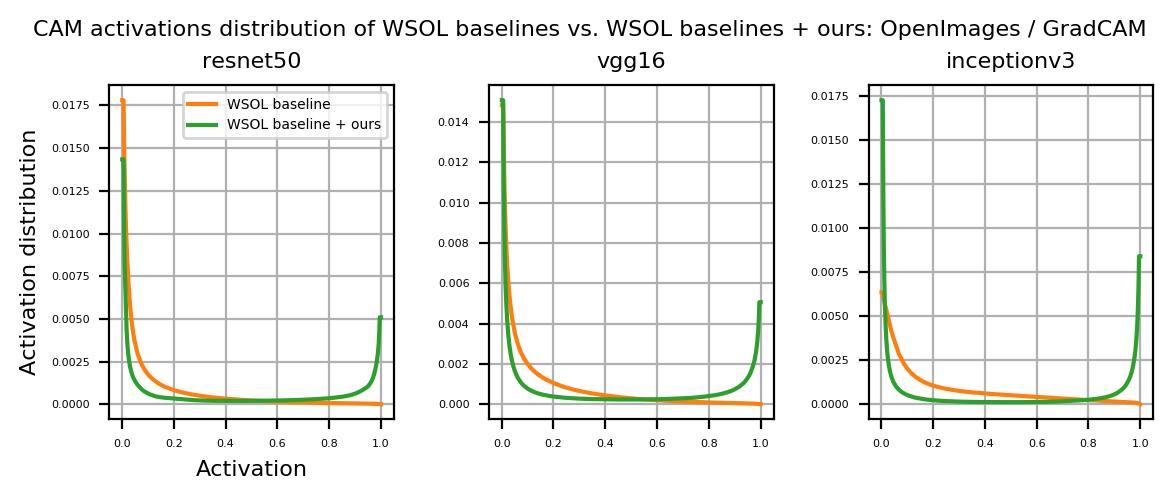}
         \caption{GradCAM}
         \label{fig:dist5oim}
     \end{subfigure}
     \\
     \begin{subfigure}[b]{0.49\textwidth}
         \centering
         \includegraphics[width=\textwidth]{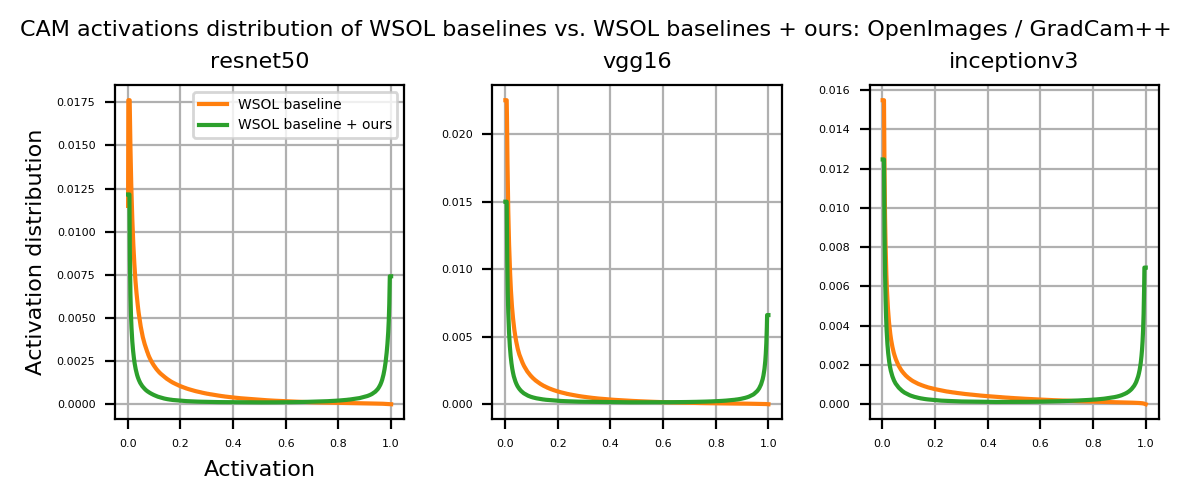}
         \caption{GradCAM++}
         \label{fig:dist4oim}
     \end{subfigure}
     \begin{subfigure}[b]{0.49\textwidth}
         \centering
         \includegraphics[width=\textwidth]{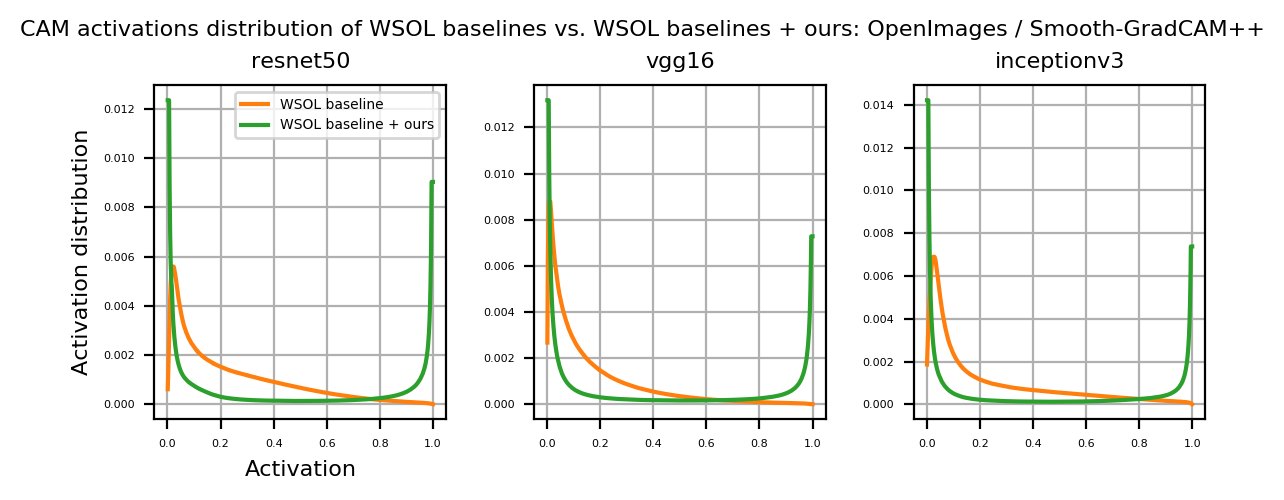}
         \caption{Smooth-GradCAM++}
         \label{fig:dis3oim}
     \end{subfigure}
     \\
     \begin{subfigure}[b]{0.49\textwidth}
         \centering
         \includegraphics[width=\textwidth]{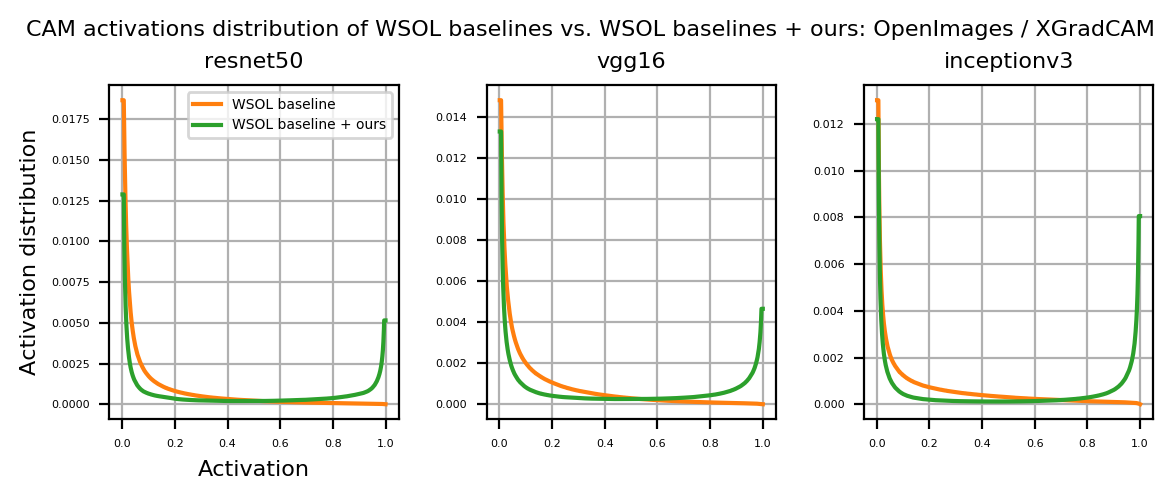}
         \caption{XGradCAM}
         \label{fig:dist2oim}
     \end{subfigure}
     \hfill
     \begin{subfigure}[b]{0.49\textwidth}
         \centering
         \includegraphics[width=\textwidth]{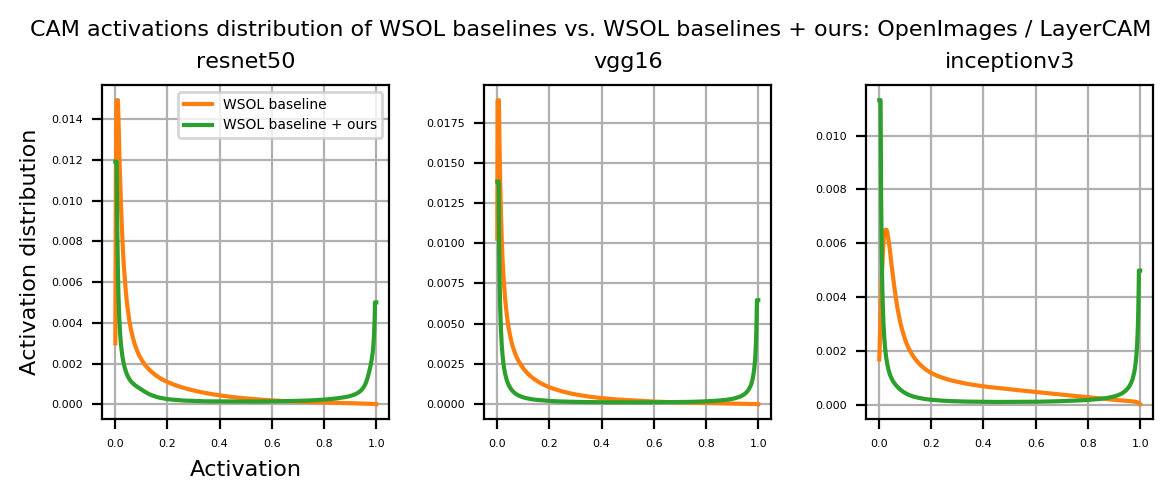}
         \caption{LayerCAM}
         \label{fig:dist1oim}
     \end{subfigure}
        \caption{CAM's activation distribution over OpenImages test set: WSOL baselines vs. WSOL baseline + ours  validated with \maxboxacc.}
        \label{fig:openimages-cam-dist}
\end{figure*}

\noindent  \textbf{10)} Comparison to CAMs activations distributions (Fig.\ref{fig:cub-cam-dist}, \ref{fig:openimages-cam-dist}).

\section{General description of CRF loss}
\label{sec:crf}
Given an input image $\bm{X}$ and the softmax activation ${\bm{S}}$ of the decoder, the CRF loss is formulated~\cite{tang2018regularized}  as,
\begin{equation}
    \label{eq:crf}
    \mathcal{R}(\bm{S}, \bm{X}) = \sum_{t=1}^{t=2} {\bm{S}^t}^{\top} \; \bm{W} \; (\bm{1} - \bm{S}^t) \;,
\end{equation}
where ${\bm{W}}$ is an affinity matrix where ${\bm{W}[i, j]}$ captures the color similarity and proximity between pixels ${i, j}$ in the image ${\bm{X}}$. We consider using Gaussian kernel to capture color and spatial similarities~\cite{KrahenbuhlK11crf}. We use the permutohedral lattice~\cite{AdamsBD10lattice} for fast computation of ${\bm{W}}$. Minimizing Eq.\ref{eq:crf} pushes the decoder to produce consistent activations for nearby pixels with similar color.

\section{Failure cases.}
\label{sec:fail}

\begin{figure}
     \centering
     \begin{subfigure}[b]{0.45\textwidth}
         \centering
         \includegraphics[width=.95\textwidth]{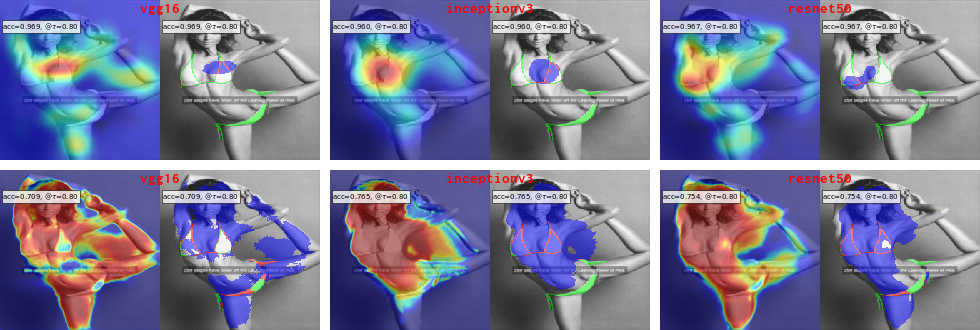}
     \end{subfigure}
     \\
     \vspace{.2cm}
     \begin{subfigure}[b]{0.45\textwidth}
         \centering
         \includegraphics[width=.95\textwidth]{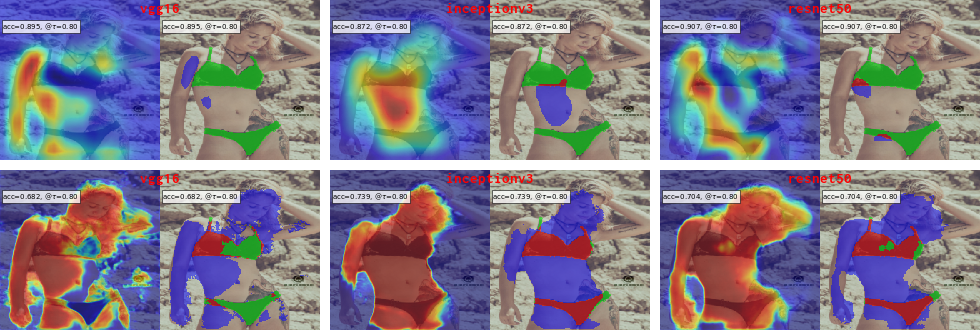}
     \end{subfigure}
     \\
     \vspace{.2cm}
     \begin{subfigure}[b]{0.45\textwidth}
         \centering
         \includegraphics[width=.95\textwidth]{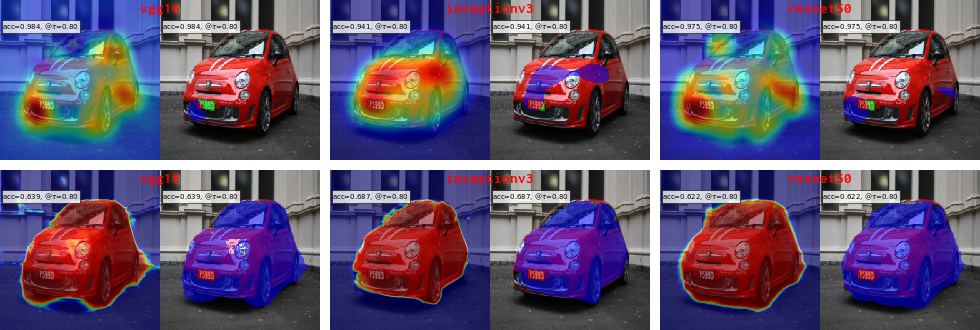}
     \end{subfigure}
        \caption{Failed cases of our method combined with CAM* over OpenImages test set.
        Colors: green false negative, red : true positive, blue: false positive. ${\tau=0.8}$.}
        \label{fig:failure}
\end{figure}

Fig.\ref{fig:failure} illustrates few examples over OpenImages test set where our method failed to localize the correct object. The fine-tuning of our method is guided mainly by the activations of the WSOL baseline. When the activations largely miss the correct object, our method learns on wrong supervisory signal leading to false localization. Detecting these cases and dealing with them remains an open issue in this work. This scenario goes under learning from highly noisy supervisory signals which is still a growing field~\cite{songsurvey2020}.

\section{Ablation study}
\label{sec:ablations}
\begin{figure}[t!]
\centering
  \centering
  \includegraphics[width=\linewidth]{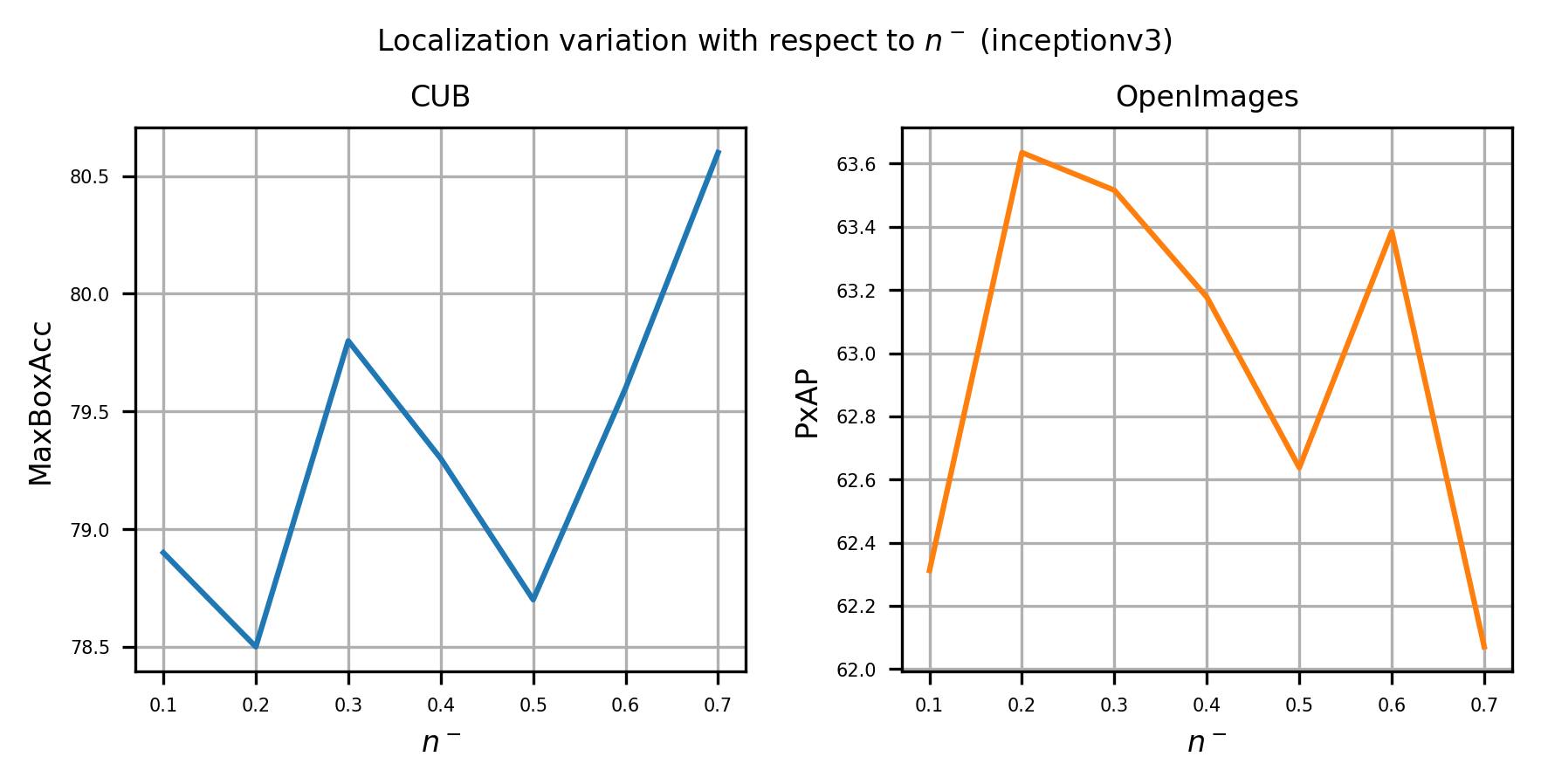}
  \caption{Impact of ${n^-}$ on the localization performance over CUB and OpenImages validation with CAM* baseline + ours. Random runs are done while the rest of hyper-parameters are fixed.}
  \label{fig:ablation-n--}
\end{figure}

Tab.\ref{tab:ablation-parts} shows the impact of adding different terms in the pixel alignment loss in our training loss on all backbones and on both datasets using CAM* method. We observe that simply adding sampling regions of foreground and background yields a large improvement of the baseline. This shows the importance of using supervisory signals at pixel level to guide CAMs. Baseline methods use only classification signal which is a global information. Adding local guidance helps better discerning foreground from background. Adding a CRF term and size constraint helps better extending the foreground while respecting object's boundaries yielding more improvement.
We report in Fig.\ref{fig:ablation-n--} the impact of the hyper-parameter ${n^-}$ on the localization performance. Note that ${n^-}$ describes the extent of the background region we are allowed to sample from. It is interconnected with the object size. Datasets with large objects result in small background and small objects lead to large background. This is reflected in these curves. On CUB, often objects are small. Therefore, assuming large background is safe. As a result, large values of ${n^-}$  yield better results. On the other hand, objects in OpenImages dataset tend to be large leaving small background, making small values of ${n^-}$ more efficient. This gives us an intuition for better range for ${n^-}$.

{
\setlength{\tabcolsep}{3pt}
\renewcommand{\arraystretch}{1.1}
\begin{table}[ht!]
\centering
\resizebox{.5\textwidth}{!}{%
\centering
\small
\begin{tabular}{lc*{3}{c}gc*{3}{c}g}
& &  \multicolumn{4}{c}{CUB (\maxboxacc)}  & & \multicolumn{4}{c}{OpenImages (\pxap)} \\
Methods  &  &  VGG & Inception & ResNet & Mean &  &  VGG &  Inception & ResNet & Mean  \\
\cline{1-1}\cline{3-6}\cline{8-11} \\
CAM*~\cite{zhou2016learning} &  &                     \tableplus{61.6} & \tableplus{58.8} & \tableplus{71.5} & \tableplus{63.9} &  & \tableplus{53.0} & \tableplus{62.7} & \tableplus{56.8} & \tableplus{57.5} \\
CAM*~\cite{zhou2016learning} + SR &  &             84.2 & 73.0 & 82.2 & 79.8    &  & 64.5 & 64.1 & 63.8 & 64.1 \\
CAM*~\cite{zhou2016learning} + SR + ASC &  &       82.9 & 74.1 & 83.2 & 80.0 &  & 63.9 & 63.4 & 62.0 & 63.1 \\
CAM*~\cite{zhou2016learning} + SR + CRF &  &       84.6 & 78.9 & 86.1 & 83.2 &  & 66.3 & 68.3 & 67.5 & 67.3 \\
CAM*~\cite{zhou2016learning} + SR + CRF + ASC &  & \tableplus{87.3} & \tableplus{82.0} & \tableplus{90.3} & \tableplus{86.5} &  & \tableplus{67.8} & \tableplus{71.9} & \tableplus{72.1} & \tableplus{70.6} \\
\cline{1-1}\cline{3-6}\cline{8-11}\\
Improvement &  &                                      \tableplus{+25.7} & \tableplus{+23.2} & \tableplus{+18.8} & \tableplus{+22.5} &  & \tableplus{+14.8} & \tableplus{+9.2} & \tableplus{+15.3} & \tableplus{+12.8} \\
\cline{1-1}\cline{3-6}\cline{8-11}\\
\end{tabular}
}
\caption{Ablation study of different terms in the pixel alignment loss  over CAM* baseline. The bottom line in green is the improvement over the WSOL baseline CAM* (top green line), when combined with our full method 4th green line.}
\label{tab:ablation-parts}
\vspace{-1em}
\end{table}
}

\begin{figure}[ht!]
\centering
  \centering
  \includegraphics[width=\linewidth]{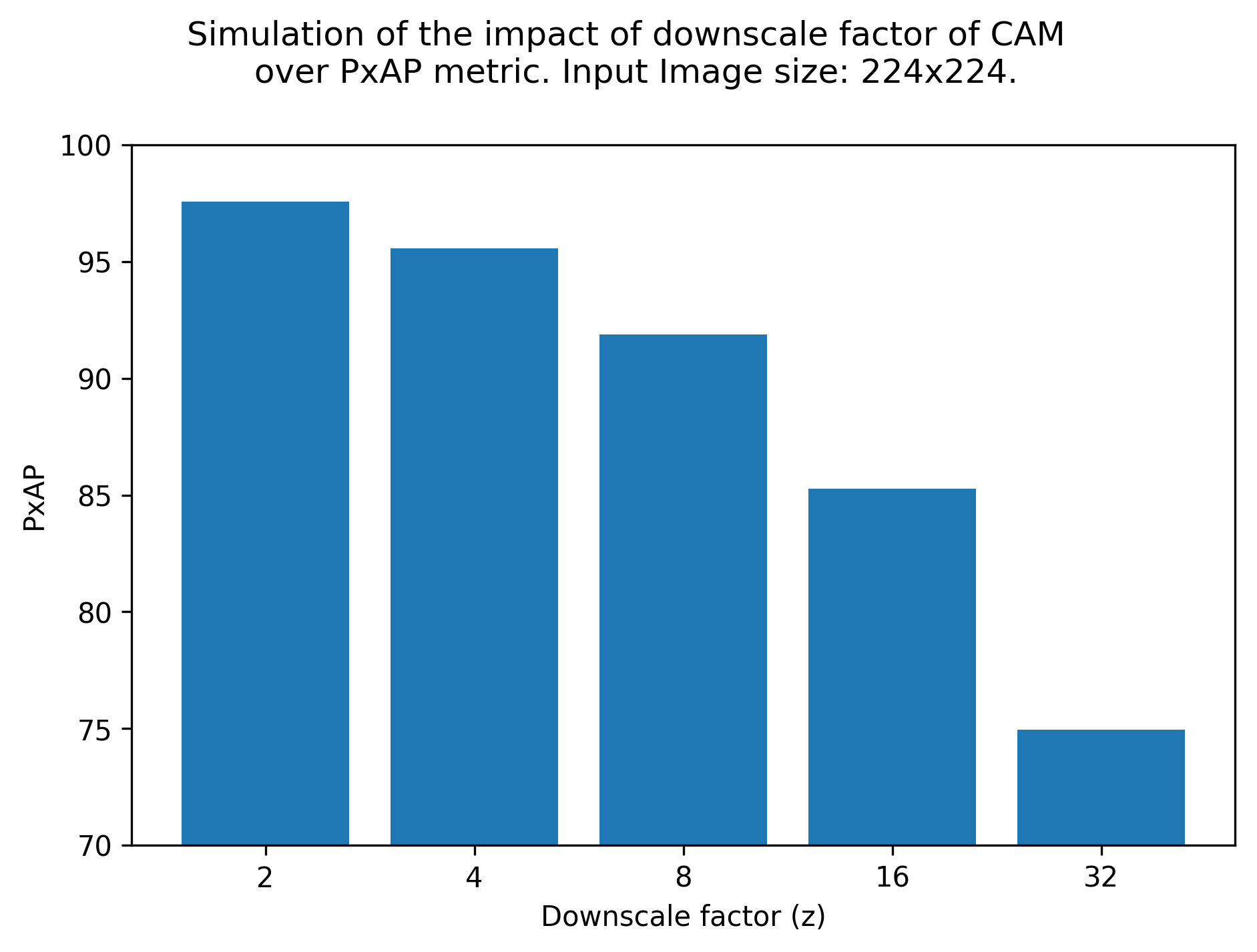}
  \caption{The localization performance over the test set of OpenImages dataset using \pxap metric with respect to the downscale factor $z$ of the simulated CAM. The simulation process is presented in Fig.\ref{fig:simulation-impact-scale}.}
  \label{fig:impact-scale}
\end{figure}

\section{Simulation of the impact of downscale factor of CAM over \pxap performance}
\label{sec:sim-impact-factor}
\begin{figure*}[ht!]
\centering
  \centering
  \includegraphics[width=.8\linewidth]{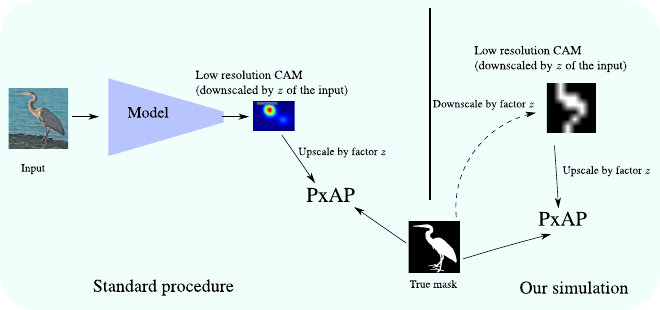}
  \caption{Simulation to evaluate the impact of CAM size over the localization performance using \pxap metric. In our simulation, we substitute the low resolution CAM predicted by the model by downscaled version of the true mask by a factor $z$ allowing us to quickly change the scale and assess its impact on the localization performance using \pxap metric. The scaling procedure is performed using bilinear interpolation.}
  \label{fig:simulation-impact-scale}
\end{figure*}

The size of the CAM is influenced by many operations in the model including convolution, and pooling operations. Standard models such as Resnet family~\cite{heZRS16} downscales the CAM by a factor ${z}$ up to ${z=32}$. In order to assess the impact of such downscale factor over the localization performance obtained through a CAM, we propose a simple simulation. For this simulation, we consider the dataset OpenImages as it provides the pixel-wise annotation in order to evaluate the \pxap performance. Instead of working directly on a model, we substitute the model's predicted low resolution CAM by a downsampled, with factor $z$, version of the true mask. Then, we measure the \pxap performance between the upscaled CAM and the true mask. This yields an almost perfect low resolution CAM but we assume it was predicted by the model. The procedure is illustrated in Fig.\ref{fig:simulation-impact-scale}. The results are presented in Fig.\ref{fig:impact-scale}. This figure shows that the low resolution size of CAMs is a major bottleneck in localization. Even when the low resolution CAM is directly the downscaled version of the true mask, upscaling back to the high resolution does not yield the exact mask due to loss of information when downscaling. \pxap values in Fig.\ref{fig:impact-scale} could be seen as an upper bound for pixel-wise localization in function of the scale factor. This implies that our method with full resolution can still yield better performance. These results suggest as well that for a better localization performance, it is better to use a low scale factor.

\section{Runtime analysis}
\label{sec:runtime}

We report in Tab.\ref{tab:complexity} the time required to build a CAM for all the studied WSOL finetuned baselines and other methods. While adding a decoder to a standard classifier increases the number of parameters, the inference time is still better than average finetuned baselines and competitive with other CAM methods. Because the inference is achieved through a single forward with fully convolutional operations, our method is expected to be fast. Other methods may require a forward and a backward to estimate a CAM. Our method is still competitive to ACoL, SPG, and ADL methods. We included other methods ScoreCAM~\cite{WangWDYZDMH20scorecam}, SSCAM~\cite{ZhangWKYH18}, and IS-CAM~\cite{naidu2020iscam}. Their slow runtime prevented us from considering them as baselines.

{
\setlength{\tabcolsep}{3pt}
\renewcommand{\arraystretch}{1.1}
\begin{table}[ht!]
\centering
\resizebox{.7\linewidth}{!}{%
\centering
\small
\begin{tabular}{lc*{1}{c}c*{1}{c}c*{1}{c}}
Backbone  &  &  \multicolumn{1}{c}{VGG} & & \multicolumn{1}{c}{Inception} && \multicolumn{1}{c}{ResNet}  \\
\cline{1-1}\cline{3-3}\cline{5-5} \cline{7-7} \\
Encoder & &    \\
\cline{1-1}\cline{3-3}\cline{5-5} \cline{7-7} \\
Layer 1  &&  128   &&  64  &&  64  \\
Layer 2  &&  256   &&  80   && 256  \\
Layer 3  &&  1024  && 288   &&  512 \\
Layer 4  &&  --    && 768   &&  1024 \\
Layer 5  &&  --    && 1024 &&   2048 \\
\cline{1-1}\cline{3-3}\cline{5-5} \cline{7-7} \\
Decoder & &    \\
\cline{1-1}\cline{3-3}\cline{5-5} \cline{7-7} \\
Layer 1  && 256    &&   256  &&  256  \\
Layer 2  && 128    &&  128   &&   128 \\
Layer 3  && 64    &&  64  &&    64\\
Layer 4  &&  --   && 32  &&    32\\
Layer 5  &&  --   && 16  &&    16\\
\end{tabular}
}
\caption{Architecture details. We use the same common backbones VGG16, InceptionV3, and Resnet50 without modification. Each backbone has its own definition of a layer and specific number of layers.}
\label{tab:arch-details}
\vspace{-1em}
\end{table}
}

{
\setlength{\tabcolsep}{3pt}
\renewcommand{\arraystretch}{1.1}
\begin{table}[ht!]
\centering
\resizebox{\linewidth}{!}{%
\centering
\small
\begin{tabular}{lc*{3}{c}g}
&  &  \multicolumn{4}{c}{CUB (\newmaxboxacc)}  \\
Methods  &  & VGG & Inception & ResNet & Mean \\
\cline{1-1}\cline{3-6}\\
CAM~\cite{zhou2016learning}{\small \emph{(cvpr,2016)}} &  &  63.7 & 56.7 & 63.0 & 61.1  \\
HaS~\cite{SinghL17}{\small \emph{(iccv,2017)}} &  &  63.7 & 53.4 & 64.7 & 60.6 \\
ACoL~\cite{ZhangWF0H18} {\small \emph{(cvpr,2018)}} &  &  57.4 & 56.2 & 66.5 & 60.0 \\
SPG~\cite{ZhangWKYH18}{\small \emph{(eccv,2018)}} &  &  56.3 & 55.9 & 60.4 & 57.5 \\
ADL~\cite{ChoeS19}  {\small \emph{(cvpr,2019)}} &  &  66.3 & 58.8 & 58.4 & 61.1 \\
CutMix~\cite{YunHCOYC19}{\small \emph{(eccv,2019)}} &  &  62.3 & 57.5 & 62.8 & 60.8 \\
\cline{1-1}\cline{3-6}\\
Best WSOL &  &  66.3 & 58.8 & 66.5 & 61.1 \\
FSL baseline &  &  71.6 & 86.6 & 82.4 & 80.2 \\
Center baseline &  &  59.7 & 59.7 & 59.7 & 59.7 \\
\cline{1-1}\cline{3-6}\\
ICL~\cite{KiU0B20icl}{\small \emph{(accv,2020)}} &  &  66.7 & 60.3 & 63.2 & 63.4 \\
\cline{1-1}\cline{3-6}\\
\cline{1-1}\cline{3-6} \\
CAM*~\cite{zhou2016learning} {\small \emph{(cvpr,2016)}}   &  &            57.0 & 54.4 & 62.1 & 57.8 \\
GradCAM~\cite{SelvarajuCDVPB17iccvgradcam} {\small \emph{(iccv,2017)}} &  &           62.7 & 57.1 & 63.3 & 61.0 \\
GradCAM++~\cite{ChattopadhyaySH18wacvgradcampp} {\small \emph{(wacv,2018)}}  &  &         73.8 & 60.7 & 70.2 & 68.2 \\
Smooth-GradCAM++~\cite{omeiza2019corr} {\small \emph{(corr,2019)}}  &  &  64.1 & 59.7 & 66.6 & 63.4 \\
XGradCAM~\cite{fu2020axiom} {\small \emph{(bmvc,2020)}}  &  &          62.8 & 56.7 & 63.2 & 60.9 \\
LayerCAM~\cite{JiangZHCW21layercam} {\small \emph{(ieee,2021)}}  &  &          74.1 & 62.6 & 72.6 & 69.7 \\
\cline{1-1}\cline{3-6}\\
\cline{1-1}\cline{3-6} \\
CAM*~\cite{zhou2016learning} + ours    &  &            79.1 & 71.2 & 79.4 & 76.5 \\
GradCAM~\cite{SelvarajuCDVPB17iccvgradcam} + ours  &  &           79.5 & 76.2 & 80.8 & 78.5 \\
GradCAM++~\cite{ChattopadhyaySH18wacvgradcampp} + ours &  &          84.1 & 73.1 & 82.7 & 79.9 \\
Smooth-GradCAM++~\cite{omeiza2019corr} + ours  &  &  83.1 & 74.0 & 81.6 & 79.5 \\
XGradCAM~\cite{fu2020axiom} + ours  &  &          80.1 & 70.6 & 80.0 & 76.9 \\
LayerCAM~\cite{JiangZHCW21layercam} + ours &  &           84.3 & 73.9 & 82.7 & 80.3 \\
\cline{1-1}\cline{3-6}\\
\cline{1-1}\cline{3-6} \\
Best WSOL + ours    &&         84.3 & 76.2 & 82.7 & 80.3 \\
\cline{1-1}\cline{3-6}\\
\end{tabular}
}
\caption{Performance on CUB using \newmaxboxacc metric.}
\label{tab:maxboxv2}
\vspace{-1em}
\end{table}
}

{
\setlength{\tabcolsep}{3pt}
\renewcommand{\arraystretch}{1.1}
\begin{table*}[ht!]
\centering
\resizebox{\linewidth}{!}{%
\centering
\small
\begin{tabular}{lc*{4}{c}c*{4}{c}c*{4}{c}}
Backbones (encoders) & & \multicolumn{4}{c}{VGG16} & & \multicolumn{4}{c}{Inception}  & & \multicolumn{4}{c}{ResNet50} \\
Methods  &  &       \#PCL & \#NFM & SFM & \#PDEC  &  & \#PCL & \#NFM & SFM & \#PDEC  &  &  \#PCL & \#NFM & SFM & \#PDEC  \\
\cline{1-1}\cline{3-6}\cline{8-11}\cline{13-16} \\
Details && $\approx$19.6 & 1024 & 28x28 & $\approx$23.1  && $\approx$25.6 & 1024 & 28x28 & $\approx$5.7 && $\approx$23.9 & 2048 & 28x28 & $\approx$9 \\
\cline{1-1}\cline{3-6}\cline{8-11}\cline{13-16} \\
CAM*~\cite{zhou2016learning} &  &                           \multicolumn{4}{c}{.2ms} && \multicolumn{4}{c}{.2ms} && \multicolumn{4}{c}{.3ms}\\
GradCAM~\cite{SelvarajuCDVPB17iccvgradcam} &  &             \multicolumn{4}{c}{7.7ms}  && \multicolumn{4}{c}{21.1ms} && \multicolumn{4}{c}{27.8ms} \\
GradCAM++~\cite{ChattopadhyaySH18wacvgradcampp} &  &        \multicolumn{4}{c}{23.5ms}  && \multicolumn{4}{c}{23.7ms} && \multicolumn{4}{c}{28.0ms}\\
Smooth-GradCAM~\cite{omeiza2019corr} &  &                   \multicolumn{4}{c}{62.0ms}  && \multicolumn{4}{c}{150.7ms } && \multicolumn{4}{c}{136.2ms} \\
XGradCAM~\cite{fu2020axiom} &  &                            \multicolumn{4}{c}{2.9ms} && \multicolumn{4}{c}{19.2ms}  && \multicolumn{4}{c}{14.2ms}\\
LayerCAM~\cite{JiangZHCW21layercam}  &  &                   \multicolumn{4}{c}{3.2ms}  && \multicolumn{4}{c}{18.2ms}  && \multicolumn{4}{c}{17.9ms} \\
\cline{1-1}\cline{3-6}\cline{8-11}\cline{13-16} \\
Mean  &  &                                                  \multicolumn{4}{c}{16.6ms}  && \multicolumn{4}{c}{38.8ms}  && \multicolumn{4}{c}{37.4ms} \\
\cline{1-1}\cline{3-6}\cline{8-11}\cline{13-16} \\
\shortstack{ours + STDCL}  &  &         \multicolumn{4}{c}{6.2ms}  && \multicolumn{4}{c}{25.5ms}  && \multicolumn{4}{c}{18.5ms} \\
\cline{1-1}\cline{3-6}\cline{8-11}\cline{13-16} \\
ACoL~\cite{ZhangWF0H18}  &  &                   \multicolumn{4}{c}{12.0ms}  && \multicolumn{4}{c}{19.2ms}  && \multicolumn{4}{c}{24.9ms} \\
SPG~\cite{ZhangWKYH18}  &  &                   \multicolumn{4}{c}{11.0ms}  && \multicolumn{4}{c}{18ms}  && \multicolumn{4}{c}{23.9ms} \\
ADL~\cite{ChoeS19}  &  &                   \multicolumn{4}{c}{6.4ms}  && \multicolumn{4}{c}{16.0}  && \multicolumn{4}{c}{14.4ms} \\
\cline{1-1}\cline{3-6}\cline{8-11}\cline{13-16} \\
ScoreCAM~\cite{WangWDYZDMH20scorecam}  &  &         \multicolumn{4}{c}{1.9sec}  && \multicolumn{4}{c}{3.4sec}  && \multicolumn{4}{c}{9.3sec} \\
SSCAM~\cite{naidu2020sscam}  &&\multicolumn{4}{c}{1min45sec} && \multicolumn{4}{c}{2min16sec}  &&\multicolumn{4}{c}{5min49sec} \\
IS-CAM~\cite{naidu2020iscam}  &  &  \multicolumn{4}{c}{30.1sec}  && \multicolumn{4}{c}{39.0sec}  && \multicolumn{4}{c}{1min39sec} \\
\end{tabular}
}
\caption{
Time required to build CAMs of different WSOL methods.
\textbf{STDCL}: standard classifier = encoder (VGG16, Inception, ResNet50) + global average pooling.
\textbf{\#PCL} (millions): number of the parameters of the classifier.
\textbf{\#NFM}: number of the feature maps at the top layer.
\textbf{SFM}: size of the feature maps at the top layer.
\textbf{\#PDEC} (millions): number of the parameters of the decoder.
\textbf{Time}: time necessary top build a full size CAM over an idle Tesla P100 GPU for one random RGB image of size ${224\times224}$ with ${200}$ classes.
Methods SSCAM~\cite{naidu2020sscam} (${N=35, \sigma=2}$), IS-CAM~\cite{naidu2020iscam} (${N=10}$), IS-CAM~\cite{naidu2020iscam} (${N=10}$) are evaluated with batch size 32 with their original hyper-parameters (${N, \text{ and } \sigma}$).
}
\label{tab:complexity}
\vspace{-1em}
\end{table*}
}

{
\setlength{\tabcolsep}{3pt}
\renewcommand{\arraystretch}{1.1}
\begin{table*}[ht!]
\centering
\resizebox{.7\textwidth}{!}{
\centering
\small
\begin{tabular}{lc*{3}{c}gc*{3}{c}g}
& &  \multicolumn{4}{c}{CUB}  & & \multicolumn{4}{c}{OpenImages}\\
Methods  &  &  VGG & Inception & ResNet & Mean &  & VGG & Inception & ResNet & Mean\\
\cline{1-1}\cline{3-6}\cline{8-11} \\
CAM~\cite{zhou2016learning} &  & 26.8 & 61.8 & 58.4 & 49.0 &  & 67.3 & 36.6 & 72.6 & 58.8 \\
HaS~\cite{SinghL17} &  & 70.9 & 69.9 & 74.5 & 71.8 &  & 60.0 & 68.4 & 74.0 & 67.5\\
ACoL~\cite{ZhangWF0H18} &  &  56.1 & 71.6 & 64.0 & 63.9 &  & 68.2 & 40.7 & 70.7 & 59.9\\
SPG~\cite{ZhangWKYH18} &  &  63.1 & 58.8 & 37.8 & 53.2 &  & 71.7 & 43.5 & 65.4 & 60.2 \\
ADL~\cite{ChoeS19} &  &  31.1 & 45.5 & 32.7 & 36.4 &  & 66.1 & 46.6 & 56.1 & 56.3 \\
CutMix~\cite{YunHCOYC19} &  &  29.2 & 70.2 & 55.9 & 51.8 &  & 68.1 & 53.1 & 73.7 & 65.0 \\
\cline{1-1}\cline{3-6}\cline{8-11}\\
\cline{1-1}\cline{3-6}\cline{8-11}\\
CAM* \cite{zhou2016learning} &  & 49.3 & 65.5 & 65.1 & 59.9 &  & 69.1 & 61.2 & 73.2 & 67.8 \\
GradCAM  &  &                     24.8 & 65.4 & 42.2 & 44.1 &  & 69.0 & 54.0 & 72.4 & 65.1 \\
GradCAM++ &  &                    24.8 & 65.2 & 42.2 & 44.0 &  & 70.3 & 69.6 & 72.3 & 70.7 \\
Smooth-GradCAM  &  &              24.8 & 65.3 & 42.2 & 44.1 &  & 69.0 & 54.0 & 67.0 & 63.3 \\
XGradCAM &  &                     24.8 & 65.4 & 65.1 & 51.7 &  & 69.0 & 69.4 & 72.0 & 70.1 \\
LayerCAM  &  &                    24.8 & 65.0 & 51.2 & 47.0 &  & 70.3 & 53.9 & 72.2 & 65.4 \\
\cline{1-1}\cline{3-6}\cline{8-11}\\
\end{tabular}
}
\caption{Classification performance of WSOL baseline methods. Model selection is performed over localization performance \maxboxacc and \pxap.}
\label{tab:main_cls}
\end{table*}
}

\FloatBarrier

\begin{figure*}
     \centering
     \begin{subfigure}[b]{0.49\textwidth}
         \centering
         \includegraphics[scale=.17]{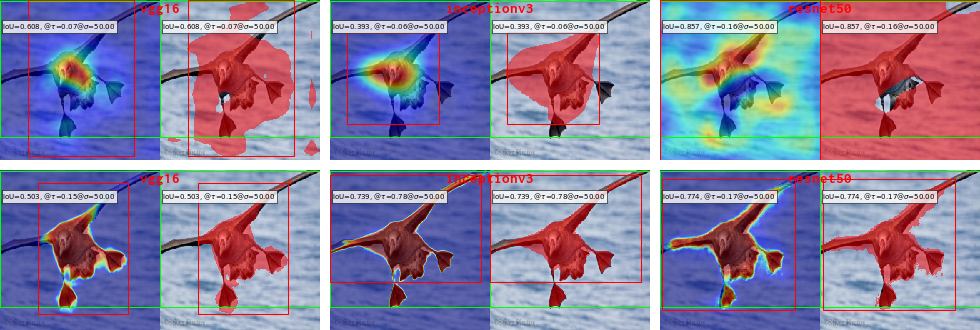}
     \end{subfigure}
     \hfill
     \begin{subfigure}[b]{0.49\textwidth}
         \centering
         \includegraphics[scale=.17]{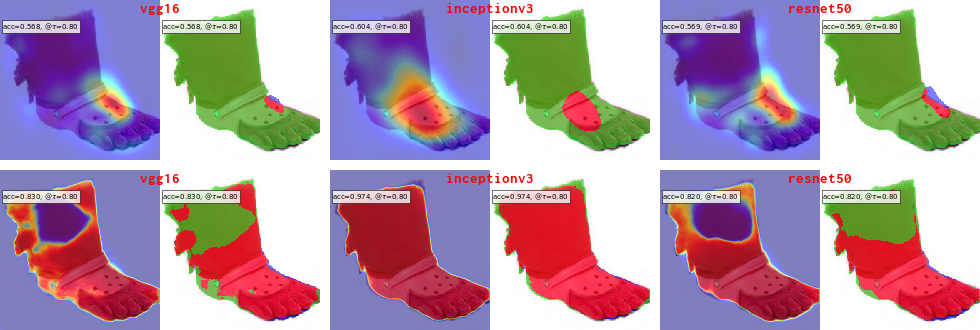}
     \end{subfigure}
     \\
     \vspace{0.1cm}
     \begin{subfigure}[b]{0.49\textwidth}
         \centering
         \includegraphics[scale=.17]{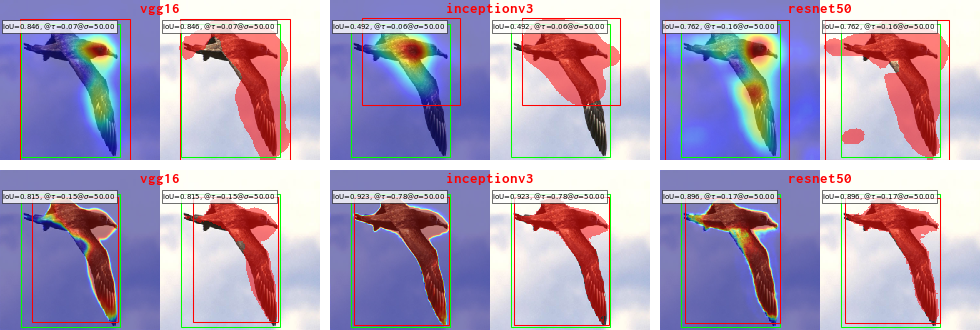}
     \end{subfigure}
     \hfill
     \begin{subfigure}[b]{0.49\textwidth}
         \centering
         \includegraphics[scale=.17]{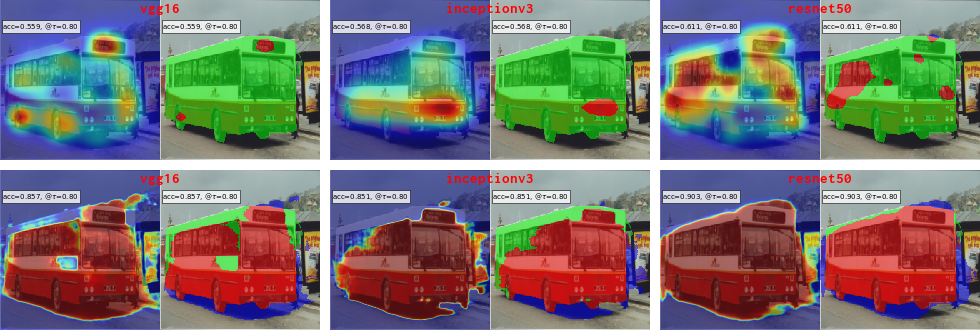}
     \end{subfigure}
     \\
     \vspace{0.1cm}
     \begin{subfigure}[b]{0.49\textwidth}
         \centering
         \includegraphics[scale=.17]{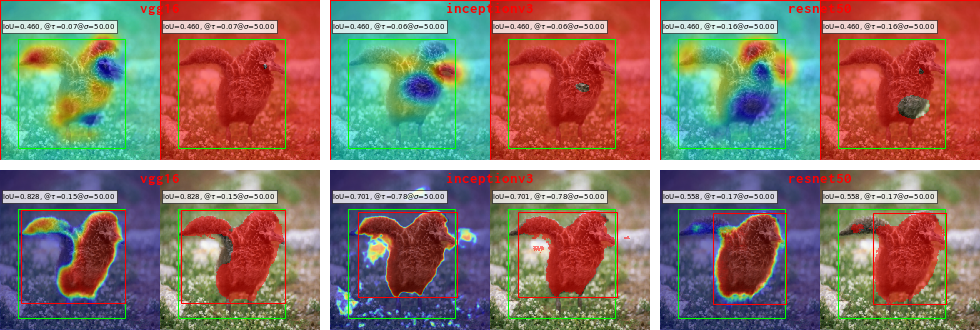}
     \end{subfigure}
     \hfill
     \begin{subfigure}[b]{0.49\textwidth}
         \centering
         \includegraphics[scale=.17]{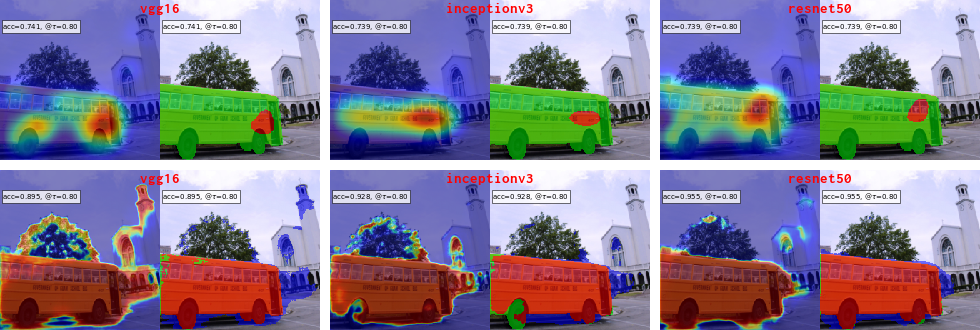}
     \end{subfigure}
     \\
     \vspace{0.1cm}
     \begin{subfigure}[b]{0.49\textwidth}
         \centering
         \includegraphics[scale=.17]{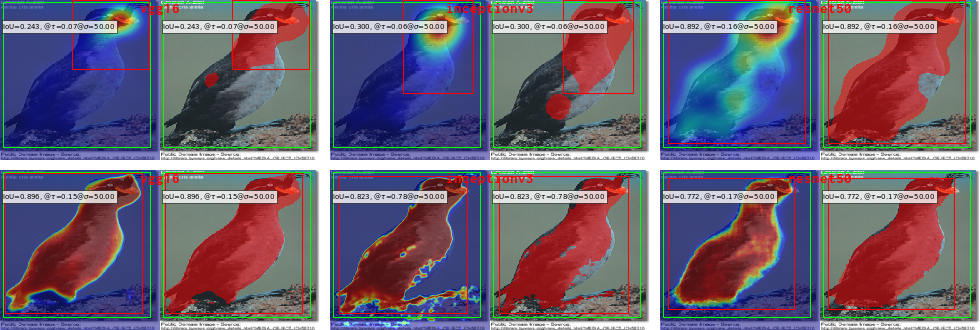}
     \end{subfigure}
     \hfill
     \begin{subfigure}[b]{0.49\textwidth}
         \centering
         \includegraphics[scale=.17]{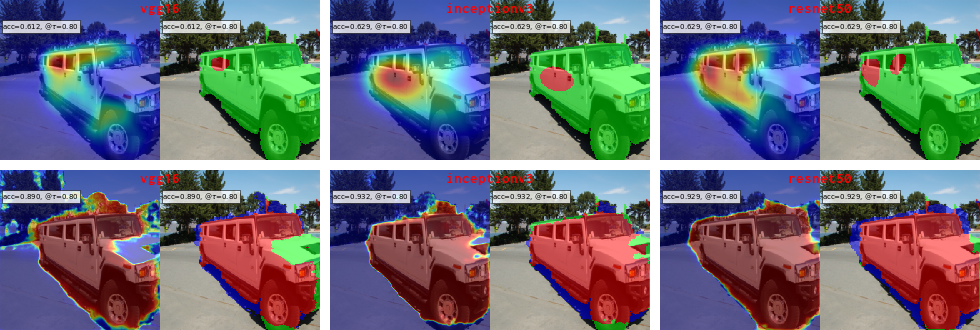}
     \end{subfigure}
     \\
     \vspace{0.1cm}
     \begin{subfigure}[b]{0.49\textwidth}
         \centering
         \includegraphics[scale=.17]{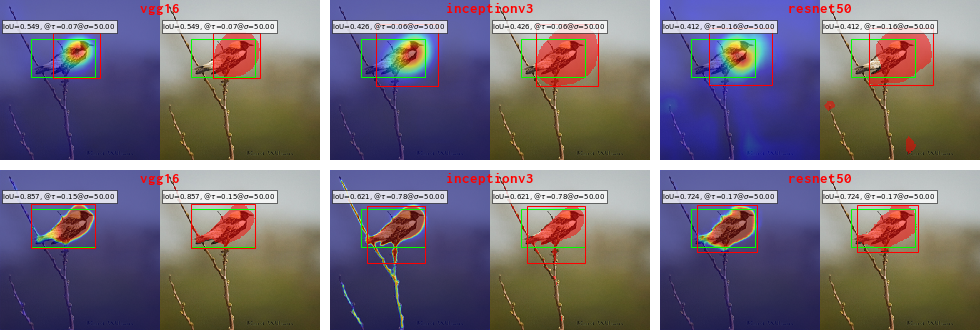}
     \end{subfigure}
     \hfill
     \begin{subfigure}[b]{0.49\textwidth}
         \centering
         \includegraphics[scale=.17]{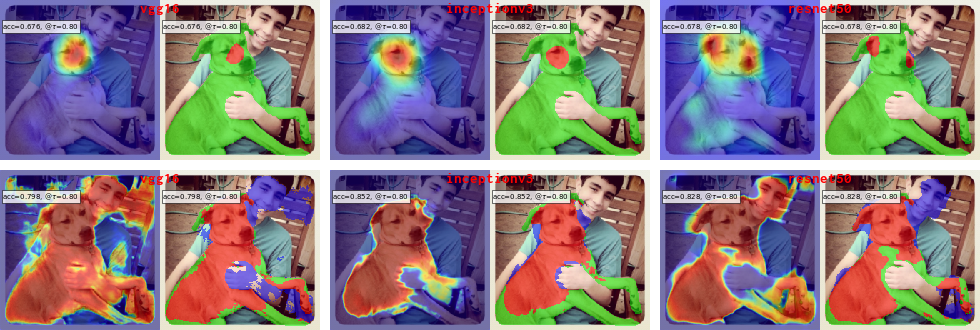}
     \end{subfigure}
     \\
     \vspace{0.1cm}
     \begin{subfigure}[b]{0.49\textwidth}
         \centering
         \includegraphics[scale=.17]{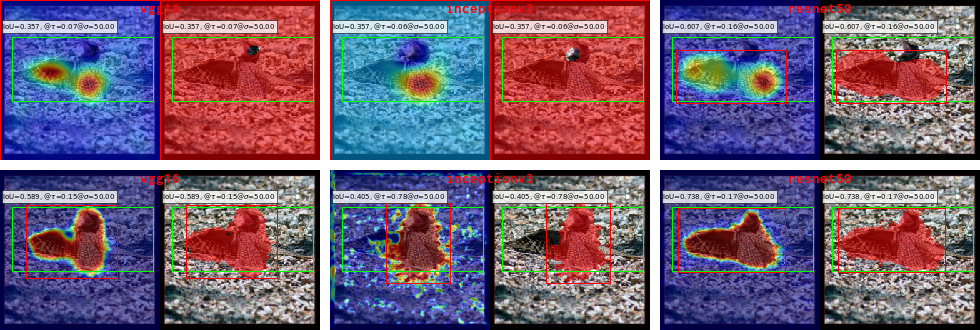}
     \end{subfigure}
     \hfill
     \begin{subfigure}[b]{0.49\textwidth}
         \centering
         \includegraphics[scale=.17]{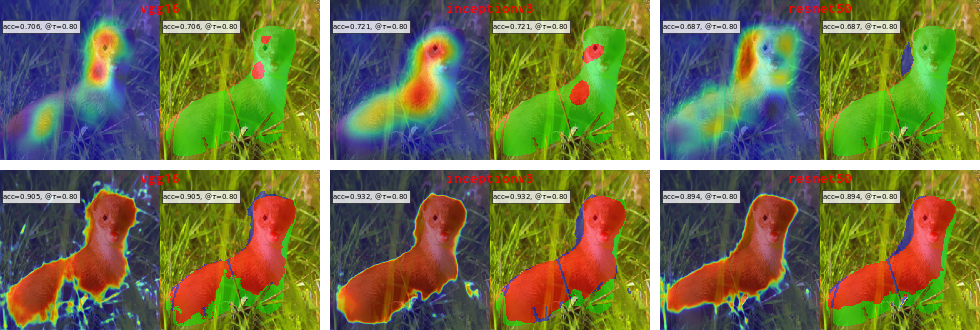}
     \end{subfigure}
      \\
     \vspace{0.1cm}
     \begin{subfigure}[b]{0.49\textwidth}
         \centering
         \includegraphics[scale=.17]{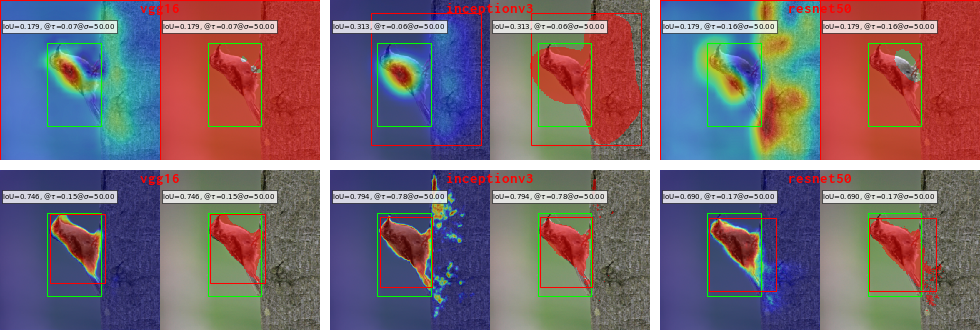}
     \end{subfigure}
     \hfill
     \begin{subfigure}[b]{0.49\textwidth}
         \centering
         \includegraphics[scale=.17]{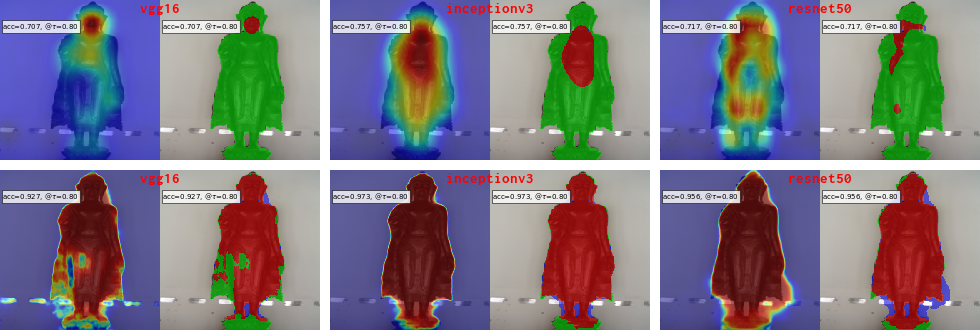}
     \end{subfigure}
      \\
     \vspace{0.1cm}
     \begin{subfigure}[b]{0.49\textwidth}
         \centering
         \includegraphics[scale=.17]{CUB-CAM-028-Brown_Creeper_Brown_Creeper_0124_24963}
     \end{subfigure}
     \hfill
     \begin{subfigure}[b]{0.49\textwidth}
         \centering
         \includegraphics[scale=.17]{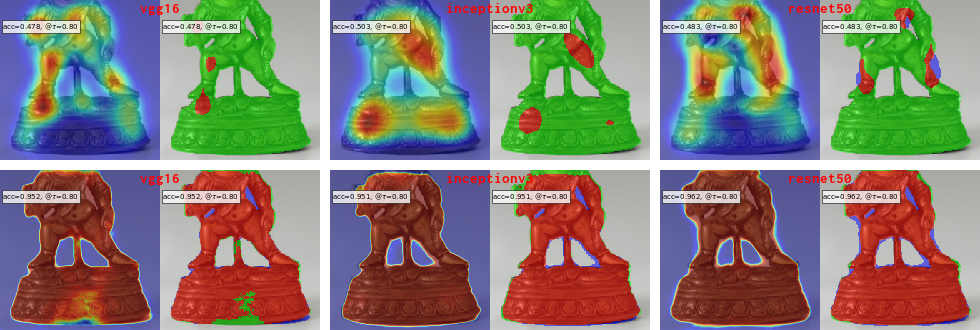}
     \end{subfigure}
      \\
     \vspace{0.1cm}
     \begin{subfigure}[b]{0.49\textwidth}
         \centering
         \includegraphics[scale=.17]{CUB-CAM-028-Brown_Creeper_Brown_Creeper_0124_24963}
     \end{subfigure}
     \hfill
     \begin{subfigure}[b]{0.49\textwidth}
         \centering
         \includegraphics[scale=.17]{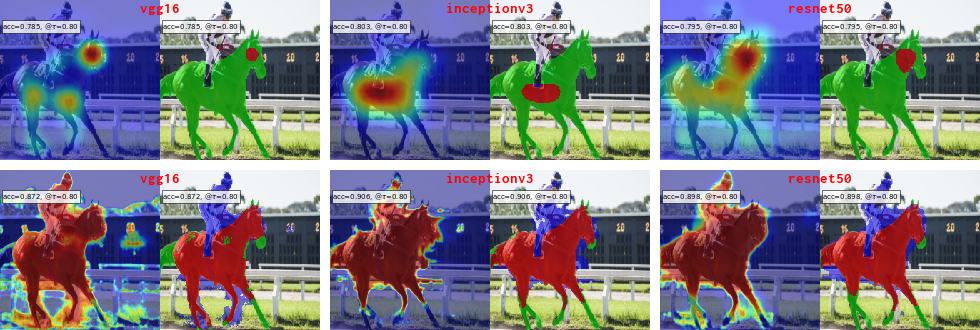}
     \end{subfigure}
     \\
     \vspace{0.1cm}
     \begin{subfigure}[b]{0.49\textwidth}
         \centering
         \includegraphics[scale=.17]{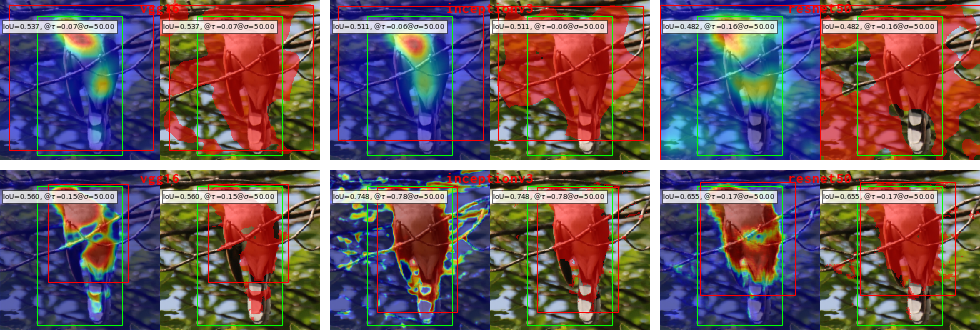}
     \end{subfigure}
     \hfill
     \begin{subfigure}[b]{0.49\textwidth}
         \centering
         \includegraphics[scale=.17]{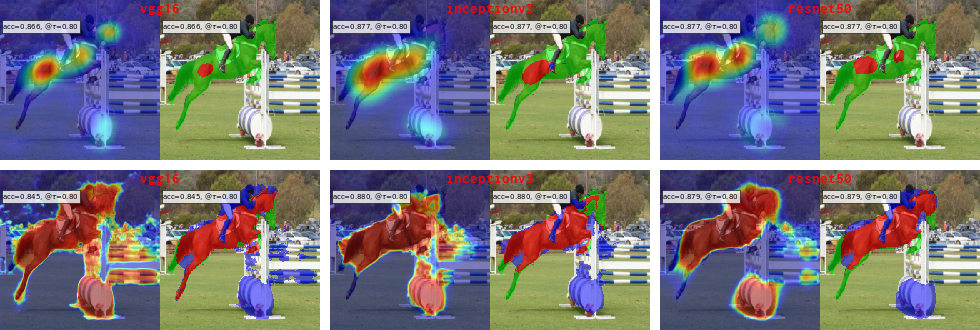}
     \end{subfigure}
        \caption{CAM* method examples for three backbones (left to right: VGG16, Inceptionv3, ResNet50): baselines (top) vs. baseline + ours (bottom)  validated with \maxboxacc. Colors: CUB (left): green box : ground truth. red box: predicted. red mask: thresholded CAM. OpenImages (right): red mask: true positive. green mask: false negative. blue mask: false positive. ${\tau=50, \sigma=0.8}$.}
        \label{fig:cam-cub-openim-example-pred}
\end{figure*}

\begin{figure*}
     \centering
     \begin{subfigure}[b]{0.49\textwidth}
         \centering
         \includegraphics[scale=.17]{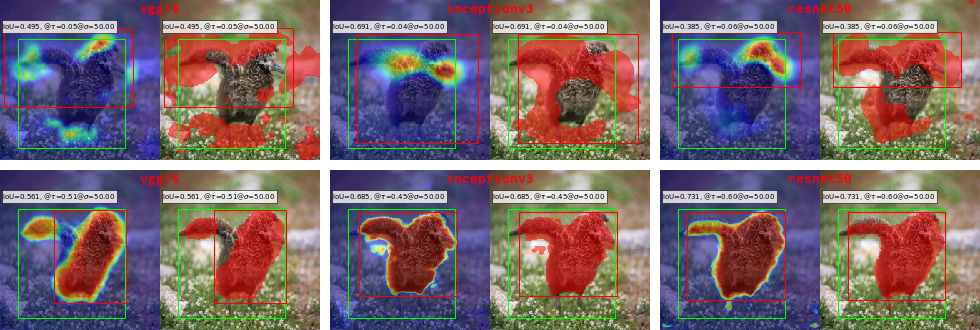}
     \end{subfigure}
     \hfill
     \begin{subfigure}[b]{0.49\textwidth}
         \centering
         \includegraphics[scale=.17]{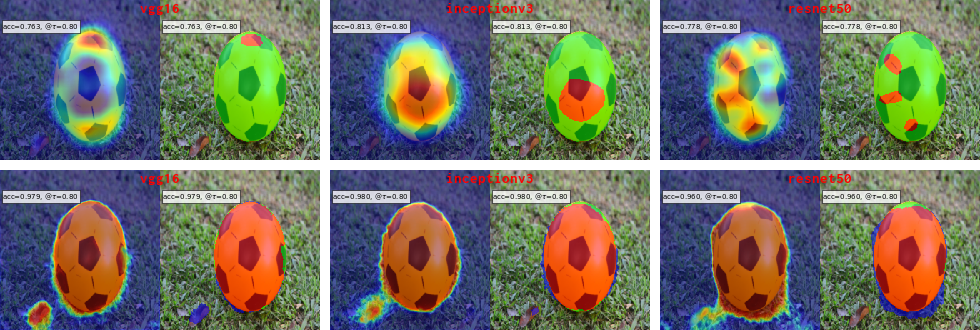}
     \end{subfigure}
     \\
     \vspace{0.1cm}
     \begin{subfigure}[b]{0.49\textwidth}
         \centering
         \includegraphics[scale=.17]{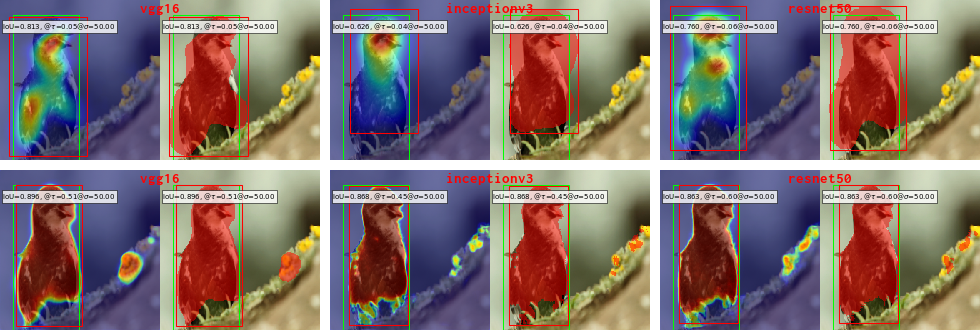}
     \end{subfigure}
     \hfill
     \begin{subfigure}[b]{0.49\textwidth}
         \centering
         \includegraphics[scale=.17]{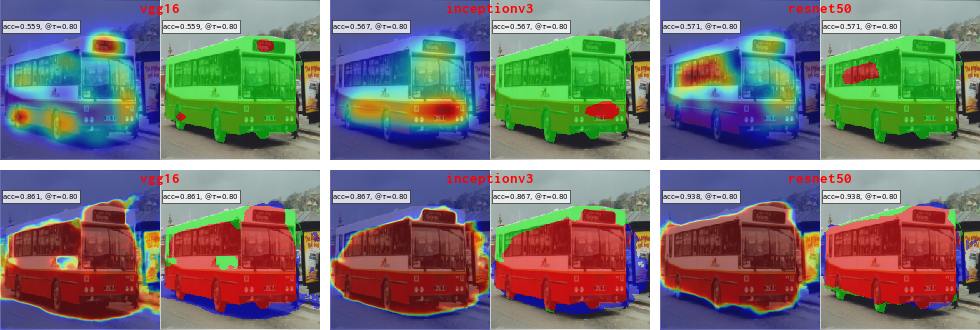}
     \end{subfigure}
     \\
     \vspace{0.1cm}
     \begin{subfigure}[b]{0.49\textwidth}
         \centering
         \includegraphics[scale=.17]{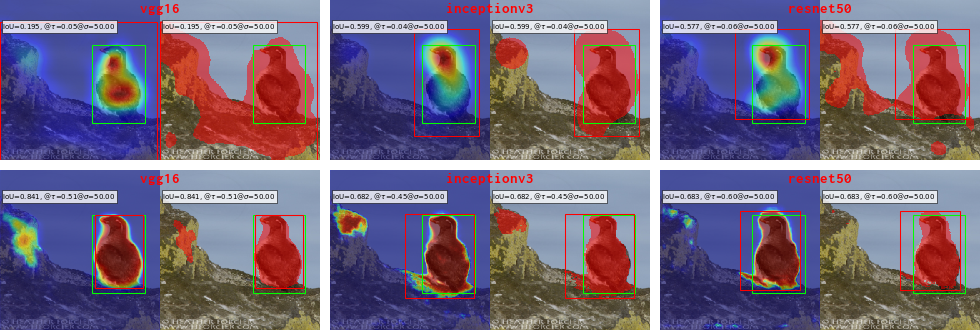}
     \end{subfigure}
     \hfill
     \begin{subfigure}[b]{0.49\textwidth}
         \centering
         \includegraphics[scale=.17]{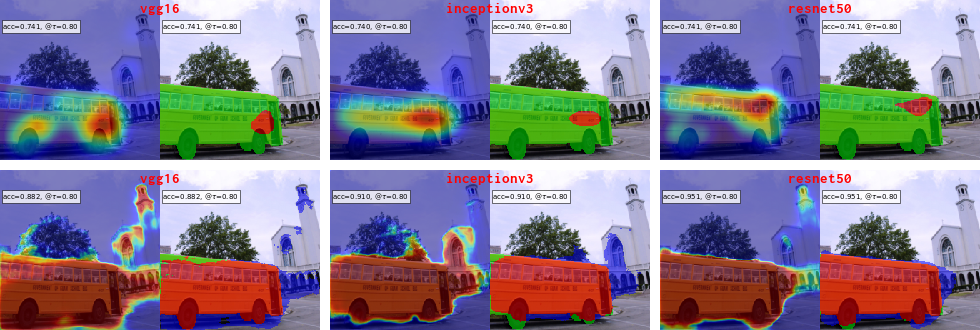}
     \end{subfigure}
     \\
     \vspace{0.1cm}
     \begin{subfigure}[b]{0.49\textwidth}
         \centering
         \includegraphics[scale=.17]{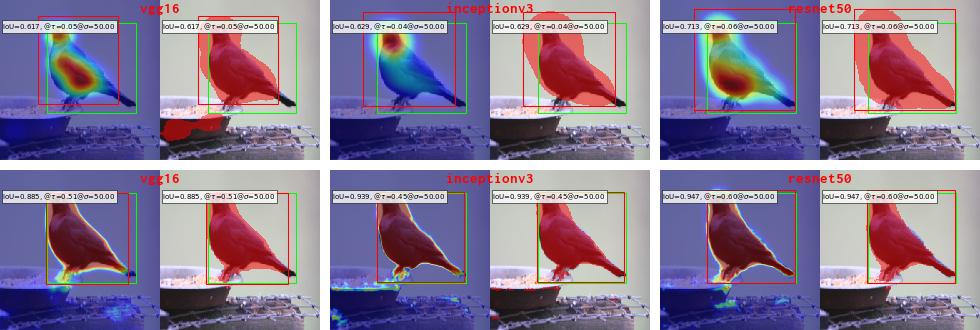}
     \end{subfigure}
     \hfill
     \begin{subfigure}[b]{0.49\textwidth}
         \centering
         \includegraphics[scale=.17]{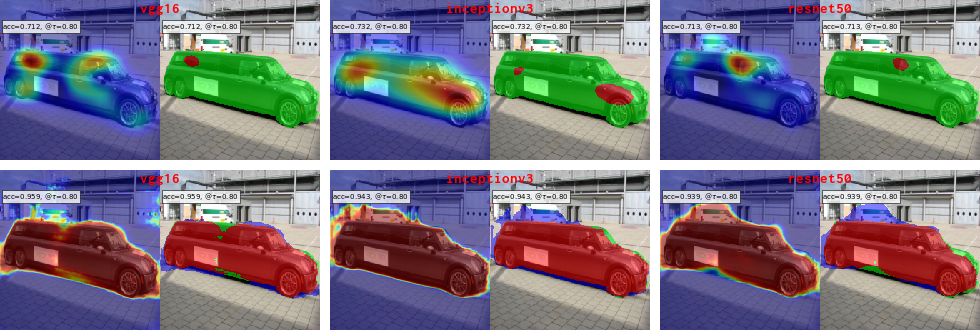}
     \end{subfigure}
     \\
     \vspace{0.1cm}
     \begin{subfigure}[b]{0.49\textwidth}
         \centering
         \includegraphics[scale=.17]{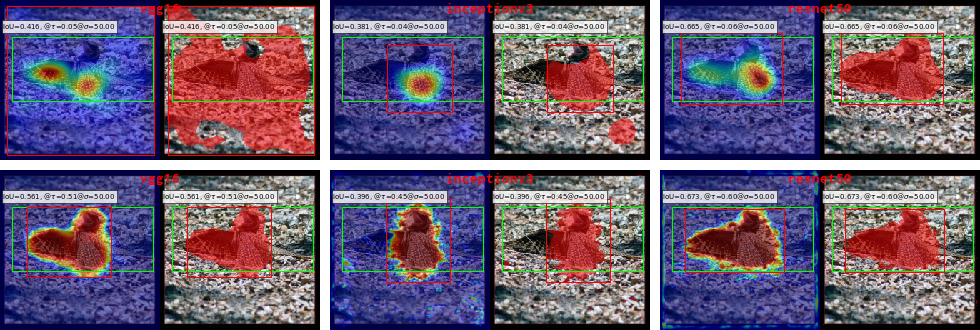}
     \end{subfigure}
     \hfill
     \begin{subfigure}[b]{0.49\textwidth}
         \centering
         \includegraphics[scale=.17]{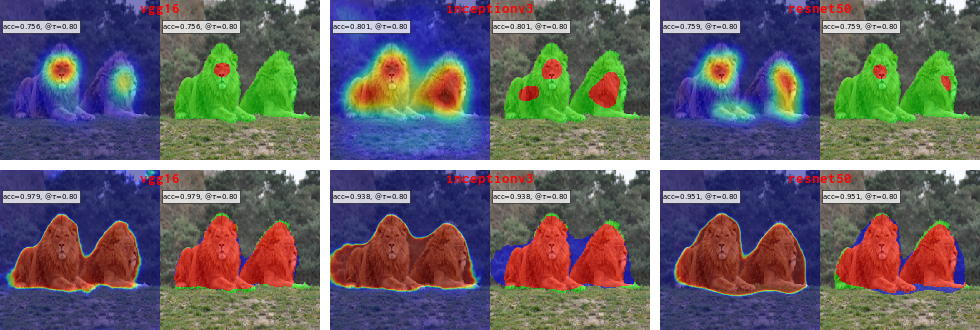}
     \end{subfigure}
     \\
     \vspace{0.1cm}
     \begin{subfigure}[b]{0.49\textwidth}
         \centering
         \includegraphics[scale=.17]{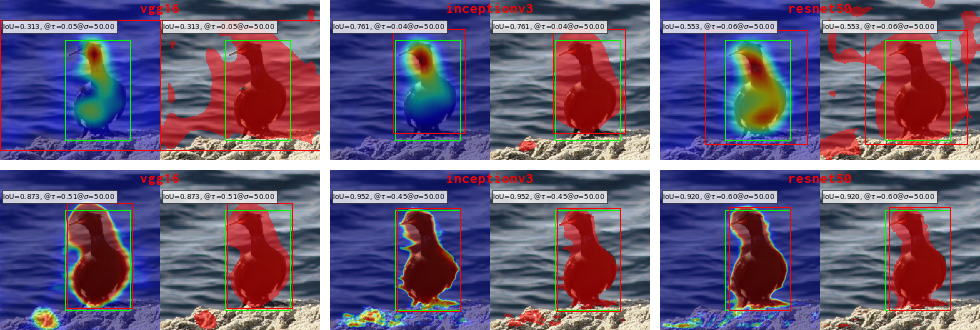}
     \end{subfigure}
     \hfill
     \begin{subfigure}[b]{0.49\textwidth}
         \centering
         \includegraphics[scale=.17]{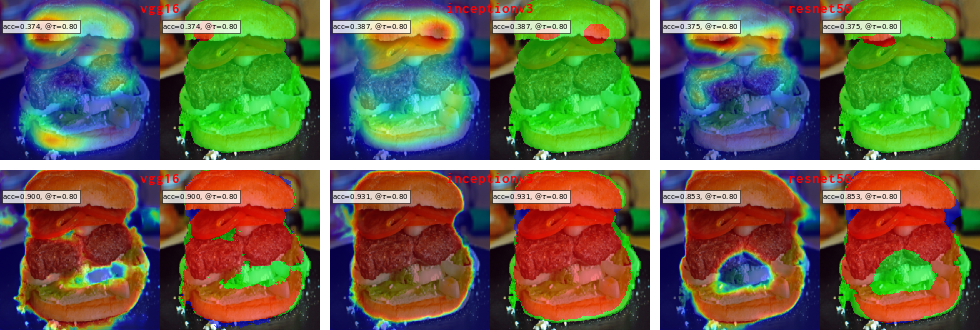}
     \end{subfigure}
      \\
     \vspace{0.1cm}
     \begin{subfigure}[b]{0.49\textwidth}
         \centering
         \includegraphics[scale=.17]{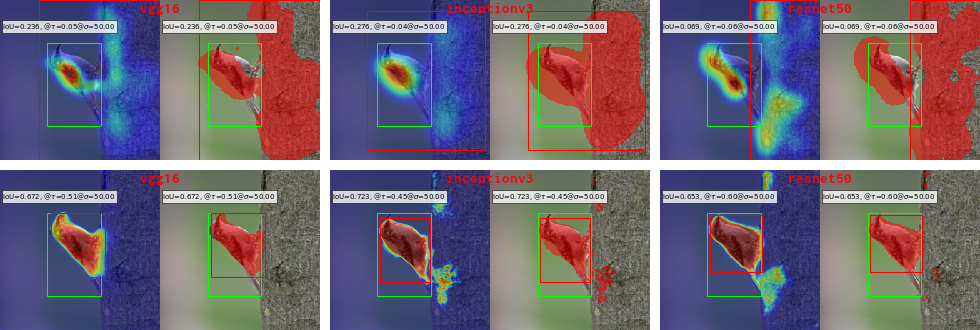}
     \end{subfigure}
     \hfill
     \begin{subfigure}[b]{0.49\textwidth}
         \centering
         \includegraphics[scale=.17]{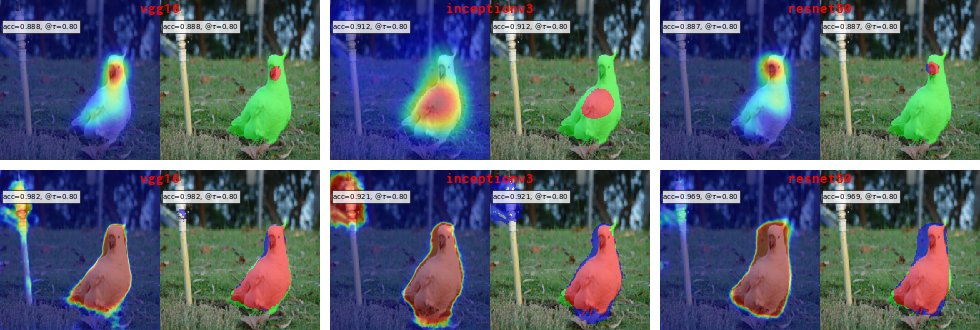}
     \end{subfigure}
      \\
     \vspace{0.1cm}
     \begin{subfigure}[b]{0.49\textwidth}
         \centering
         \includegraphics[scale=.17]{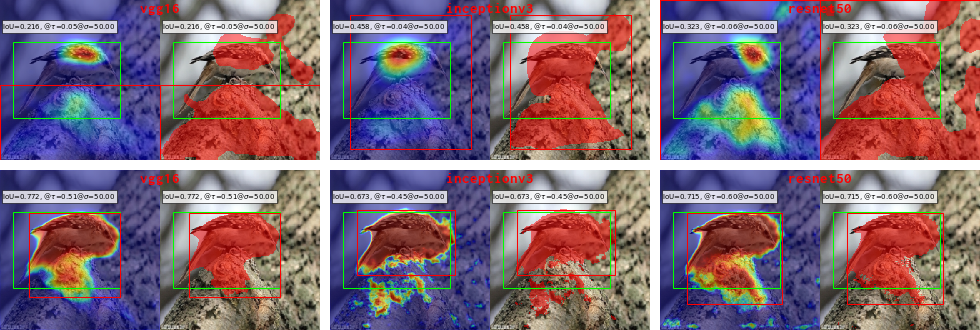}
     \end{subfigure}
     \hfill
     \begin{subfigure}[b]{0.49\textwidth}
         \centering
         \includegraphics[scale=.17]{OpenImages-GradCam-test_0gv1x_9ea27c73dc7d75d9}
     \end{subfigure}
      \\
     \vspace{0.1cm}
     \begin{subfigure}[b]{0.49\textwidth}
         \centering
         \includegraphics[scale=.17]{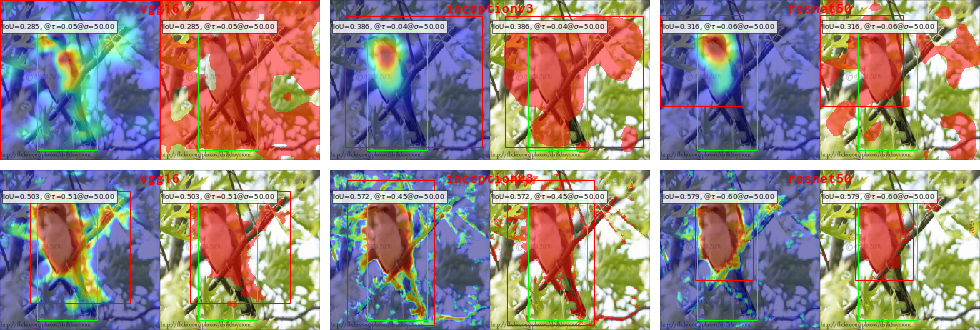}
     \end{subfigure}
     \hfill
     \begin{subfigure}[b]{0.49\textwidth}
         \centering
         \includegraphics[scale=.17]{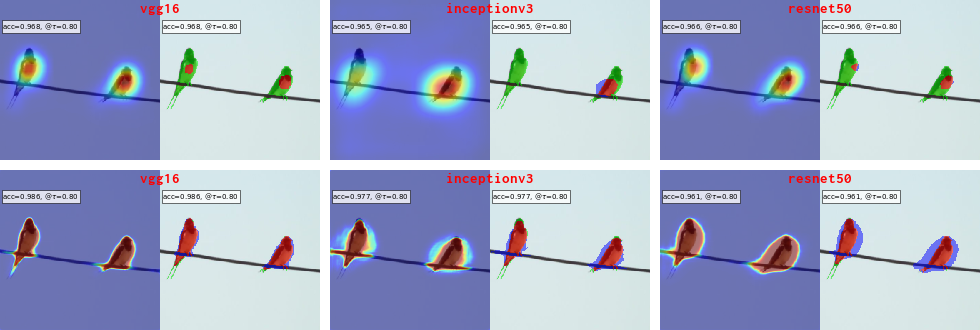}
     \end{subfigure}
     \\
     \vspace{0.1cm}
     \begin{subfigure}[b]{0.49\textwidth}
         \centering
         \includegraphics[scale=.17]{CUB-GradCam-042-Vermilion_Flycatcher_Vermilion_Flycatcher_0028_42197}
     \end{subfigure}
     \hfill
     \begin{subfigure}[b]{0.49\textwidth}
         \centering
         \includegraphics[scale=.17]{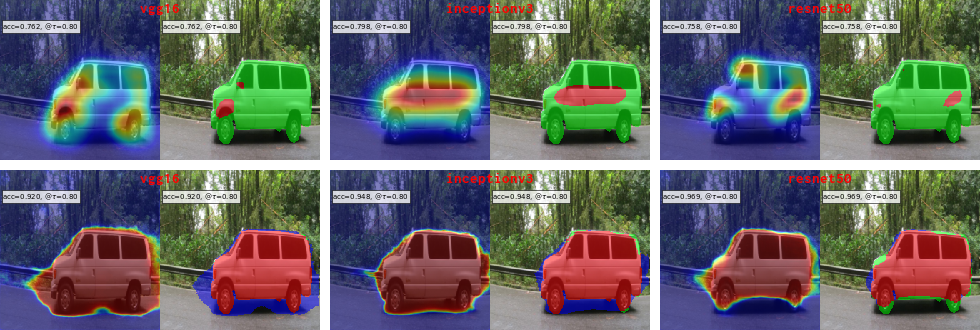}
     \end{subfigure}
        \caption{GradCAM method examples for three backbones (left to right: VGG16, Inceptionv3, ResNet50): baselines (top) vs. baseline + ours (bottom)  validated with \maxboxacc. Colors: CUB (left): green box : ground truth. red box: predicted. red mask: thresholded CAM. OpenImages (right): red mask: true positive. green mask: false negative. blue mask: false positive. ${\tau=50, \sigma=0.8}$.}
        \label{fig:gradcam-cub-openim-example-pred}
\end{figure*}

\begin{figure*}
     \centering
     \begin{subfigure}[b]{0.49\textwidth}
         \centering
         \includegraphics[scale=.17]{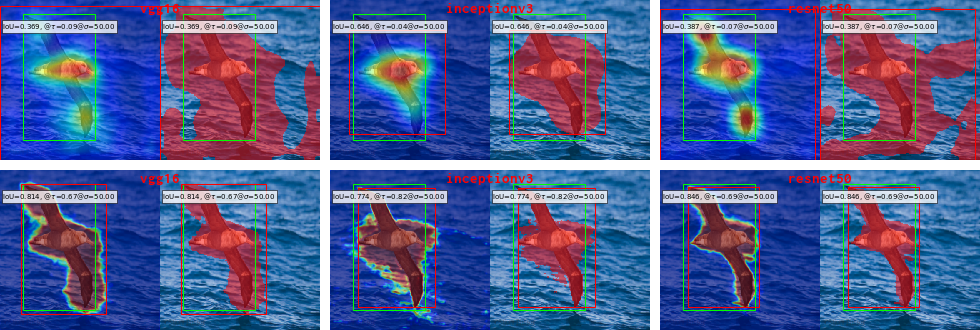}
     \end{subfigure}
     \hfill
     \begin{subfigure}[b]{0.49\textwidth}
         \centering
         \includegraphics[scale=.17]{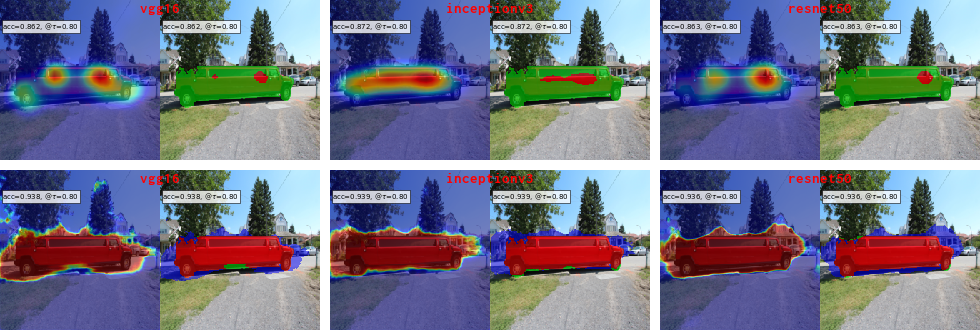}
     \end{subfigure}
     \\
     \vspace{0.1cm}
     \begin{subfigure}[b]{0.49\textwidth}
         \centering
         \includegraphics[scale=.17]{CUB-GradCAMpp-028-Brown_Creeper_Brown_Creeper_0006_25034}
     \end{subfigure}
     \hfill
     \begin{subfigure}[b]{0.49\textwidth}
         \centering
         \includegraphics[scale=.17]{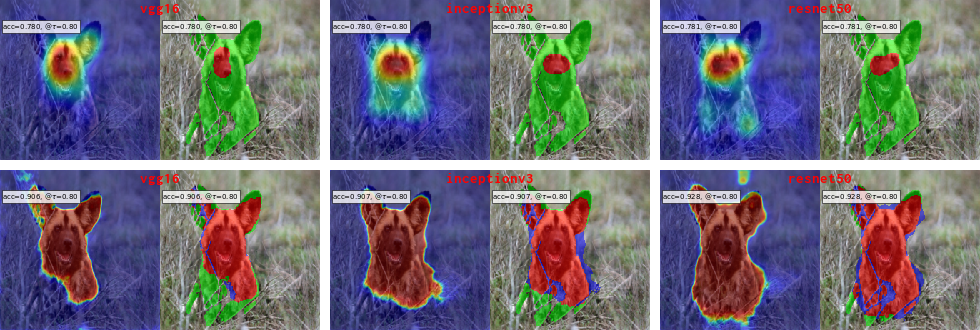}
     \end{subfigure}
     \\
     \vspace{0.1cm}
     \begin{subfigure}[b]{0.49\textwidth}
         \centering
         \includegraphics[scale=.17]{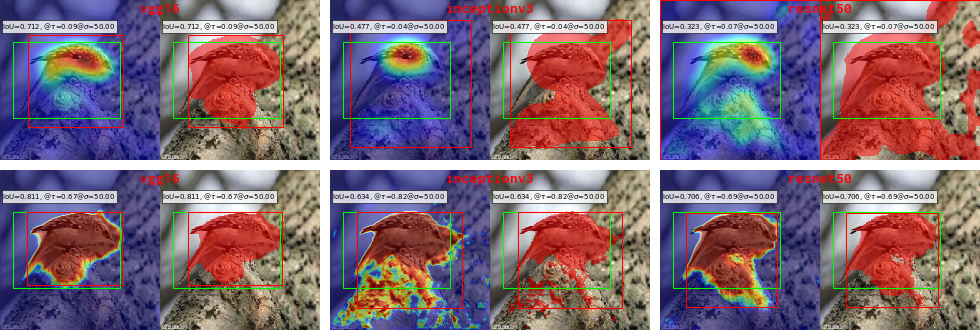}
     \end{subfigure}
     \hfill
     \begin{subfigure}[b]{0.49\textwidth}
         \centering
         \includegraphics[scale=.17]{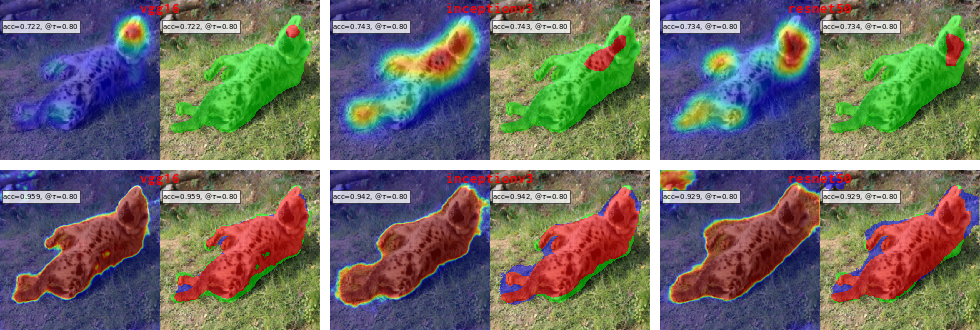}
     \end{subfigure}
     \\
     \vspace{0.1cm}
     \begin{subfigure}[b]{0.49\textwidth}
         \centering
         \includegraphics[scale=.17]{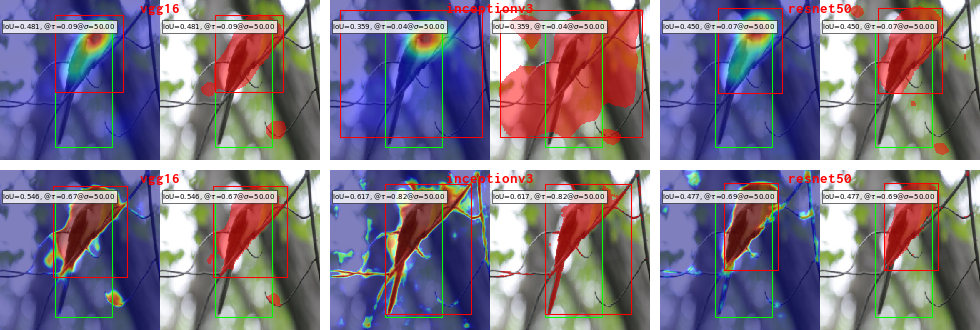}
     \end{subfigure}
     \hfill
     \begin{subfigure}[b]{0.49\textwidth}
         \centering
         \includegraphics[scale=.17]{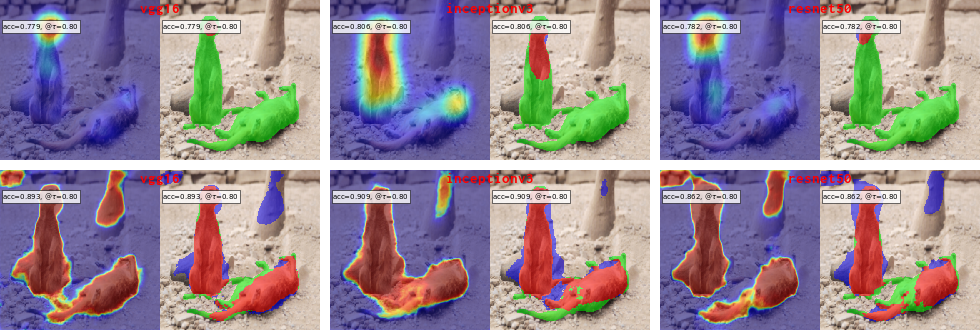}
     \end{subfigure}
     \\
     \vspace{0.1cm}
     \begin{subfigure}[b]{0.49\textwidth}
         \centering
         \includegraphics[scale=.17]{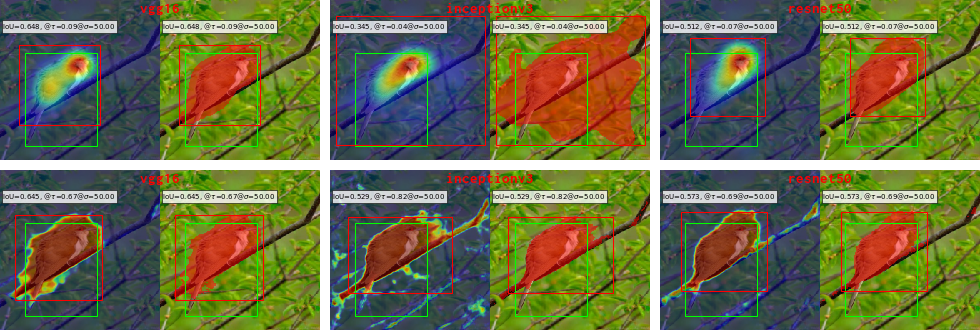}
     \end{subfigure}
     \hfill
     \begin{subfigure}[b]{0.49\textwidth}
         \centering
         \includegraphics[scale=.17]{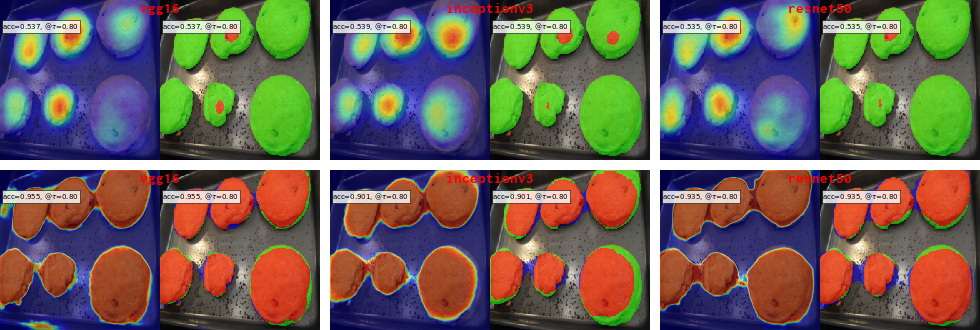}
     \end{subfigure}
     \\
     \vspace{0.1cm}
     \begin{subfigure}[b]{0.49\textwidth}
         \centering
         \includegraphics[scale=.17]{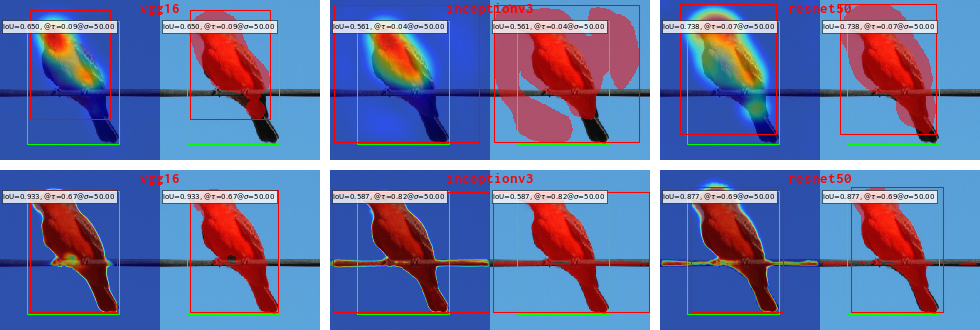}
     \end{subfigure}
     \hfill
     \begin{subfigure}[b]{0.49\textwidth}
         \centering
         \includegraphics[scale=.17]{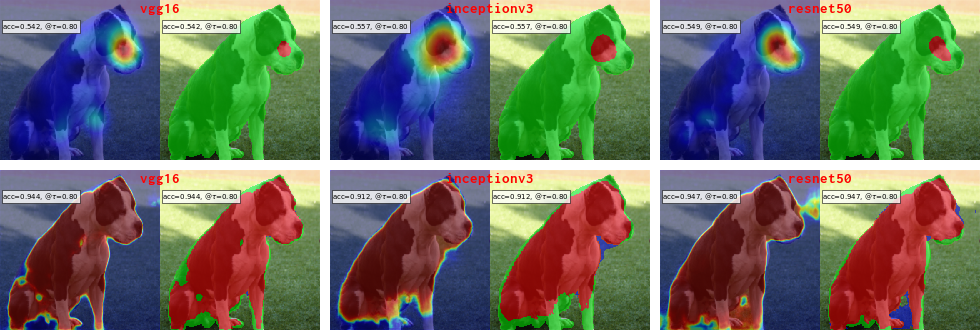}
     \end{subfigure}
      \\
     \vspace{0.1cm}
     \begin{subfigure}[b]{0.49\textwidth}
         \centering
         \includegraphics[scale=.17]{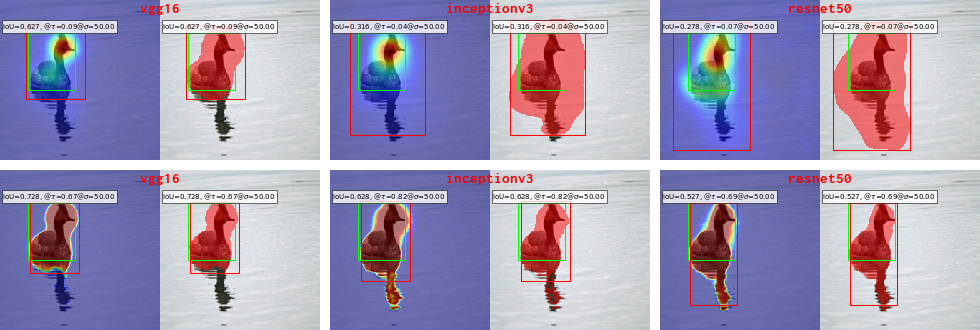}
     \end{subfigure}
     \hfill
     \begin{subfigure}[b]{0.49\textwidth}
         \centering
         \includegraphics[scale=.17]{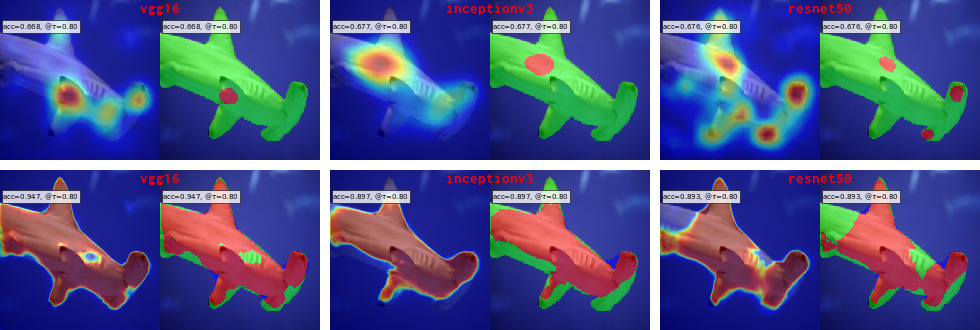}
     \end{subfigure}
      \\
     \vspace{0.1cm}
     \begin{subfigure}[b]{0.49\textwidth}
         \centering
         \includegraphics[scale=.17]{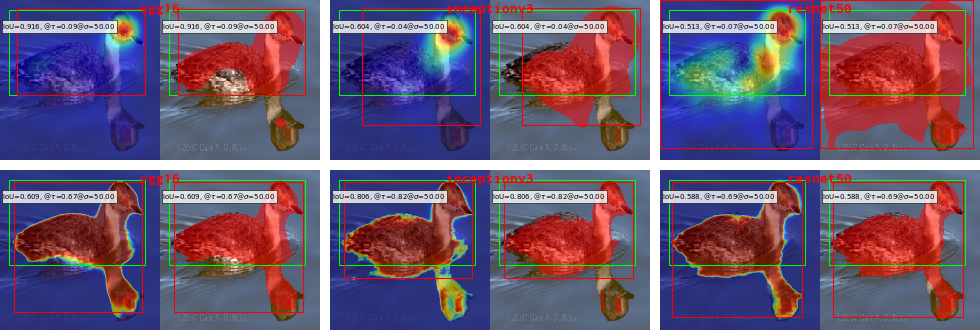}
     \end{subfigure}
     \hfill
     \begin{subfigure}[b]{0.49\textwidth}
         \centering
         \includegraphics[scale=.17]{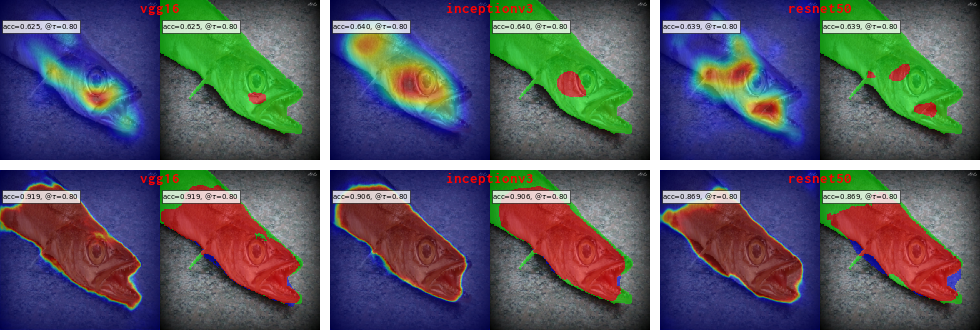}
     \end{subfigure}
      \\
     \vspace{0.1cm}
     \begin{subfigure}[b]{0.49\textwidth}
         \centering
         \includegraphics[scale=.17]{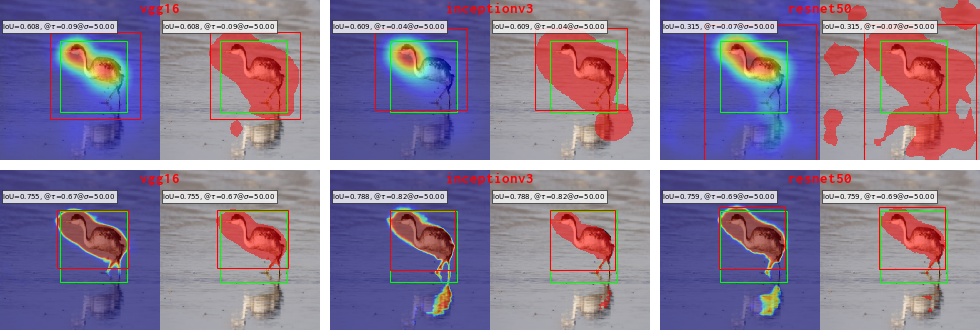}
     \end{subfigure}
     \hfill
     \begin{subfigure}[b]{0.49\textwidth}
         \centering
         \includegraphics[scale=.17]{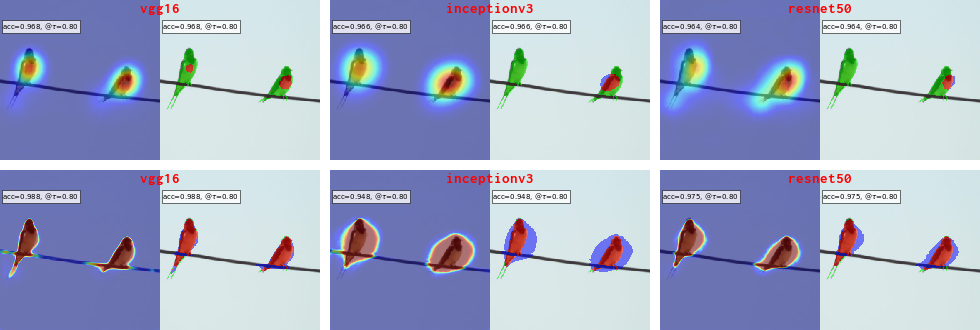}
     \end{subfigure}
     \\
     \vspace{0.1cm}
     \begin{subfigure}[b]{0.49\textwidth}
         \centering
         \includegraphics[scale=.17]{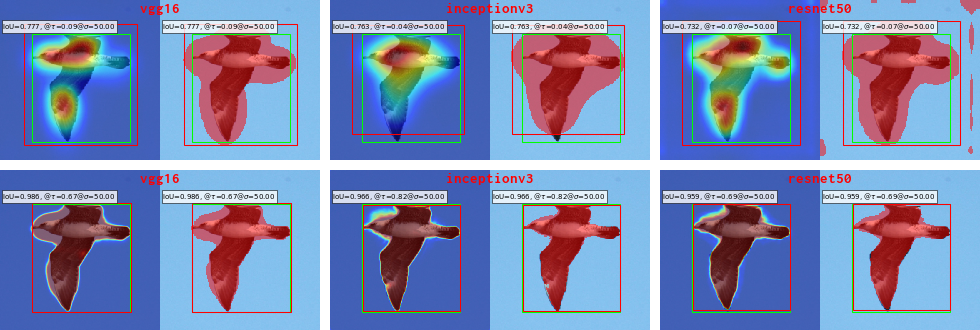}
     \end{subfigure}
     \hfill
     \begin{subfigure}[b]{0.49\textwidth}
         \centering
         \includegraphics[scale=.17]{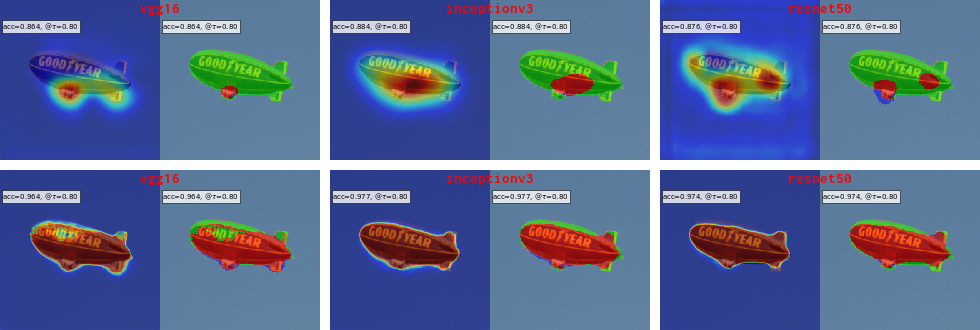}
     \end{subfigure}
        \caption{GradCAM++ method examples for three backbones (left to right: VGG16, Inceptionv3, ResNet50): baselines (top) vs. baseline + ours (bottom)  validated with \maxboxacc. Colors: CUB (left): green box : ground truth. red box: predicted. red mask: thresholded CAM. OpenImages (right): red mask: true positive. green mask: false negative. blue mask: false positive. ${\tau=50, \sigma=0.8}$.}
        \label{fig:gradcampp-cub-openim-example-pred}
\end{figure*}

\begin{figure*}
     \centering
     \begin{subfigure}[b]{0.49\textwidth}
         \centering
         \includegraphics[scale=.17]{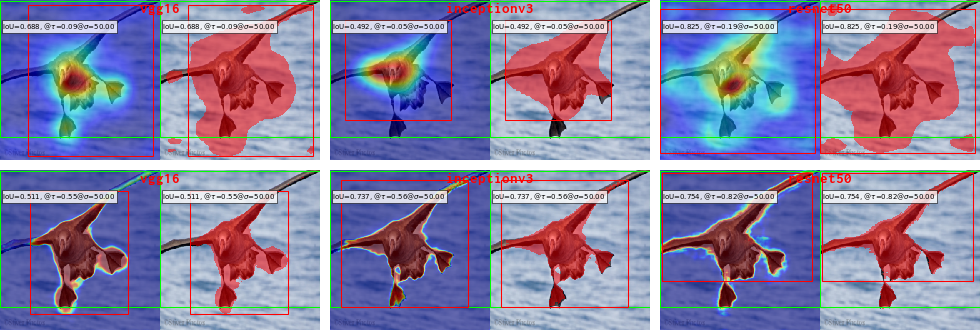}
     \end{subfigure}
     \hfill
     \begin{subfigure}[b]{0.49\textwidth}
         \centering
         \includegraphics[scale=.17]{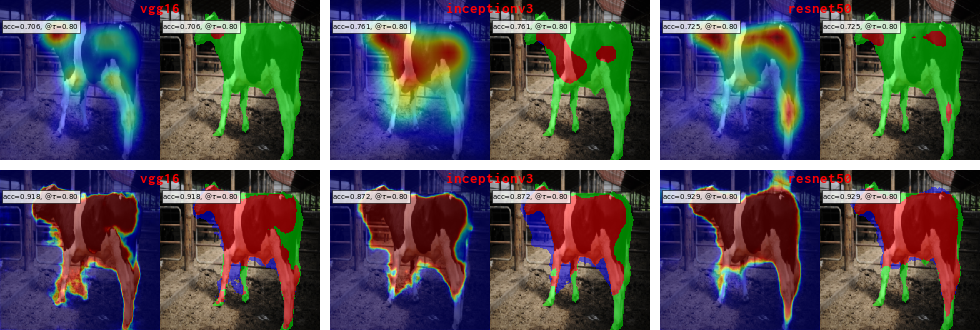}
     \end{subfigure}
     \\
     \vspace{0.1cm}
     \begin{subfigure}[b]{0.49\textwidth}
         \centering
         \includegraphics[scale=.17]{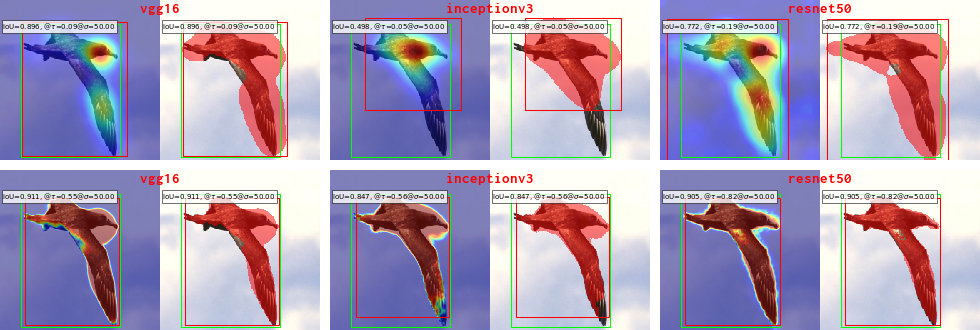}
     \end{subfigure}
     \hfill
     \begin{subfigure}[b]{0.49\textwidth}
         \centering
         \includegraphics[scale=.17]{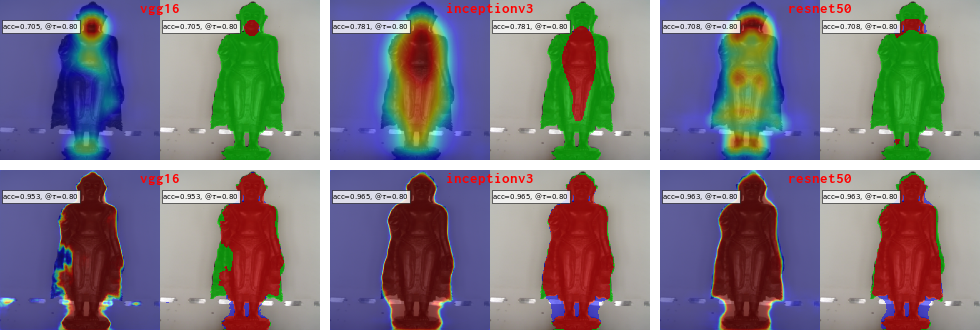}
     \end{subfigure}
     \\
     \vspace{0.1cm}
     \begin{subfigure}[b]{0.49\textwidth}
         \centering
         \includegraphics[scale=.17]{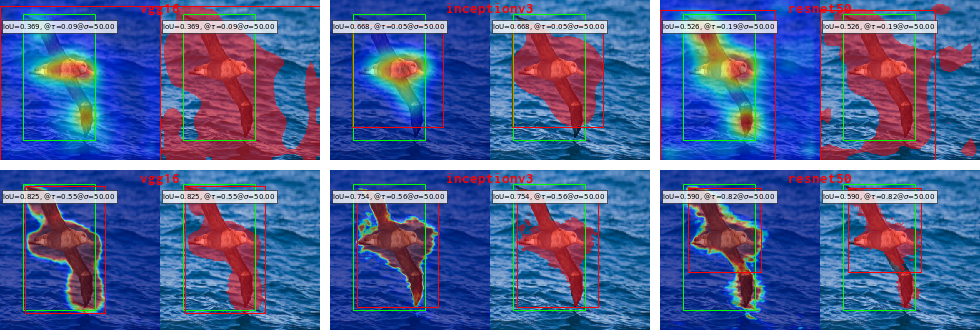}
     \end{subfigure}
     \hfill
     \begin{subfigure}[b]{0.49\textwidth}
         \centering
         \includegraphics[scale=.17]{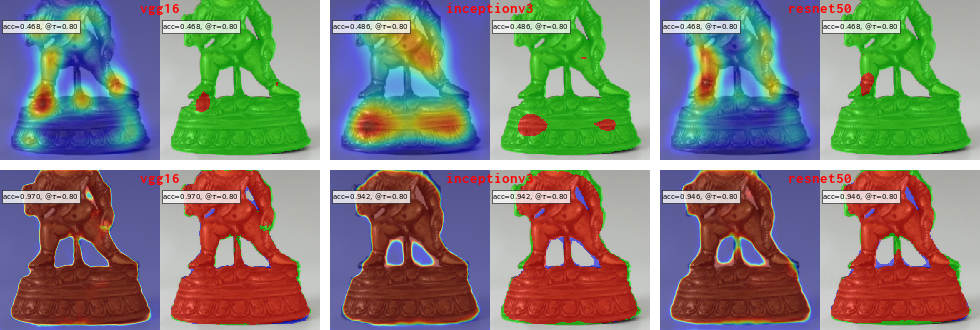}
     \end{subfigure}
     \\
     \vspace{0.1cm}
     \begin{subfigure}[b]{0.49\textwidth}
         \centering
         \includegraphics[scale=.17]{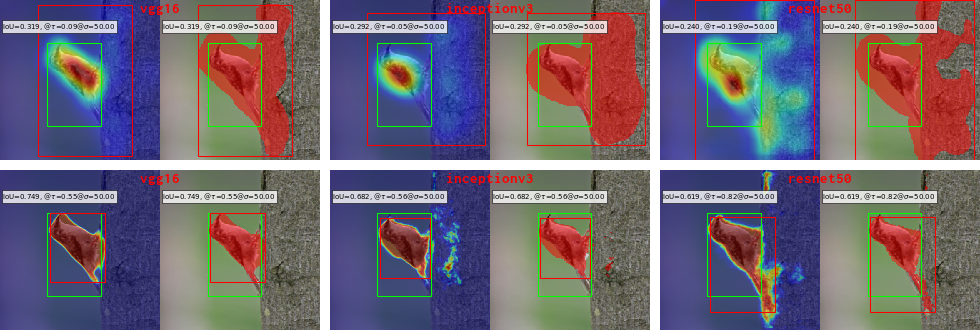}
     \end{subfigure}
     \hfill
     \begin{subfigure}[b]{0.49\textwidth}
         \centering
         \includegraphics[scale=.17]{OpenImages-LayerCAM-test_09f_2_bbff4ee30d6ac044}
     \end{subfigure}
     \\
     \vspace{0.1cm}
     \begin{subfigure}[b]{0.49\textwidth}
         \centering
         \includegraphics[scale=.17]{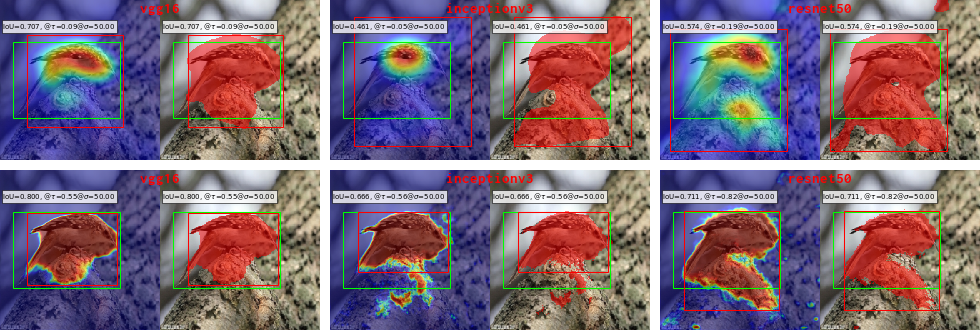}
     \end{subfigure}
     \hfill
     \begin{subfigure}[b]{0.49\textwidth}
         \centering
         \includegraphics[scale=.17]{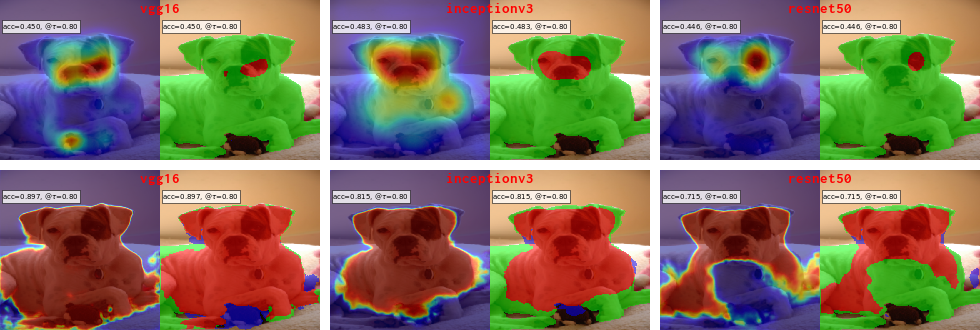}
     \end{subfigure}
     \\
     \vspace{0.1cm}
     \begin{subfigure}[b]{0.49\textwidth}
         \centering
         \includegraphics[scale=.17]{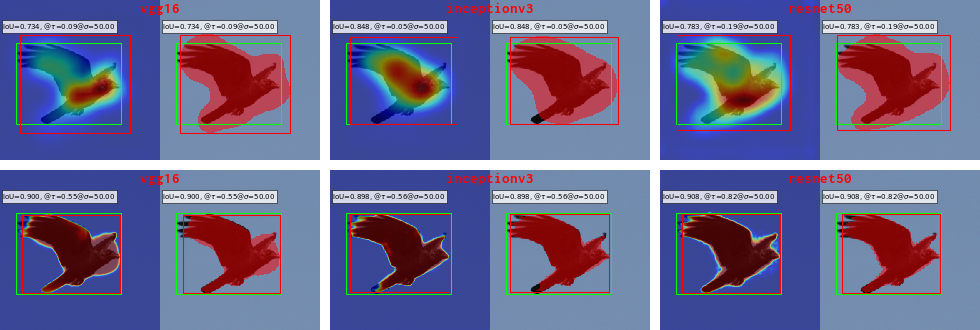}
     \end{subfigure}
     \hfill
     \begin{subfigure}[b]{0.49\textwidth}
         \centering
         \includegraphics[scale=.17]{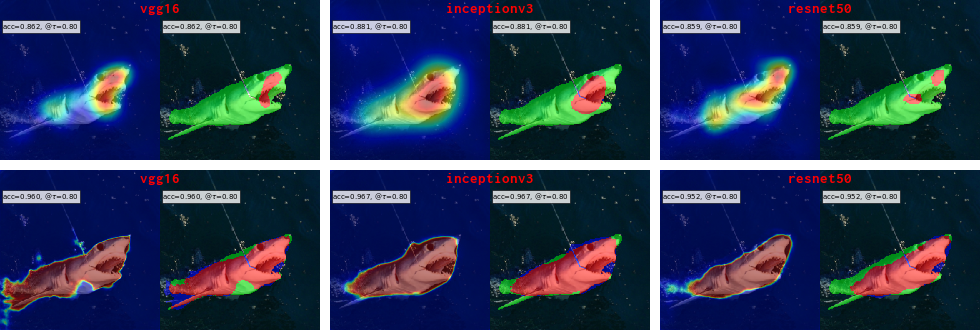}
     \end{subfigure}
      \\
     \vspace{0.1cm}
     \begin{subfigure}[b]{0.49\textwidth}
         \centering
         \includegraphics[scale=.17]{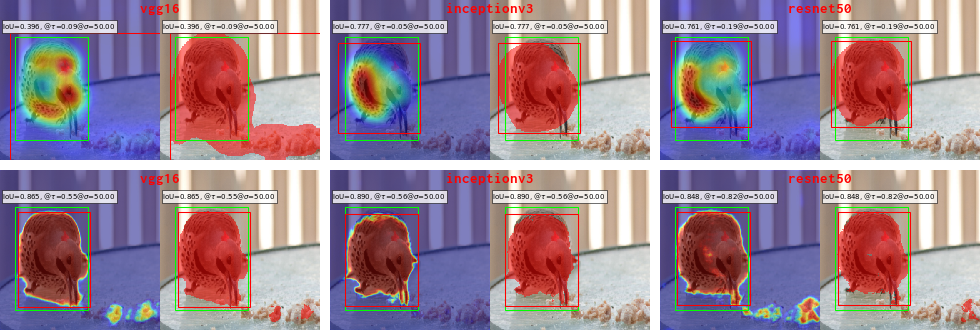}
     \end{subfigure}
     \hfill
     \begin{subfigure}[b]{0.49\textwidth}
         \centering
         \includegraphics[scale=.17]{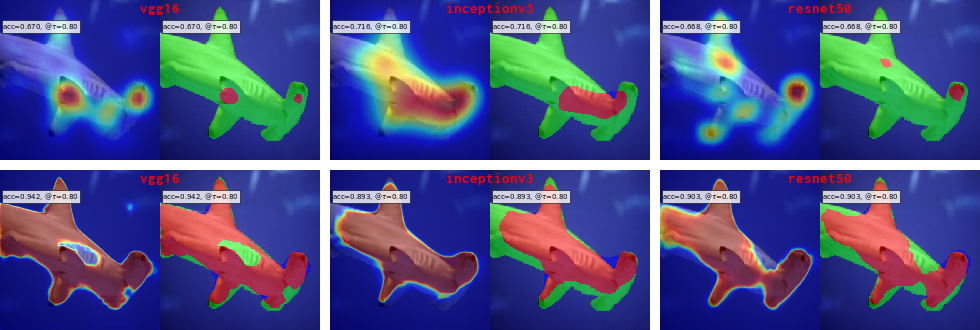}
     \end{subfigure}
      \\
     \vspace{0.1cm}
     \begin{subfigure}[b]{0.49\textwidth}
         \centering
         \includegraphics[scale=.17]{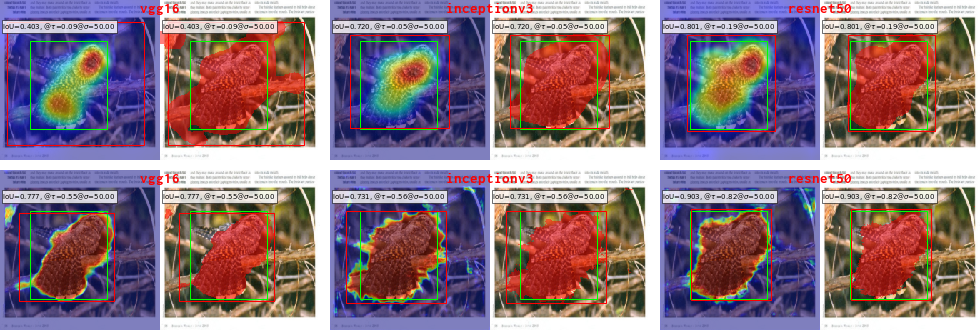}
     \end{subfigure}
     \hfill
     \begin{subfigure}[b]{0.49\textwidth}
         \centering
         \includegraphics[scale=.17]{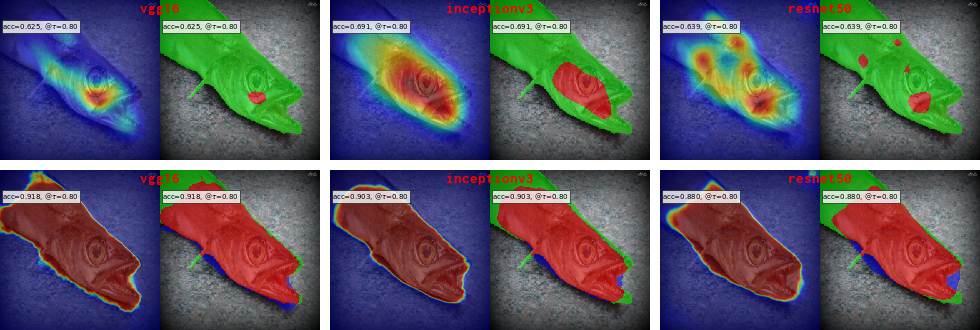}
     \end{subfigure}
      \\
     \vspace{0.1cm}
     \begin{subfigure}[b]{0.49\textwidth}
         \centering
         \includegraphics[scale=.17]{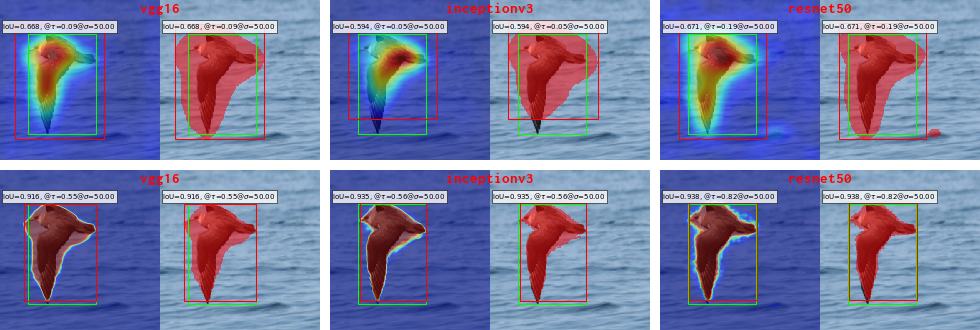}
     \end{subfigure}
     \hfill
     \begin{subfigure}[b]{0.49\textwidth}
         \centering
         \includegraphics[scale=.17]{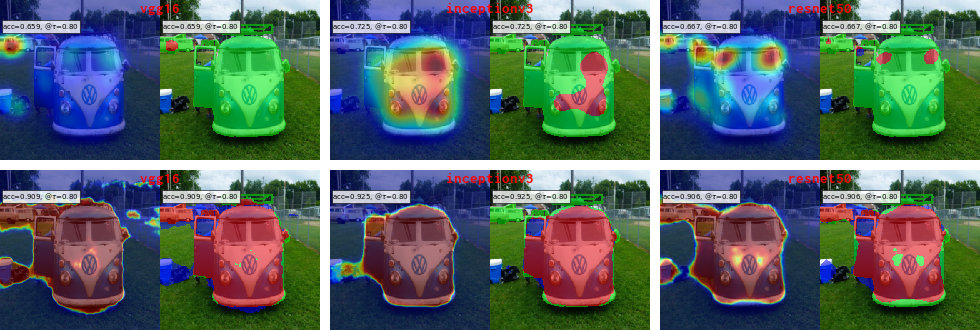}
     \end{subfigure}
     \\
     \vspace{0.1cm}
     \begin{subfigure}[b]{0.49\textwidth}
         \centering
         \includegraphics[scale=.17]{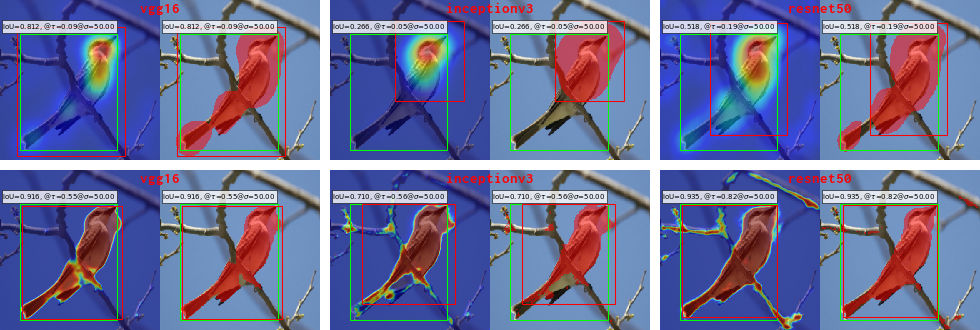}
     \end{subfigure}
     \hfill
     \begin{subfigure}[b]{0.49\textwidth}
         \centering
         \includegraphics[scale=.17]{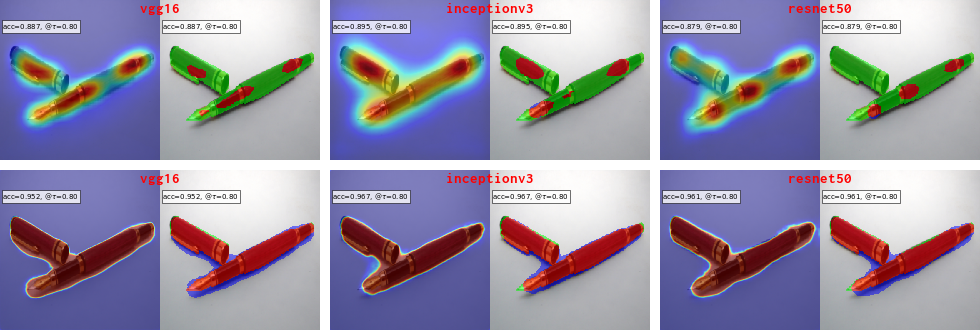}
     \end{subfigure}
        \caption{LayerCAM method examples for three backbones (left to right: VGG16, Inceptionv3, ResNet50): baselines (top) vs. baseline + ours (bottom)  validated with \maxboxacc. Colors: CUB (left): green box : ground truth. red box: predicted. red mask: thresholded CAM. OpenImages (right): red mask: true positive. green mask: false negative. blue mask: false positive. ${\tau=50, \sigma=0.8}$.}
        \label{fig:layercam-cub-openim-example-pred}
\end{figure*}

\begin{figure*}
     \centering
     \begin{subfigure}[b]{0.49\textwidth}
         \centering
         \includegraphics[scale=.17]{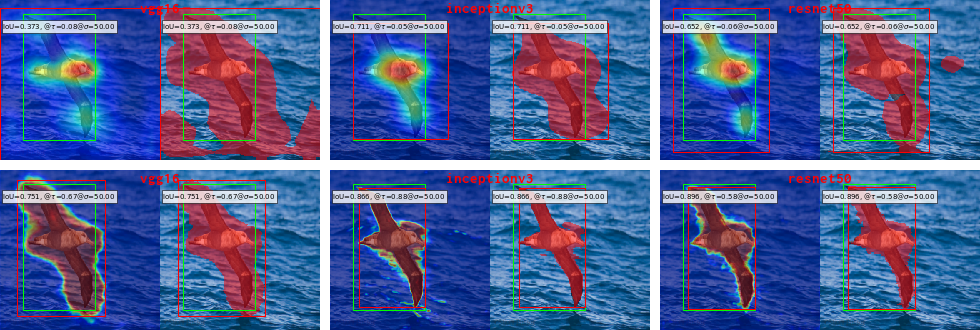}
     \end{subfigure}
     \hfill
     \begin{subfigure}[b]{0.49\textwidth}
         \centering
         \includegraphics[scale=.17]{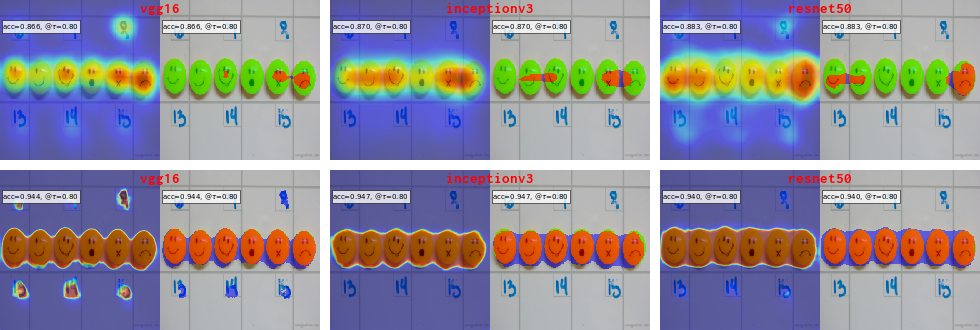}
     \end{subfigure}
     \\
     \vspace{0.1cm}
     \begin{subfigure}[b]{0.49\textwidth}
         \centering
         \includegraphics[scale=.17]{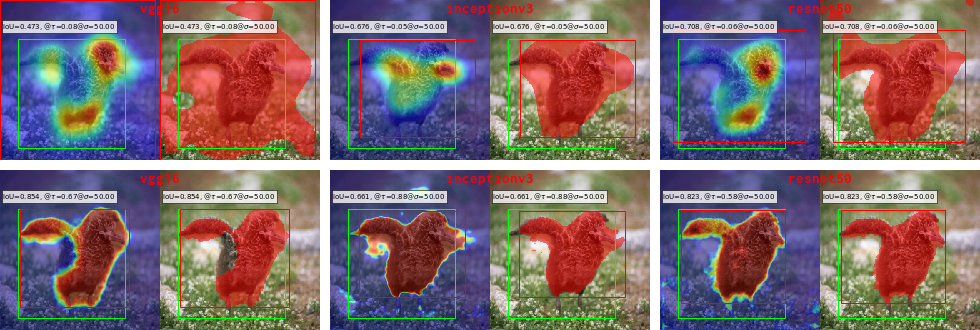}
     \end{subfigure}
     \hfill
     \begin{subfigure}[b]{0.49\textwidth}
         \centering
         \includegraphics[scale=.17]{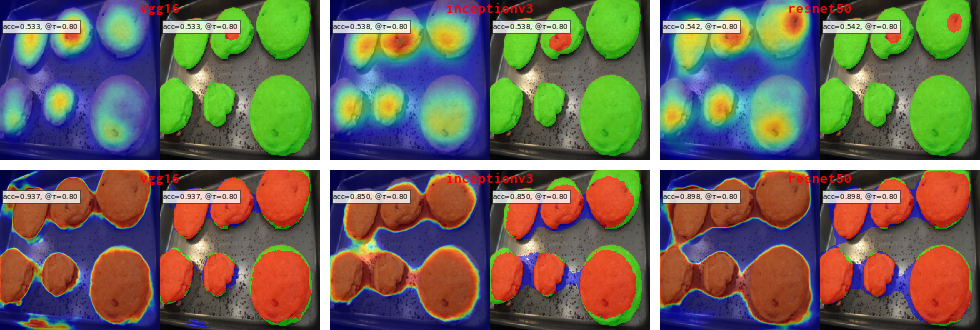}
     \end{subfigure}
     \\
     \vspace{0.1cm}
     \begin{subfigure}[b]{0.49\textwidth}
         \centering
         \includegraphics[scale=.17]{CUB-SmoothGradCAMpp-002-Laysan_Albatross_Laysan_Albatross_0098_621}
     \end{subfigure}
     \hfill
     \begin{subfigure}[b]{0.49\textwidth}
         \centering
         \includegraphics[scale=.17]{OpenImages-SmoothGradCAMpp-test_021mn_26407df923f45ca0}
     \end{subfigure}
     \\
     \vspace{0.1cm}
     \begin{subfigure}[b]{0.49\textwidth}
         \centering
         \includegraphics[scale=.17]{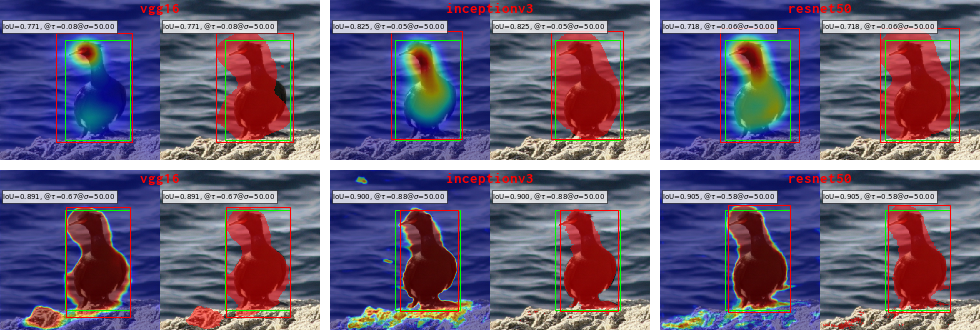}
     \end{subfigure}
     \hfill
     \begin{subfigure}[b]{0.49\textwidth}
         \centering
         \includegraphics[scale=.17]{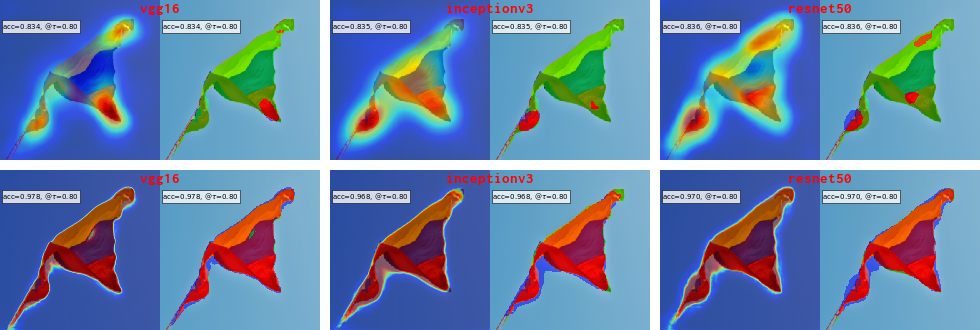}
     \end{subfigure}
     \\
     \vspace{0.1cm}
     \begin{subfigure}[b]{0.49\textwidth}
         \centering
         \includegraphics[scale=.17]{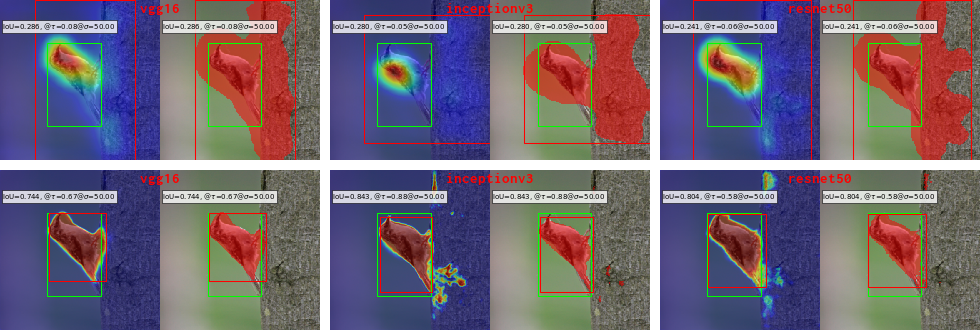}
     \end{subfigure}
     \hfill
     \begin{subfigure}[b]{0.49\textwidth}
         \centering
         \includegraphics[scale=.17]{OpenImages-SmoothGradCAMpp-test_04_sv_e64b1b3421e948a5}
     \end{subfigure}
     \\
     \vspace{0.1cm}
     \begin{subfigure}[b]{0.49\textwidth}
         \centering
         \includegraphics[scale=.17]{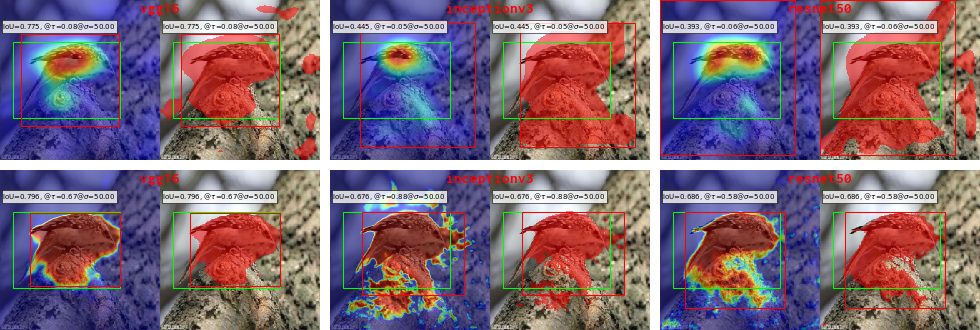}
     \end{subfigure}
     \hfill
     \begin{subfigure}[b]{0.49\textwidth}
         \centering
         \includegraphics[scale=.17]{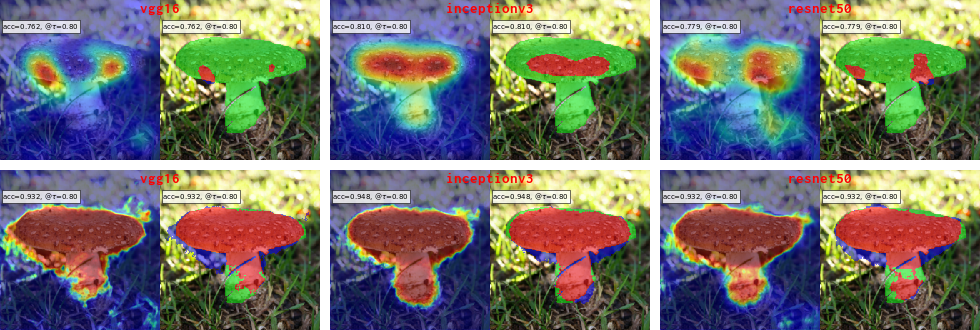}
     \end{subfigure}
      \\
     \vspace{0.1cm}
     \begin{subfigure}[b]{0.49\textwidth}
         \centering
         \includegraphics[scale=.17]{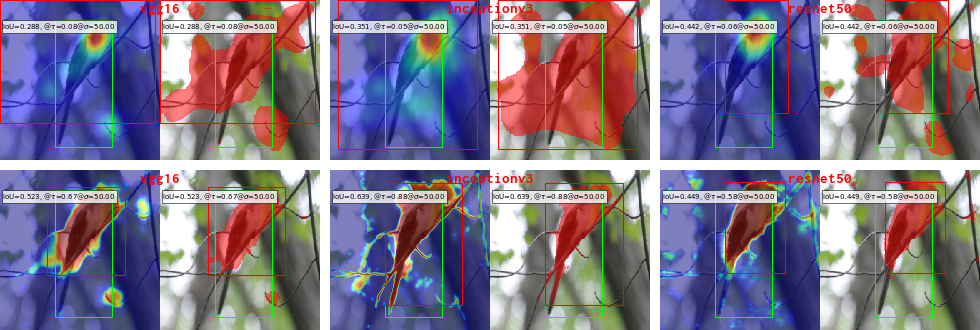}
     \end{subfigure}
     \hfill
     \begin{subfigure}[b]{0.49\textwidth}
         \centering
         \includegraphics[scale=.17]{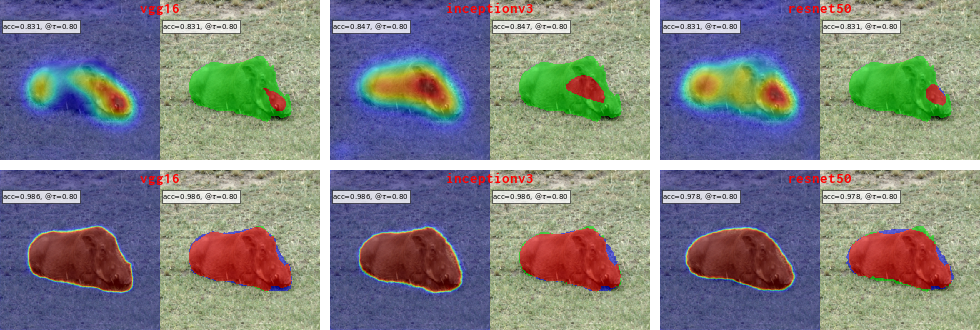}
     \end{subfigure}
      \\
     \vspace{0.1cm}
     \begin{subfigure}[b]{0.49\textwidth}
         \centering
         \includegraphics[scale=.17]{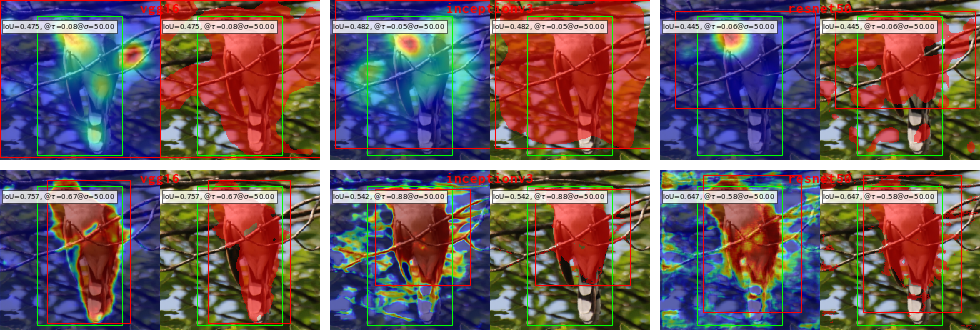}
     \end{subfigure}
     \hfill
     \begin{subfigure}[b]{0.49\textwidth}
         \centering
         \includegraphics[scale=.17]{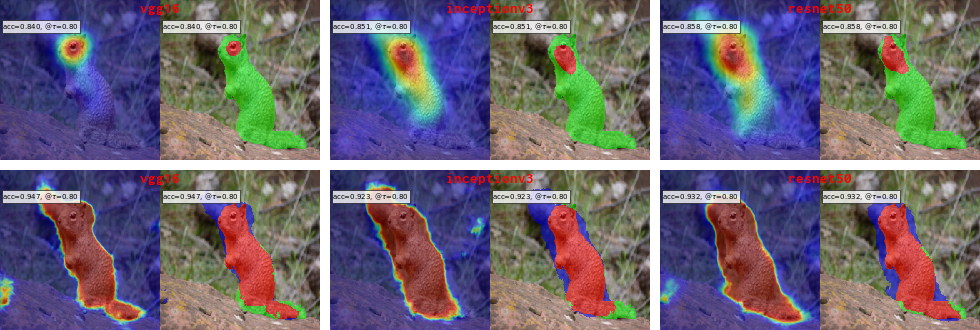}
     \end{subfigure}
      \\
     \vspace{0.1cm}
     \begin{subfigure}[b]{0.49\textwidth}
         \centering
         \includegraphics[scale=.17]{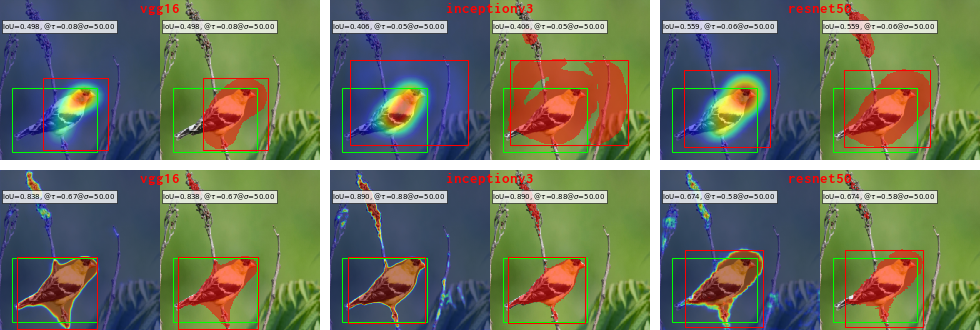}
       \end{subfigure}
     \hfill
     \begin{subfigure}[b]{0.49\textwidth}
         \centering
         \includegraphics[scale=.17]{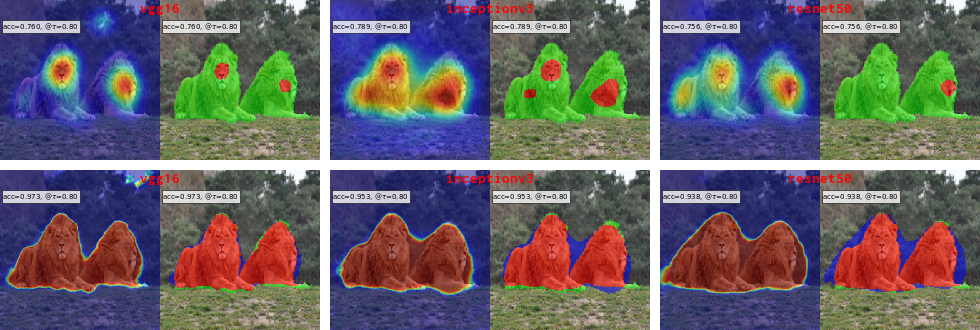}
     \end{subfigure}
     \\
     \vspace{0.1cm}
     \begin{subfigure}[b]{0.49\textwidth}
         \centering
         \includegraphics[scale=.17]{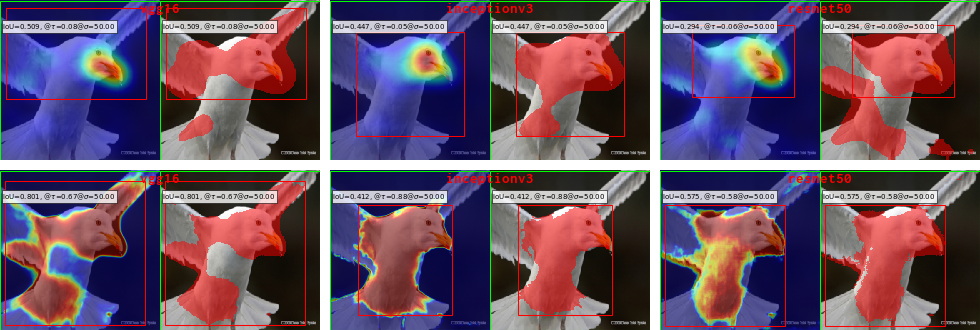}
     \end{subfigure}
     \hfill
     \begin{subfigure}[b]{0.49\textwidth}
         \centering
         \includegraphics[scale=.17]{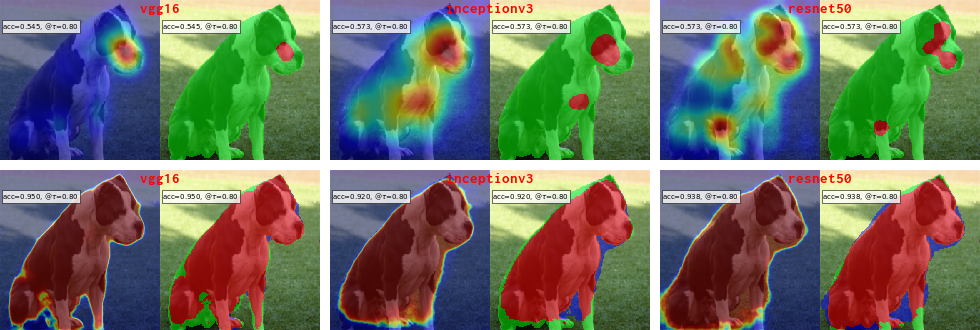}
     \end{subfigure}
        \caption{SmoothGradCAM++ method examples for three backbones (left to right: VGG16, Inceptionv3, ResNet50): baselines (top) vs. baseline + ours (bottom)  validated with \maxboxacc. Colors: CUB (left): green box : ground truth. red box: predicted. red mask: thresholded CAM. OpenImages (right): red mask: true positive. green mask: false negative. blue mask: false positive. ${\tau=50, \sigma=0.8}$.}
        \label{fig:smoothgradcampp-cub-openim-example-pred}
\end{figure*}

\begin{figure*}
     \centering
     \begin{subfigure}[b]{0.49\textwidth}
         \centering
         \includegraphics[scale=.17]{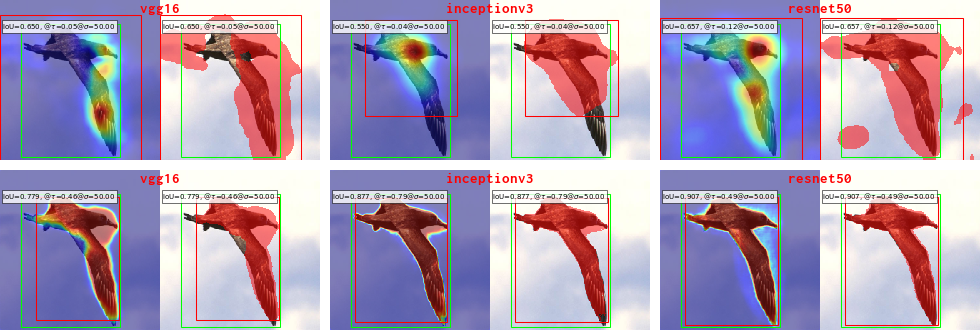}
     \end{subfigure}
     \hfill
     \begin{subfigure}[b]{0.49\textwidth}
         \centering
         \includegraphics[scale=.17]{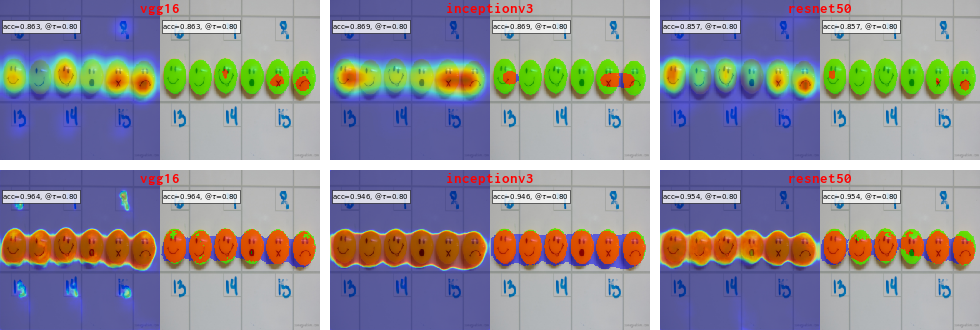}
     \end{subfigure}
     \\
     \vspace{0.1cm}
     \begin{subfigure}[b]{0.49\textwidth}
         \centering
         \includegraphics[scale=.17]{CUB-XGradCAM-001-Black_footed_Albatross_Black_Footed_Albatross_0024_796089}
     \end{subfigure}
     \hfill
     \begin{subfigure}[b]{0.49\textwidth}
         \centering
         \includegraphics[scale=.17]{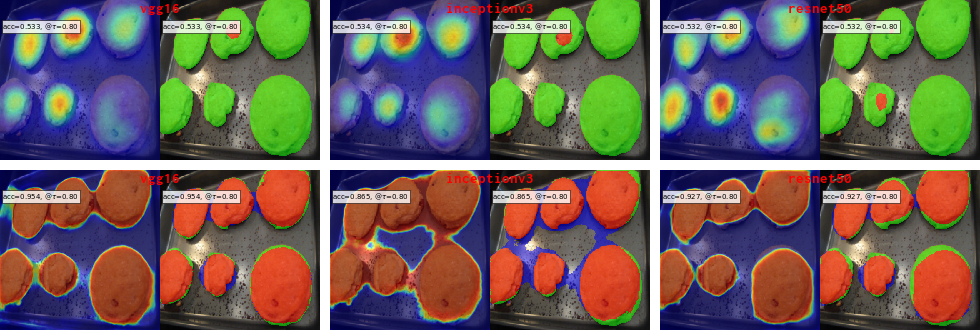}
     \end{subfigure}
     \\
     \vspace{0.1cm}
     \begin{subfigure}[b]{0.49\textwidth}
         \centering
         \includegraphics[scale=.17]{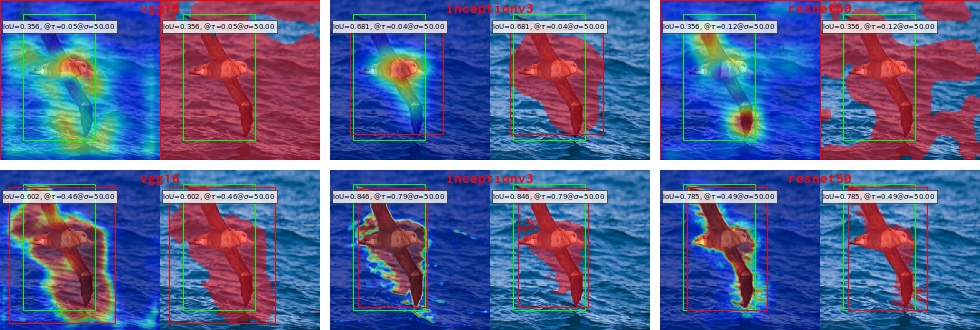}
     \end{subfigure}
     \hfill
     \begin{subfigure}[b]{0.49\textwidth}
         \centering
         \includegraphics[scale=.17]{OpenImages-XGradCAM-test_06j2d_18878c051c43ae73}
     \end{subfigure}
     \\
     \vspace{0.1cm}
     \begin{subfigure}[b]{0.49\textwidth}
         \centering
         \includegraphics[scale=.17]{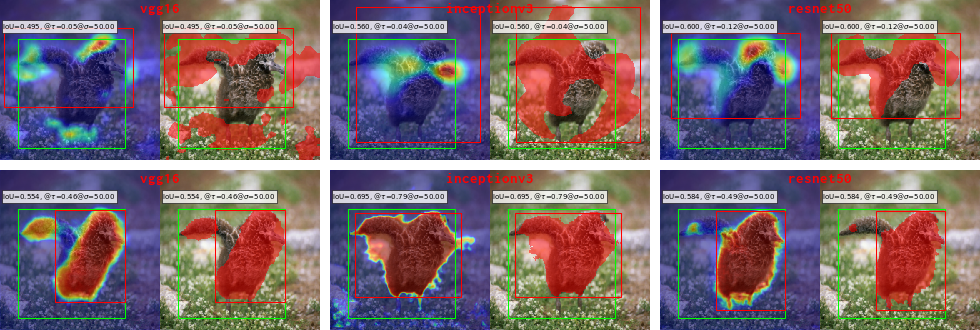}
     \end{subfigure}
     \hfill
     \begin{subfigure}[b]{0.49\textwidth}
         \centering
         \includegraphics[scale=.17]{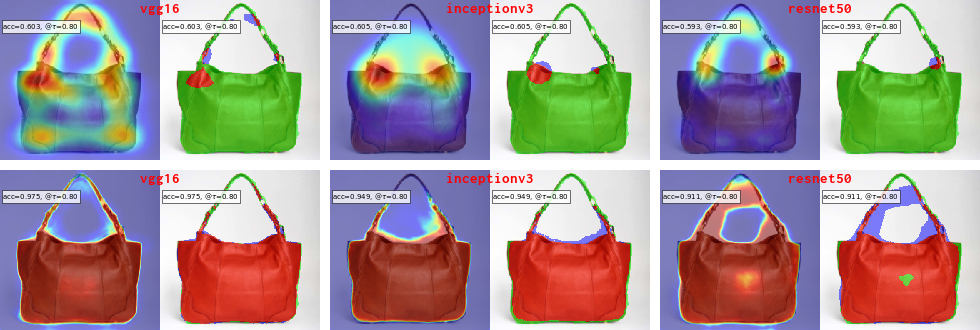}
     \end{subfigure}
     \\
     \vspace{0.1cm}
     \begin{subfigure}[b]{0.49\textwidth}
         \centering
         \includegraphics[scale=.17]{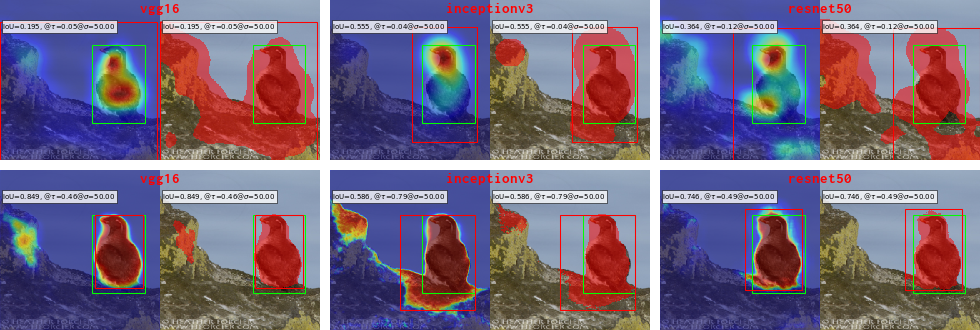}
     \end{subfigure}
     \hfill
     \begin{subfigure}[b]{0.49\textwidth}
         \centering
         \includegraphics[scale=.17]{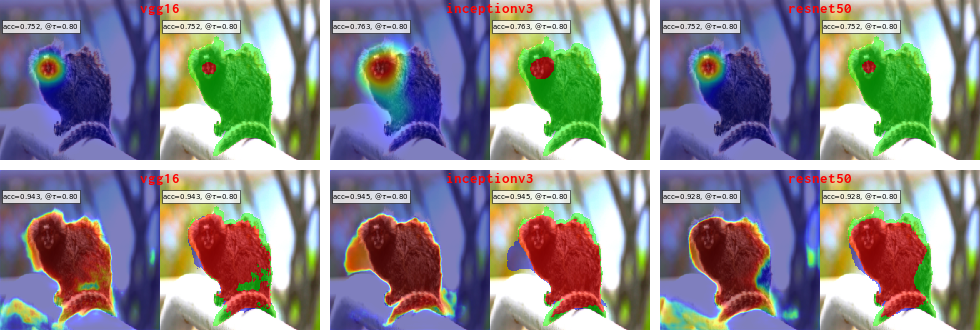}
     \end{subfigure}
     \\
     \vspace{0.1cm}
     \begin{subfigure}[b]{0.49\textwidth}
         \centering
         \includegraphics[scale=.17]{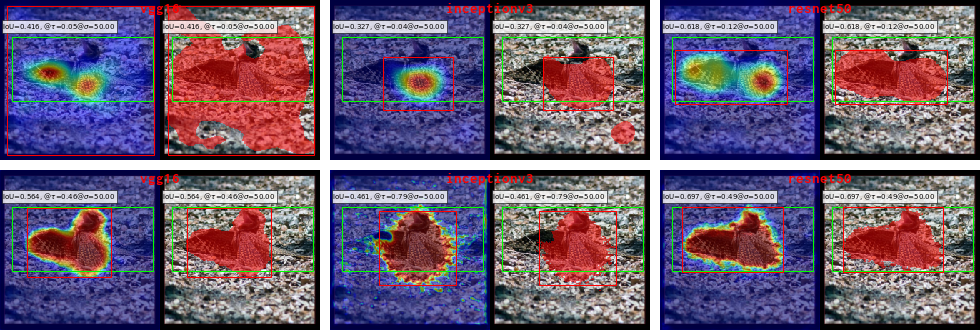}
     \end{subfigure}
     \hfill
     \begin{subfigure}[b]{0.49\textwidth}
         \centering
         \includegraphics[scale=.17]{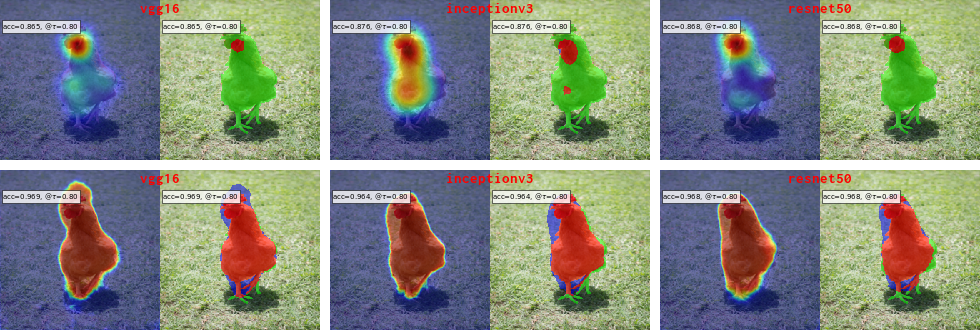}
     \end{subfigure}
      \\
     \vspace{0.1cm}
     \begin{subfigure}[b]{0.49\textwidth}
         \centering
         \includegraphics[scale=.17]{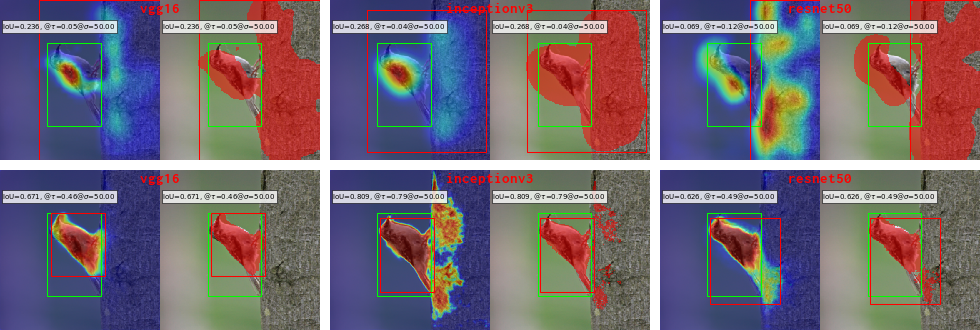}
     \end{subfigure}
     \hfill
     \begin{subfigure}[b]{0.49\textwidth}
         \centering
         \includegraphics[scale=.17]{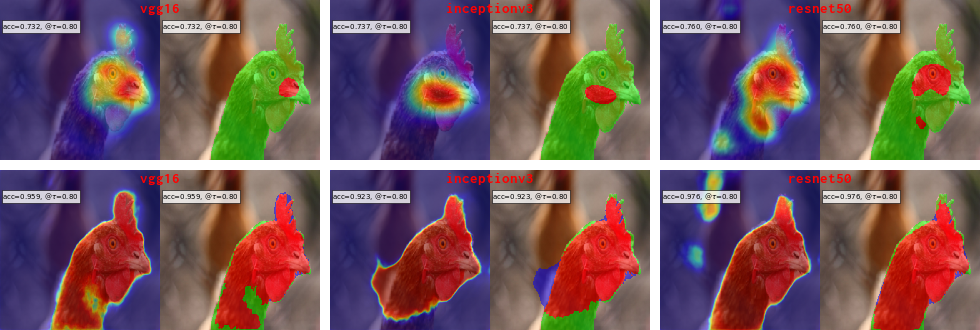}
     \end{subfigure}
      \\
     \vspace{0.1cm}
     \begin{subfigure}[b]{0.49\textwidth}
         \centering
         \includegraphics[scale=.17]{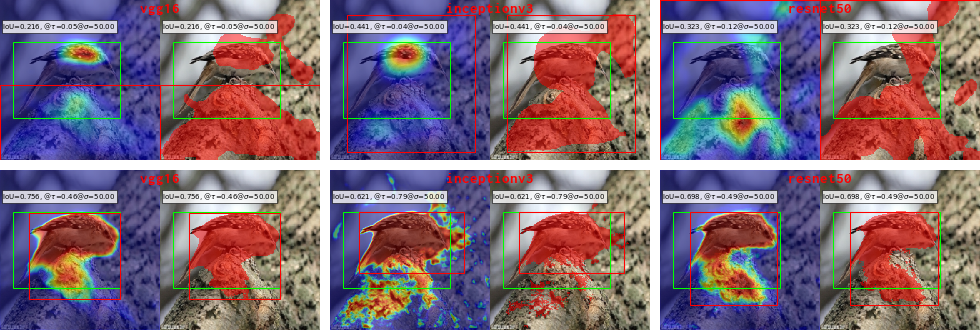}
     \end{subfigure}
     \hfill
     \begin{subfigure}[b]{0.49\textwidth}
         \centering
         \includegraphics[scale=.17]{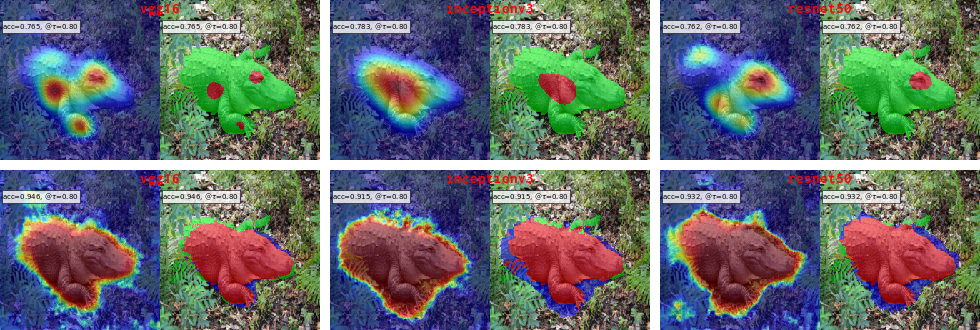}
     \end{subfigure}
      \\
     \vspace{0.1cm}
     \begin{subfigure}[b]{0.49\textwidth}
         \centering
         \includegraphics[scale=.17]{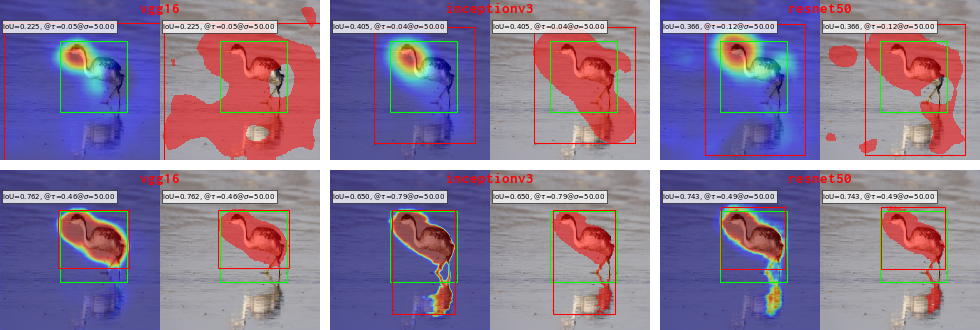}
     \end{subfigure}
     \hfill
     \begin{subfigure}[b]{0.49\textwidth}
         \centering
         \includegraphics[scale=.17]{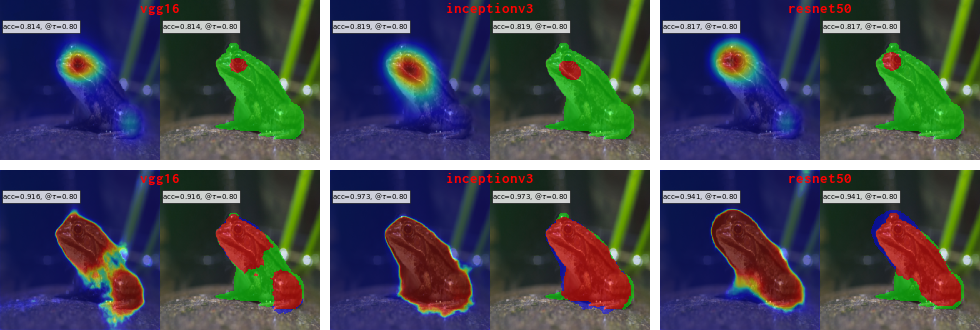}
     \end{subfigure}
     \\
     \vspace{0.1cm}
     \begin{subfigure}[b]{0.49\textwidth}
         \centering
         \includegraphics[scale=.17]{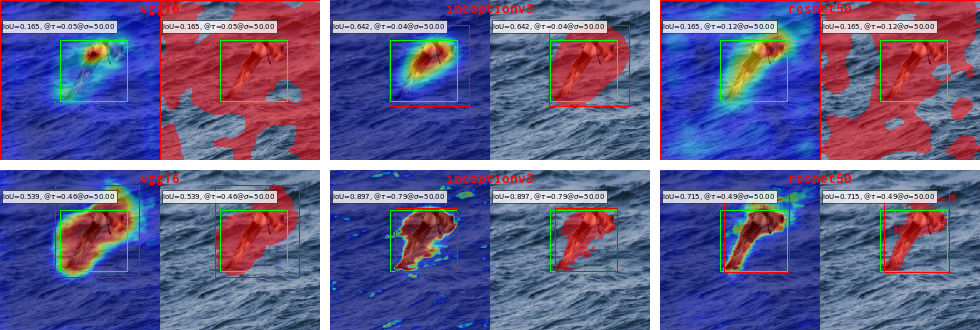}
     \end{subfigure}
     \hfill
     \begin{subfigure}[b]{0.49\textwidth}
         \centering
         \includegraphics[scale=.17]{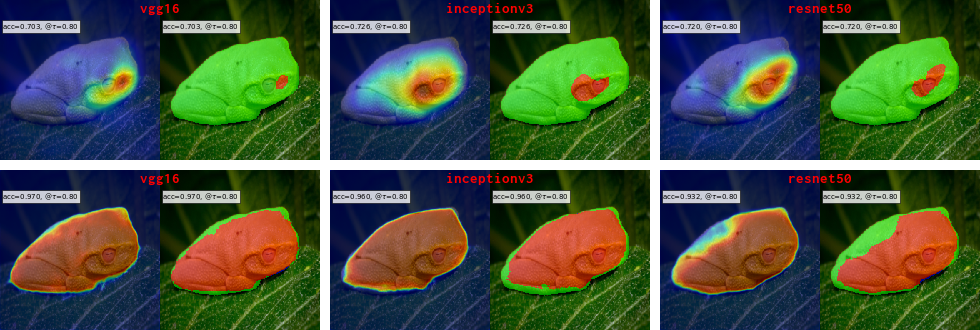}
     \end{subfigure}
        \caption{XGradCAM method examples for three backbones (left to right: VGG16, Inceptionv3, ResNet50): baselines (top) vs. baseline + ours (bottom)  validated with \maxboxacc. Colors: CUB (left): green box : ground truth. red box: predicted. red mask: thresholded CAM. OpenImages (right): red mask: true positive. green mask: false negative. blue mask: false positive. ${\tau=50, \sigma=0.8}$.}
        \label{fig:xgradcam-cub-openim-example-pred}
\end{figure*}

\FloatBarrier
\clearpage
\newpage

\bibliographystyle{abbrv}
\bibliography{biblio}

\end{document}

%% file: macros.tex
\usepackage{times}
\usepackage[numbers]{natbib}
\usepackage[english]{babel}
\usepackage{blindtext}
\usepackage{graphicx}
\usepackage{amsmath, amsthm, amssymb, bbm, bm}
\usepackage{enumerate}
\usepackage{float}      
\usepackage{subcaption}  
\usepackage{wrapfig}
\usepackage[margin=0cm]{caption}
\usepackage[titletoc, toc]{appendix}
\usepackage{tabularx}

\usepackage{multirow}
\usepackage{makecell}
\usepackage{placeins}  

\usepackage[x11names, usenames, dvipsnames, svgnames, table]{xcolor}
\definecolor{firebrick}{rgb}{.698,.133,.133}
\definecolor{mybluelight}{rgb}{0.9, 0.9, 1.}
\definecolor{myorangelight}{rgb}{1., 0.9, 0.9}

\usepackage[utf8]{inputenc} 
\usepackage[T1]{fontenc}    
\usepackage{url}            
\usepackage{booktabs, colortbl}       
\usepackage{amsfonts}       
\usepackage{nicefrac}       
\usepackage{microtype}      

\usepackage{csquotes}
\usepackage{latexsym}

\usepackage{pifont}
\usepackage[boxruled, vlined, linesnumbered]{algorithm2e}
\SetAlFnt{\small}
\SetAlCapFnt{\small}
\SetAlCapNameFnt{\small}
\usepackage{algorithmic}
\algsetup{linenosize=\tiny}

\let\oldnl\nl
\newcommand{\nonl}{\renewcommand{\nl}{\let\nl\oldnl}}

\usepackage{paralist}

\usepackage{xspace}
\usepackage{soul}
\usepackage{dsfont}
\usepackage{stmaryrd}
\usepackage[textwidth=15mm]{todonotes}
\usepackage{dirtytalk}
\usepackage{pbox}
\usepackage{cprotect}

\usepackage{verbatim}
\usepackage{textcomp}
\usepackage[normalem]{ulem}

\usepackage{mathtools}
\usepackage{etextools}
\usepackage[inline]{enumitem}

\usepackage{flushend}  

\usepackage[colorlinks=true,allcolors=firebrick,bookmarks=false]{hyperref}

\newcommand\maxboxacc{\texttt{MaxBoxAcc}\xspace}
\newcommand\newmaxboxacc{\texttt{MaxBoxAccV2}\xspace}

\newcommand\pxap{\texttt{PxAP}\xspace}
\newcommand\topone{\texttt{top-1}\xspace}
\newcommand\topfive{\texttt{top-5}\xspace}

\definecolor{darkergreen}{RGB}{21, 152, 56}
\definecolor{red2}{RGB}{252, 54, 65}
\definecolor{Gray}{gray}{0.85}
\newcolumntype{g}{>{\columncolor{Gray}}c}

\newcommand\tableplus[1]{\textcolor{darkergreen}{#1}}

\let\OLDthebibliography\thebibliography
\renewcommand\thebibliography[1]{
  \OLDthebibliography{#1}
  \setlength{\parskip}{0pt}
  \setlength{\itemsep}{0pt plus 0.3ex}
}

\newcommand{\reals}{\mathbb{R}}

\newcommand{\abs}[1]{\ensuremath \left| #1 \right|}

\theoremstyle{definition}

\DeclarePairedDelimiterX{\divx}[2]{(}{)}{%
  #1\;\delimsize\|\;#2%
}

\newcommand*{\ie}{\emph{i.e.}\@\xspace}